\documentclass[10pt,twocolumn,letterpaper]{article}

\usepackage[pagenumbers]{cvpr} 

\definecolor{cvprblue}{rgb}{0.21,0.49,0.74}
\usepackage[pagebackref,breaklinks,colorlinks,allcolors=cvprblue]{hyperref}

\def\paperID{12201} 
\def\confName{CVPR}
\def\confYear{2026}

\usepackage[utf8]{inputenc} 
\usepackage[T1]{fontenc}    
\usepackage{hyperref}       
\usepackage{url}            
\usepackage{booktabs}       
\usepackage{amsfonts}       
\usepackage{nicefrac}       
\usepackage{microtype}      
\usepackage{xcolor}         
\usepackage{amsmath}
\usepackage{pifont}
\newcommand{\cmark}{\textcolor{green!60!black}{\ding{51}}}  
\newcommand{\xmark}{\textcolor{red}{\ding{55}}}             
\usepackage{tabularx}

\usepackage{graphicx}
\usepackage{subcaption}
\usepackage{multirow}

\usepackage{caption}
\usepackage{float}
\usepackage{placeins}

\title{VIABench: A Comprehensive Video Benchmark Collected from Blind Individuals for Visual Impairment Assistance}

\author{
Yunfeng Liu\textsuperscript{1,*}
\quad
Yuandong Yang\textsuperscript{1,*}
\quad
Jiarui Han\textsuperscript{1}
\quad
Zhenpeng Huang\textsuperscript{1}
\\
Yuqing Tang\textsuperscript{1}
\quad
Xiangyu Zeng\textsuperscript{1,2}
\quad
Gangshan Wu\textsuperscript{1}
\quad
Limin Wang\textsuperscript{1,2,\ensuremath{\dagger}}
\\[0.5ex]
\textsuperscript{1}Nanjing University
\qquad
\textsuperscript{2}Shanghai AI Laboratory
}

\begin{document}

\maketitle
\begingroup
\renewcommand{\thefootnote}{}
\footnotetext{\textsuperscript{*}Equal contribution.
\textsuperscript{\ensuremath{\dagger}}Corresponding author.}
\addtocounter{footnote}{-1}
\endgroup

\begin{abstract}
  Visually impaired individuals (VIIs) encounter significant daily challenges due to limited access to visual information. Although Multimodal Large Language Models (MLLMs) have achieved impressive results on general vision and language tasks, their practical utility in real-world blind assistance still remains largely underexplored. To fill this gap, we introduce \textbf{VIABench}, a comprehensive video benchmark specifically designed to evaluate MLLMs in \textbf{V}isually \textbf{I}mpaired \textbf{A}ssistance scenarios using first-person videos recorded or shared by VIIs themselves. VIABench defines three core tasks, each targeting a distinct requirement in visual assistance. \textbf{Proactive Reminder}: Assesses the model’s ability to interpret ongoing video content while proactively anticipating and verbally describing upcoming navigation-critical events; \textbf{Visual Question Answering (VQA)}: Evaluates the model's capacity to answer user-posed questions about the environment or objects within the video; \textbf{Vision-Guided Interaction}: Tests context-aware reasoning to accomplish intentional interactions between user and environment. To ensure a robust and fair evaluation, we propose a rigorous benchmarking pipeline that supports both online (real-time) and offline settings. Our experiments demonstrate that current MLLMs still struggle to deliver comprehensive support for VIIs, especially in the \textbf{Proactive Reminder} task, which demands accurate anticipation and real-time responsiveness. We hope VIABench will drive future research toward developing customized MLLMs for real-world assistance, ultimately improving navigation and interaction experiences for visually impaired individuals. Code and data will be released at \url{https://github.com/MCG-NJU/VIABench}.
\end{abstract}

\begin{figure*}[t]
    \centering
    \includegraphics[width=\linewidth]{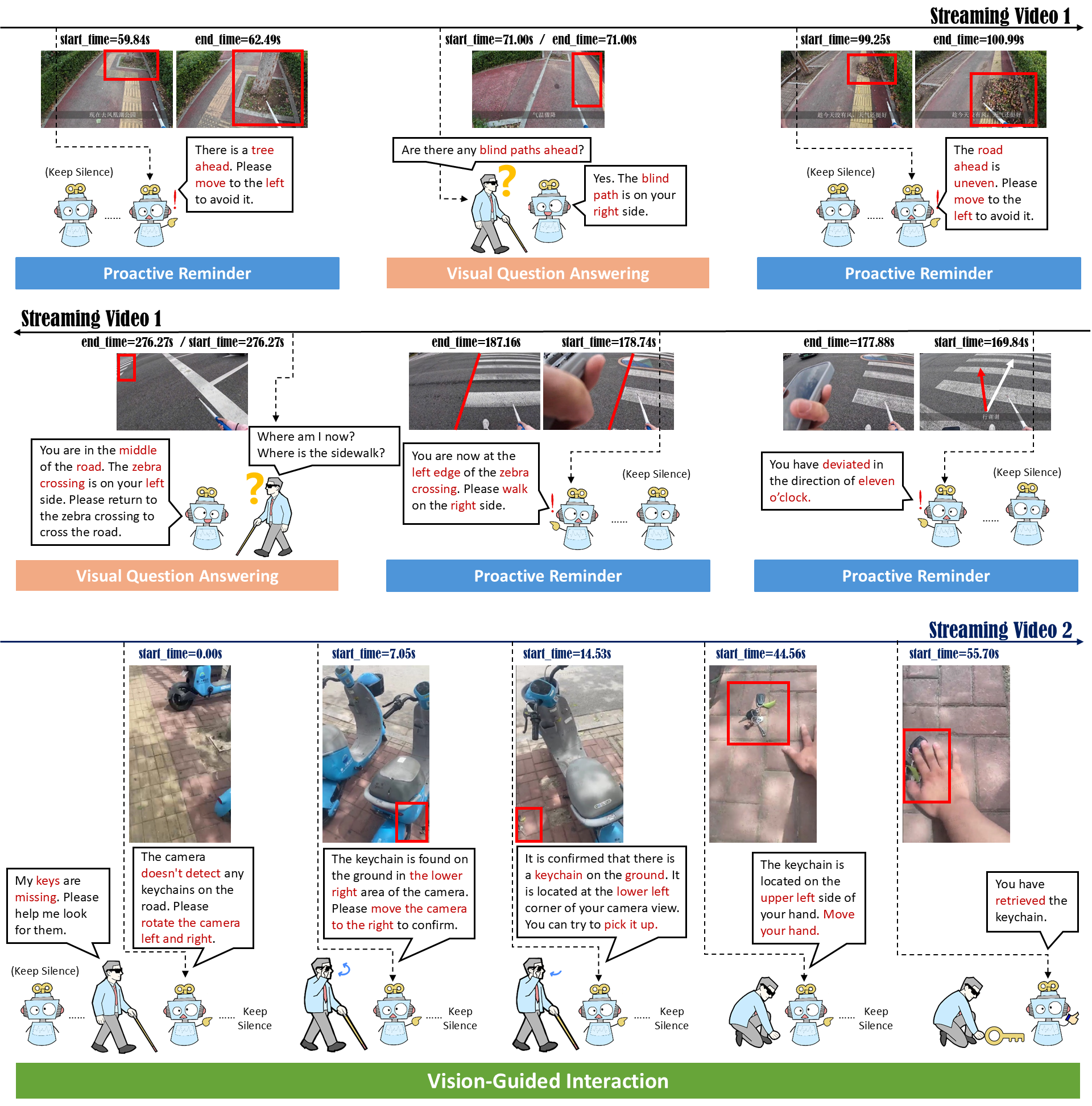}
    \caption{\textbf{Task definitions in VIABench}. Our benchmark contains 3 core tasks for visual impairment assistance, corresponding to three types of user behavior when interacting with the model: (1) doing nothing, (2) asking a question, and (3) receiving and acting on iterative guidance to complete a goal. We highlight the object relevant to the annotation in a red box for better visibility.}
    \label{fig:fig_1}
\end{figure*}

\section{Introduction}
Vision plays a critical role in perceiving surroundings, navigating environments, reading signs, identifying objects, and interpreting social cues. Visually impaired individuals (VIIs) encounter significant daily challenges due to limited access to visual information. For example, even seemingly simple tasks—such as crossing a street, finding a store entrance, or reading a menu—can become major obstacles. While traditional mobility aids like white canes and guide dogs can provide some support, they offer only partial solutions and fail to convey high-level semantic or contextual information.

Although a number of traditional vision algorithms \citep{yu2019street,yu2019lytnet,xia2023dataset,tian2021dynamic} have been proposed for assistive navigation, they remain limited in practical applicability due to their single-task design and poor generalization to real-world scenarios. Recent advances in Multimodal Large Language Models (MLLMs) \citep{team2024gemini,wang2024qwen2,yao2024minicpm,videochat} have already demonstrated exceptional performance in video understanding tasks, excelling on the existing video benchmarks \citep{mvbench,cgbench,fu2024video,wang2024lvbench,zhou2024mlvu}. With their inherent support for multi-task learning, powerful zero-shot performance, and ability to interact with users through natural language, a Jarvis-like intelligent visual assistant for the visually impaired is becoming increasingly feasible.

However, evaluating a general-purpose visual assistant for individuals with visual impairments requires a comprehensive benchmark that closely mirrors real-world usage scenarios. Although several recent datasets \citep{yuan2024walkvlm,xiao2025egoblind} have been introduced to assess the capabilities of large models in assistive contexts, they fall short in capturing the complexity and dynamic nature of everyday experiences encountered by visually impaired users. For instance, the WAD dataset \citep{yuan2024walkvlm} comprises 3-second video clips sourced from YouTube travel vlogs—content that is typically edited and curated for sighted audiences, thereby introducing a domain gap. While EgoBlind \citep{xiao2025egoblind} features videos recorded by blind users themselves, its tasks remain limited to passive, user-initiated interactions—primarily framed as visual question answering—where the model only responds upon receiving an explicit prompt. Therefore, there still lacks a comprehensive and versatile video benchmark collected by blind individuals for visual impairment assistance in real-world scenarios.

To fill this gap, we propose VIABench, a comprehensive, time-aligned, egocentric video benchmark tailored for evaluating general-purpose assistive vision systems. As shown in Figure~\ref{fig:fig_1}, VIABench defines three core task types that reflect increasing levels of assistive complexity for visually impaired individuals:
\begin{itemize}
    \item \textbf{Proactive Reminder}, where the model must autonomously notify users of potential risks or useful cues without explicit prompts.
    
    \item \textbf{Visual Question Answering (VQA)}, where users ask spontaneous questions during navigation, and the model answers based on the current and past visual context.
    
    \item \textbf{Vision-Guided Interaction}, where the model provides iterative guidance based on users' instructions to help them adjust their viewpoint or physical movements until the target task is completed.
\end{itemize}

Specifically, VIABench offers \textbf{a large-scale collection of real-world, long-form egocentric videos from VIIs}, with substantially greater total duration and temporal richness than prior visual-assistance video benchmarks. It further unifies three essential assistive tasks within a single benchmark, capturing \textbf{both autonomous and user-initiated} behaviors for comprehensive assessment.




To assess the real-world utility of state-of-the-art multimodal large models (MLLMs) in assistive scenarios for visually impaired users, we evaluate several leading video understanding models on the proposed VIABench benchmark. Our evaluation primarily focuses on the Proactive Reminder task, which poses a unique challenge: models must proactively identify and communicate critical events while remaining silent when no relevant information is present. However, existing online video understanding benchmarks \citep{lin2024streamingbench,li2025ovo} are not well aligned with the requirements of reminder-style tasks. Their evaluation pipelines are often heavy and rely on sparsely annotated events; for example, StreamingBench \citep{lin2024streamingbench} repeatedly queries the model on frames near the ground-truth event intervals, leading to high computational cost and insufficient modeling of temporal dynamics.

To address these limitations, we introduce \textbf{Token-Level Prompt Activation Decoding (TPAD)}, a novel two-stage evaluation framework designed for Reminder-style tasks. TPAD enables efficient and reliable assessment of a model’s ability to balance action and inaction in real-time video streams. In summary, our contributions are three-fold:

\begin{enumerate}
    \item We introduce VIABench, a large-scale, real-world video benchmark for blind assistance, annotated for three key tasks (Proactive Reminder, VQA, and Vision-Guided Interaction). With its real-time nature and diverse content, VIABench serves as a challenging and realistic testbed for video understanding.

    \item We propose \textbf{TPAD}, a two-stage evaluation framework that addresses the core challenge of proactive alert generation in long, continuous videos, enabling efficient and meaningful assessment of Reminder tasks.

    \item We conduct a comprehensive evaluation of existing MLLMs on VIABench, providing valuable insights into their current limitations and potential in real-world assistive applications.
\end{enumerate}

\begin{table*}[htbp]
\centering
\caption{Comparison of datasets for visual impairment assistance across key aspects: 
\textbf{Modality} (image or video), 
\textbf{Anno.} (annotation: M-Manual, A-Automatic), 
\textbf{Videos} (number of videos), 
\textbf{Samples} (number of annotated samples), 
\textbf{Avg./Max. Duration (s)} (average and maximum duration in seconds), 
\textbf{Visually Impaired} (data collected by VIIs), 
\textbf{Reminder} (proactive reminder tasks), 
\textbf{VQA} (visual question answering tasks), 
\textbf{Interaction} (vision-guided interaction tasks), 
\textbf{TA} (time-aligned annotations), 
\textbf{Robust} (robustness evaluation).}
\label{tab:dataset_comparison}
\resizebox{\textwidth}{!}{%
\begin{tabular}{l c c c c c c c c c c c c}
\toprule
\textbf{Datasets} & \textbf{Modality} & \textbf{Anno.} & \textbf{Videos} & 
\textbf{Samples} & 
\begin{tabular}[c]{@{}c@{}}\textbf{Avg./Max.}\\\textbf{Duration(s)}\end{tabular} & 
\begin{tabular}[c]{@{}c@{}}\textbf{Total}\\\textbf{Duration(h)}\end{tabular} & 
\begin{tabular}[c]{@{}c@{}}\textbf{Visually}\\\textbf{Impaired}\end{tabular} & 
\textbf{Reminder} & \textbf{VQA} & \textbf{Interaction} & \textbf{TA} & \textbf{Robust} \\
\midrule
VizWiz \citep{gurari2018vizwiz} & I & M & -- & 31,173 & -- & - & \cmark & \xmark & \cmark & \xmark & \xmark & \xmark \\
EgoBlind \citep{xiao2025egoblind} & V & M / A & 1,392 & 5,311 & 40 / 120 + & 15.5 & \cmark & \xmark & \cmark & \xmark & \cmark & \xmark \\
WalkVLM \citep{yuan2024walkvlm} & I \& V & M / A & 12K & 12K & 3 / 3 & 12 & \xmark & \cmark & \cmark & \xmark & \xmark & \xmark \\
\midrule
\textbf{VIABench (Ours)} & V & M & 761 & 14,526 & 222 / 1959 & 47 & \cmark & \cmark & \cmark & \cmark & \cmark & \cmark \\
\bottomrule
\end{tabular}
}
\end{table*}


\section{Related Work}

\textbf{Multimodal Large Language Models.} MLLMs have rapidly advanced in their ability to jointly process visual and linguistic information, driven by large-scale multimodal pretraining, unified modeling frameworks, and increasingly powerful visual encoders. Representative models such as Gemini 1.5 \citep{team2024gemini}, Qwen2-VL \citep{wang2024qwen2}, MiniCPM-V \citep{yao2024minicpm}, and VideoChat \citep{videochat} showcase strong general-purpose video understanding capabilities across a wide range of tasks, including temporal reasoning, event localization, and open-ended video question answering. These models consistently achieve the state-of-the-art results on standardized video benchmarks \citep{mvbench,cgbench,fu2024video,wang2024lvbench,zhou2024mlvu}, demonstrating broad generalization and adaptability. Despite this progress, the existing capability-oriented benchmarks primarily assess models through carefully designed, challenge-style tasks. This leaves their goal-oriented performance in real-world, socially impactful scenarios (such as visual impairment assistance) largely underexplored. In this paper, we mainly focus on exploiting the MLLM for the task of VIA.

\noindent \textbf{Visual Impairment Assistance Datasets.}
Vision-and-language datasets for visually impaired individuals (VIIs) have evolved from static images to egocentric videos. VizWiz \citep{gurari2018vizwiz} pioneered this direction by collecting over 31K real-world images and spoken questions from blind users, highlighting challenges unique to VII-created content—low-quality visual content, conversational queries, and unanswerable questions. Subsequent image-based datasets, including VIAssist \citep{yang2024viassist} and GuideDog \citep{kim2025guidedog}, explore spatial understanding and user interaction, but remain limited to single-image inputs without temporal context. To incorporate temporal reasoning, recent efforts have turned to egocentric video. EgoBlind \citep{xiao2025egoblind} was the first large-scale VideoQA dataset captured by blind users, offering 1,392 videos with 5,311 questions to evaluate MLLMs. VIEW-QA \citep{song2024viewqa} extends this to 360° video, but both focus on post-hoc QA rather than live assistance. Moreover, their scope is restricted: general VQA tasks offer only narrow evaluations of models as comprehensive VI assistants. WalkVLM \citep{yuan2024walkvlm} came closer to proactive guidance by training on 12K walking-scene video clips (each $\sim$3 seconds) for real-time navigational reminders. However, the extremely short duration provides insufficient temporal context, and its curated YouTube travel videos differ substantially from the real-world egocentric videos encountered in visual impairment assistance, resulting in a significant distribution shift.



\section{VIABench}



\subsection{Task Definition}\label{task_definitions}
To reflect the diverse needs of real-world blind assistance, VIABench defines three core tasks, each addressing a complementary aspect of how AI models can support VIIs. These tasks are derived from practical assistance scenarios encountered in first-person videos recorded by VIIs, and together they form a holistic evaluation framework. Across all tasks, models are required to produce responses that are \textit{accurate}, \textit{concise}, and \textit{informative}.

\textbf{Proactive Reminder} serves as the central task in VIABench, designed to evaluate a model’s ability to perform online video understanding in dynamic, real-world scenarios. It requires models not only to recognize navigation-critical events but also to anticipate and proactively describe them before they occur, providing timely and actionable support to visually impaired users. The task comprises 21 fine-grained sub-tasks, with detailed definitions provided in the appendix.

\textbf{Visual Question Answering} evaluates a model’s ability to comprehend dynamic scenes and provide informative responses to user queries. This task reflects situations where users actively seek information about unfamiliar objects, surroundings, or signage within their field of view.



\textbf{Vision-Guided Interaction} evaluates the model's capacity to provide iterative, context-aware guidance to help a user complete a specific physical goal. Unlike VQA, this is a multi-turn, closed-loop task. The model must provide step-by-step instructions and continuously adapt its guidance based on the user's actions and the resulting change in visual input, until the goal is achieved.

\begin{figure*}
  \centering
  \includegraphics[width=\linewidth]{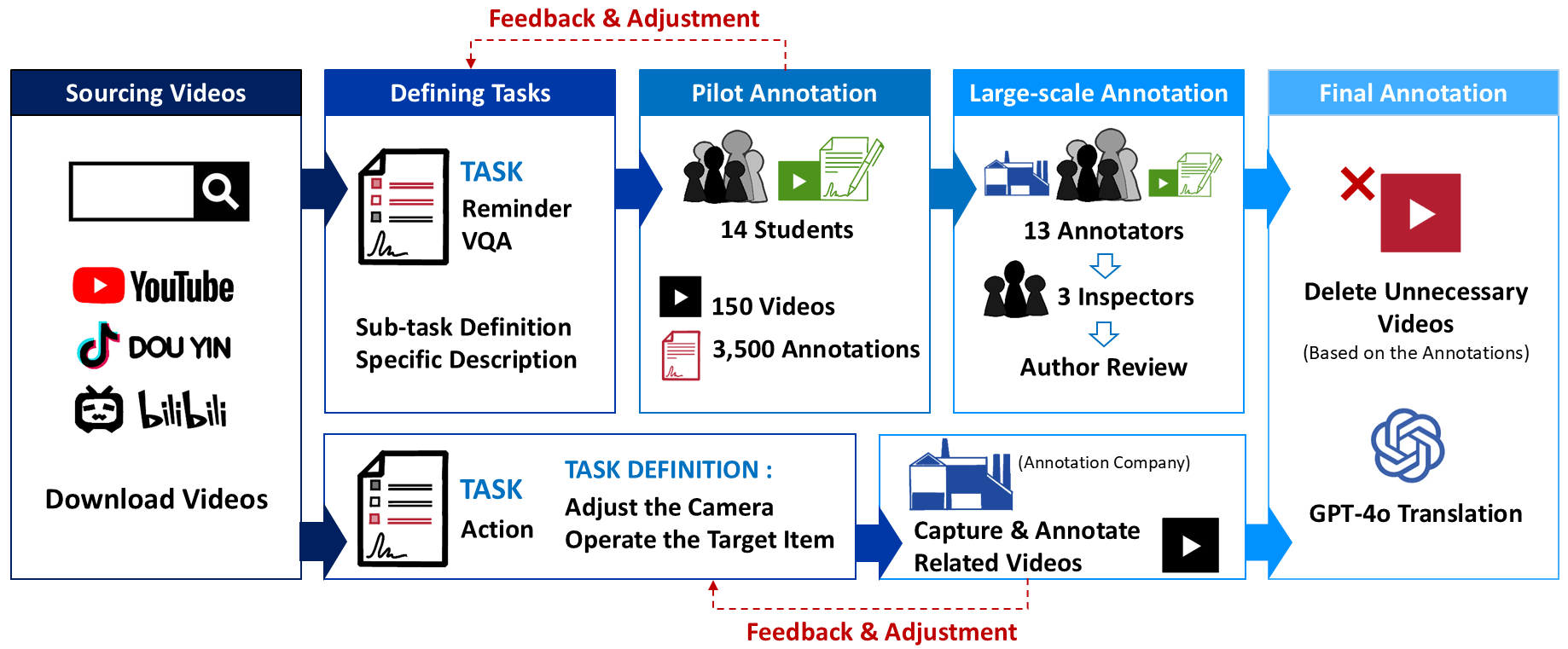}
  \caption{Overview of the VIABench data generation pipeline. The process consists of five key stages: (1) Sourcing Videos, (2) Defining Tasks, (3) Pilot Annotation, (4) Large-scale Annotation, and (5) Final Annotation. The pipeline highlights two parallel task-definition tracks and incorporates feedback loops to ensure data quality.}
  \label{fig:pipeline}
\end{figure*}

\subsection{Data Collection and Annotation}


We built VIABench entirely from scratch through a five-stage data construction pipeline, as illustrated in Figure~\ref{fig:pipeline}.

\textbf{1. Sourcing Videos:} 
Our data was sourced from two primary streams to ensure authenticity and task coverage.
\begin{itemize}
    \item \textbf{Real-World VII Videos:} We curated videos from visually impaired creators on platforms such as YouTube, Bilibili, and DouYin through keyword search and manual screening, selecting clips that reflect authentic daily scenarios and meet quality and ethical standards.
    
    \item \textbf{Informed Simulated Videos:} Since the \textit{Vision-Guided Interaction} scenarios are scarce in public data, we collected \textbf{231 purpose-filmed simulated videos}. These were recorded by trained annotators \textit{after} they had completed labeling real VII videos, enabling them to realistically reproduce VII behaviors, camera motions, and iterative interaction patterns.
\end{itemize}

\textbf{2. Defining Tasks:} 
Based on extensive examination of first-person videos from VIIs—as well as user feedback found in video comment sections—we identified three major categories of VIABench. The task definitions are illustrated in Section~\ref{task_definitions}.

\textbf{3. Pilot Annotation:} 
A group of 14 trained student annotators conducted pilot labeling over 150 videos, generating 3.5K annotations. These trials helped identify common pitfalls, improve clarity in task definitions, and validate the feasibility of fine-grained labeling—including precise start and end timestamps for events in the Proactive Reminder task.

\textbf{4. Large-scale Annotation:} 
Following the pilot, we collaborated with a professional annotation vendor to scale the labeling process. To ensure high annotation quality, we provided structured video-based training materials, conducted multi-stage QA with in-house experts, and implemented feedback loops between annotators and reviewers. 

\textbf{5. Final Annotation:} 
All finalized annotations were reviewed to remove redundant or irrelevant videos. Annotations were then translated into English using GPT-4o \citep{gpt4o} to ensure cross-lingual consistency and accessibility for the research community.

\begin{figure}[htbp]
\centering
\begin{subfigure}{0.45\textwidth}
    \includegraphics[width=\linewidth]{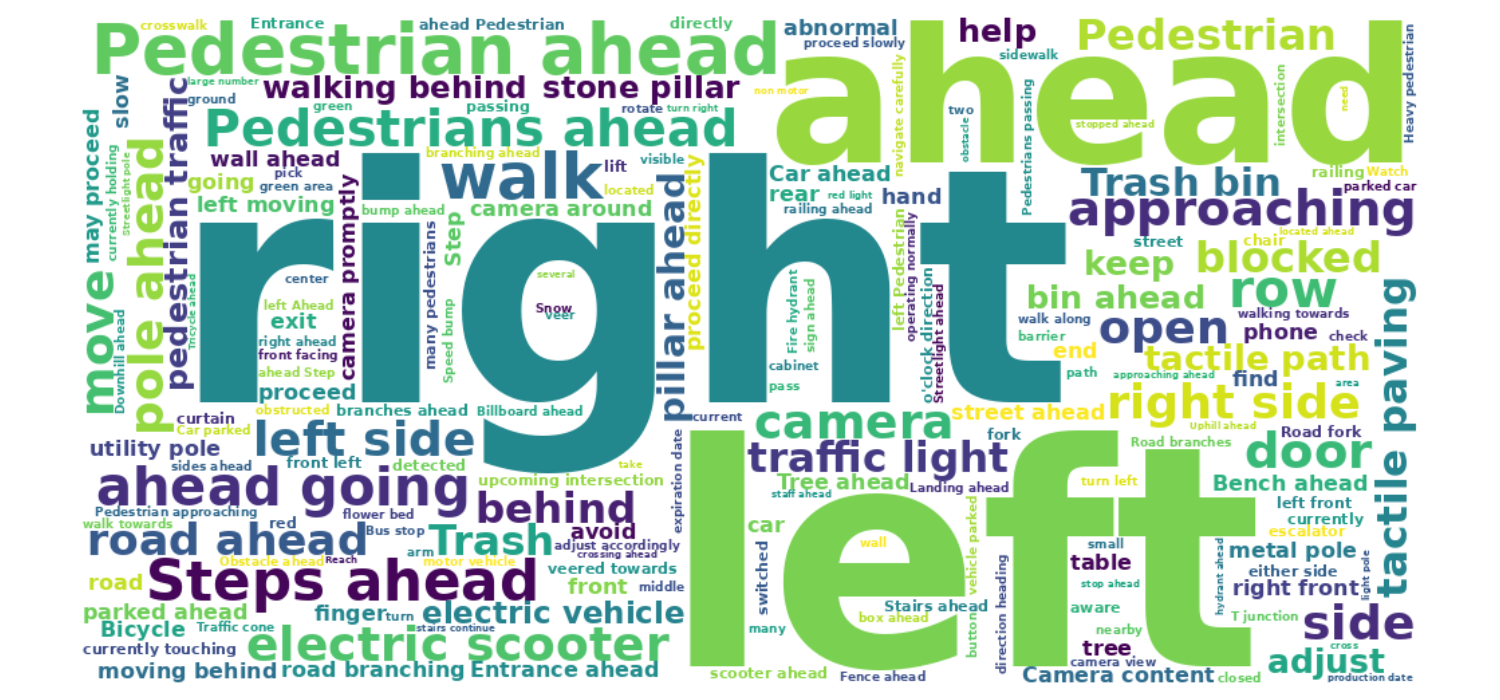}
\end{subfigure}
\begin{subfigure}{0.45\textwidth}
    \includegraphics[width=\linewidth]{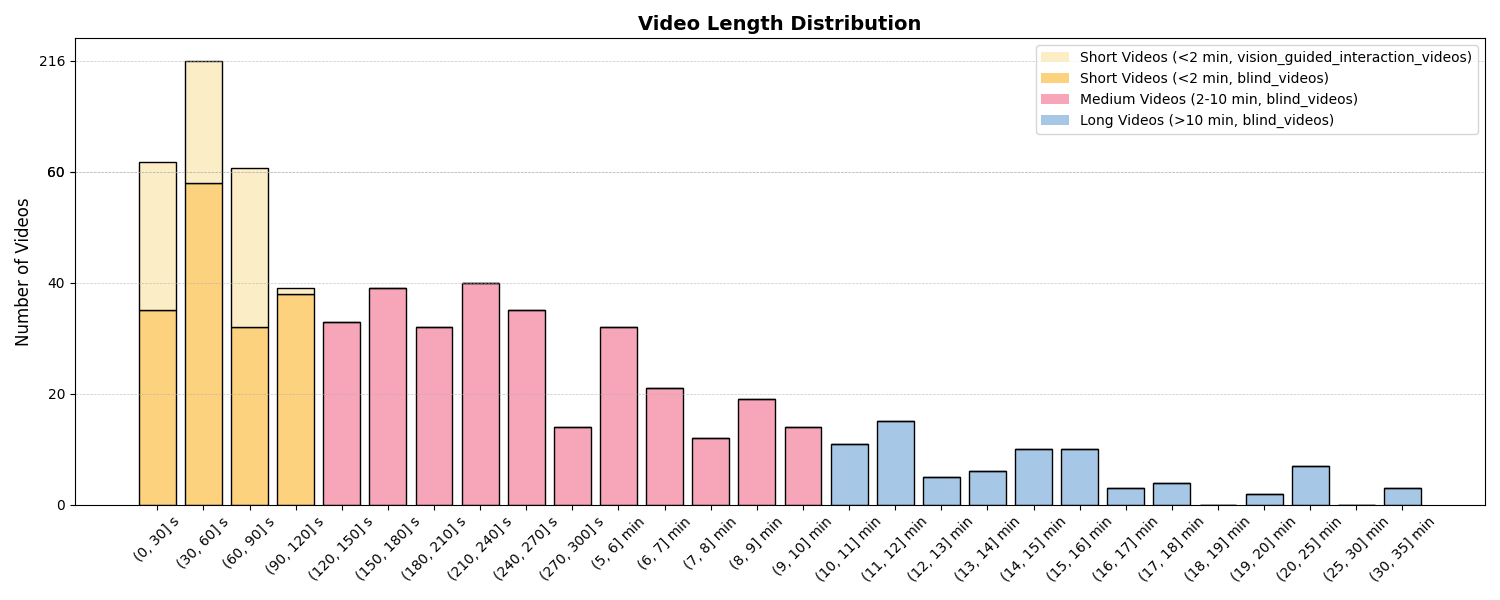}
\end{subfigure}
\caption{\textbf{(Up)} The word cloud of annotations highlights a strong focus on navigation and obstacle detection. \textbf{(Down)} A histogram illustrating the diverse distribution of video lengths, with a notable presence of long-form content.}
\label{fig:visual_distributions}
\end{figure}

\subsection{Statistics and Comparative Analysis}

To highlight the unique strengths of VIABench, we provide a statistical overview and comparison with existing datasets for visual impairment assistance (Table~\ref{tab:dataset_comparison}), supported by visual distributions in Figure~\ref{fig:visual_distributions}.

\textbf{Large-Scale, High-Quality, and Long-Form Video Data.}
VIABench is the most temporally rich dataset in this domain, containing \textbf{761 videos} and \textbf{14,526 manually curated annotations}, totaling \textbf{46.9 hours} of footage---substantially longer than prior datasets (e.g., approximately 15.5 hours in EgoBlind and 12 hours in WalkVLM). Crucially, VIABench emphasizes long-form video: the average duration is \textbf{222 seconds} (max: 1959s), compared with only 40s in EgoBlind and 3s in WalkVLM. The diverse length distribution in Figure~\ref{fig:visual_distributions} (\textbf{Down}) highlights our substantial collection of both short clips and extended videos, providing significantly richer temporal context for evaluating long-horizon reasoning and proactive understanding.

\textbf{Authentic Real-World Scenarios and Goal-Oriented Task Design.}
Unlike capability-driven datasets that artificially construct difficult queries, VIABench is grounded in real-world assistive needs. It comprises \textbf{530 real VII videos} ($\sim$44 hours) and \textbf{231 informed simulated videos} designed to complete the coverage of the \textit{Vision-Guided Interaction} task. The semantic distribution of annotations in Figure~\ref{fig:visual_distributions} (\textbf{Up})—dominated by directional cues (e.g., “right”, “left”, “ahead”) and obstacle identifiers (e.g., “pedestrian”, “scooter”, “steps”)—further demonstrates the dataset's strong alignment with navigation-centered, goal-oriented assistance rather than contrived test scenarios.

\textbf{Robustness Under Realistic Visual Degradation.}
A key challenge in visual impairment assistance is robustness to low-quality egocentric video. This challenge is inherently present in VIABench, as \textbf{94\%} of the data (\textbf{44 hours}) comes from visually impaired users, whose recordings often contain occlusion, abnormal exposure, and camera inversion/rotation. Similar observations were reported in VizWiz \citep{gurari2018vizwiz}, where visually impaired users cannot reliably control framing or visual quality. VIABench explicitly includes a robustness evaluation dimension (Table~\ref{tab:dataset_comparison}), requiring models to detect when the visual input itself becomes unreliable and promptly alert the user to adjust the camera. Failing to provide such warnings can lead to safety-critical situations for visually impaired users, making this capability essential for real-world deployment.


\section{Method}

Our goal is to systematically evaluate how existing Multimodal Large Language Models (MLLMs) can be adapted to assistive scenarios for visually impaired users. 
VIABench focuses on creating a unified evaluation framework that bridges the gap between conventional MLLM capabilities and real-world assistive demands. 
This section introduces how different categories of MLLMs are integrated into our benchmark, followed by a detailed description of our proposed \textbf{Token-Level Prompt Activation Decoding (TPAD)} mechanism, which enables offline MLLMs to participate in proactive reminder evaluation.

\subsection{Adapting MLLMs to Assistive Scenarios}

To comprehensively assess the applicability of existing Multimodal Large Language Models (MLLMs) in real assistive contexts, we design VIABench around three representative \textbf{assistive scenarios} faced by visually impaired users: \textbf{navigation}, \textbf{blind questioning}, and \textbf{interaction}.  
Each scenario corresponds to a distinct type of cognitive and perceptual challenge and is evaluated through one of the three core tasks in VIABench—\textbf{Proactive Reminder}, \textbf{Visual Question Answering (VQA)}, and \textbf{Vision-Guided Interaction}, respectively.

\paragraph{Navigation Scenario $\Rightarrow$ Proactive Reminder Task.} When navigating outdoor or indoor environments, visually impaired users rely on assistive systems to proactively warn them of potential hazards—such as steps, obstacles, or sidewalk boundaries—without explicit requests. This requires models not only to understand scene semantics but also to determine \textit{when} an alert should be issued. To assess this capability, we introduce the \textbf{Proactive Reminder} task, in which models must continuously monitor streaming visual input and trigger alerts at appropriate moments. While some online MLLMs—such as \textbf{VideoLLM-Online} \citep{chen2024videollm}—are designed for real-time video processing and can generate responses proactively, many existing MLLMs operate in an offline mode intended for static inference. To allow these offline models to participate in proactive reminder evaluation, we propose a lightweight adaptation mechanism, \textbf{Token-Level Prompt Activation Decoding (TPAD)}, which converts standard offline decoding into a frame-wise proactive scoring process (detailed in the next subsection). 

\paragraph{Blind Questioning Scenario $\Rightarrow$ Visual Question Answering Task.} In many daily situations, blind users may actively query their surroundings—for example, `What color is the traffic light now?'' The corresponding benchmark task, \textbf{Visual Question Answering (VQA)}, measures a model’s ability to provide accurate, real-time responses to such queries. Following the design philosophy of \textbf{OVOBench} \citep{li2025ovo}, we adopt an \textbf{online VQA setting}, where each question is answered based only on visual information available \textit{before} the query time. This prevents access to future frames and encourages temporal grounding and causal reasoning in video understanding. 

\paragraph{Interaction Scenario $\Rightarrow$ Vision-Guided Interaction Task.} Beyond navigation and query answering, visually impaired users also rely on assistive systems to provide a sequence of actionable instructions—because they cannot visually verify their progress, the model must continually update its guidance as the scene evolves. To capture this requirement, we introduce the \textbf{Vision-Guided Interaction} task, which simulates multi-turn, vision-grounded assistance. Each annotated interaction segment corresponds to a conversational turn in which the model is prompted with the accumulated dialogue history and the visual context up to that moment, and must produce the next instructive response. This setting evaluates the model’s ability to maintain coherent multi-step guidance, adapt to changes in the visual stream, and ground its instructions in the user’s ongoing situation.

Together, these three tasks form a unified evaluation protocol that connects core assistive requirements with concrete computational challenges.  
They jointly cover proactive perception (when to act), reactive reasoning (how to answer), and interactive communication (how to assist)—providing a holistic assessment of MLLM capabilities in real-world assistive scenarios.

\begin{figure}[t]
  \centering
  \includegraphics[width=\linewidth]{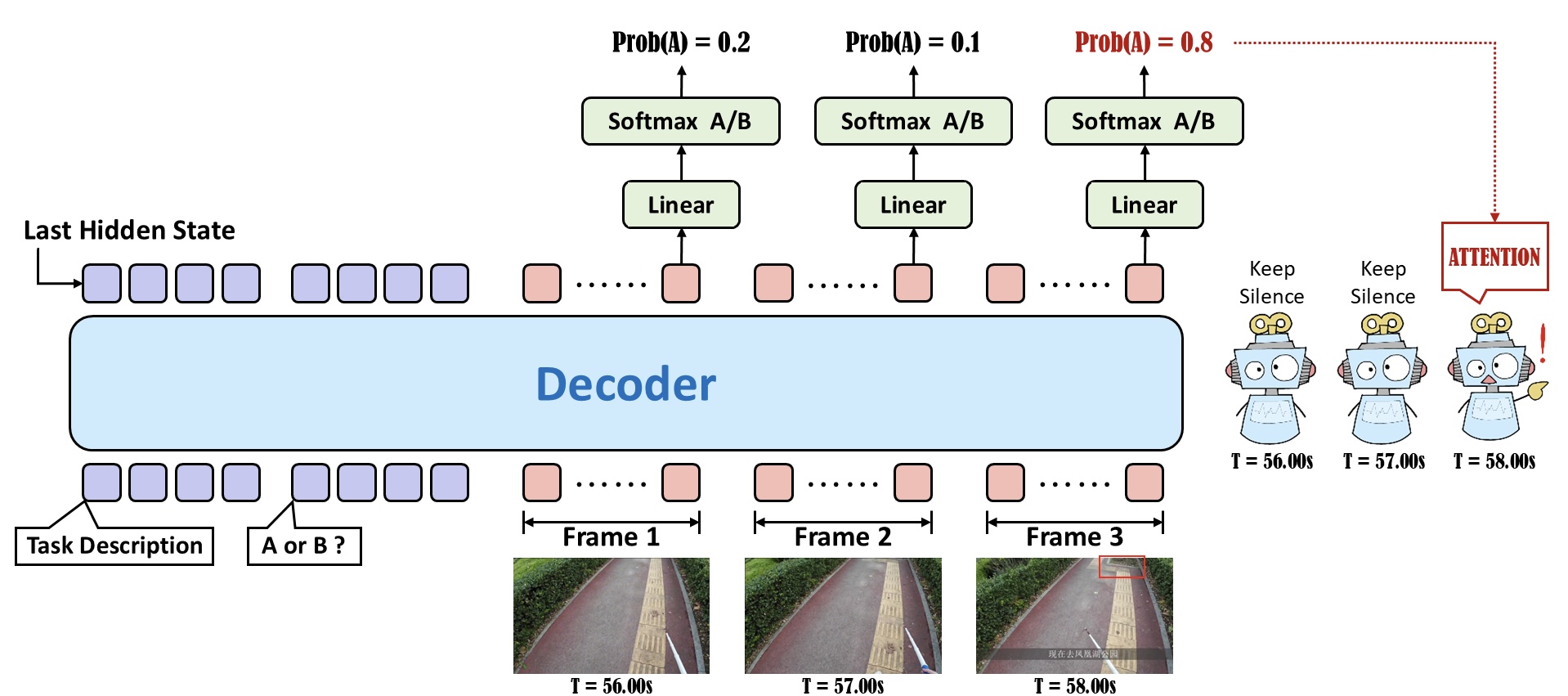}
  \caption{
  Overview of Token-Level Prompt Activation Decoding (TPAD). 
  Given a fixed forced-choice prompt, the MLLM processes a concatenated token sequence consisting of the prompt and video frame representations. 
  For each frame, the hidden state of its final token is projected via the language modeling head to obtain the probability of issuing an alert (option A). 
  TPAD enables efficient, context-aware alert detection for offline models through a single forward pass. 
  In this example, an alert is triggered only at T = 58.00s.}
  \label{fig:tpad}
\end{figure}

\subsection{Token-Level Prompt Activation Decoding}

As discussed above, proactive assistance in navigation requires models to issue timely alerts based on continuous visual input—an ability that most existing MLLMs lack. 
They typically operate in an offline, prompt-driven fashion, processing static images or short clips only when explicitly queried. 
To bridge this gap and make offline MLLMs evaluable in the \textbf{Proactive Reminder} task, we introduce a lightweight adaptation mechanism called \textbf{Token-Level Prompt Activation Decoding (TPAD)}.

\vspace{0.5em}
\noindent\textbf{Core Idea.}
TPAD transforms a pretrained MLLM into a frame-wise proactive detector \textit{without any fine-tuning}. 
It leverages the intrinsic sequential structure of Transformer-based architectures \citep{vaswani2017attention, dosovitskiy2020image} to interpret each frame’s hidden representation as a potential activation signal. 
Given a fixed forced-choice prompt such as “\textit{Should the user be warned? (A) Yes; (B) No}”, TPAD feeds both the textual prompt and all video frames into the model as a single concatenated token sequence. 
Passing this sequence through the MLLM produces the final-layer hidden states 
$H_{\text{last}} \in \mathbb{R}^{L \times d}$, where $L$ is the total number of tokens and $d$ the hidden dimension.

\vspace{0.5em}
\noindent\textbf{Frame-Level Scoring.}
For each video frame $F_k$, we locate its final token index $idx_k^*$ within the concatenated sequence and extract the corresponding hidden state $h_{idx_k^*} \in \mathbb{R}^d$. 
This representation is projected into the vocabulary space through the MLLM’s language modeling head ($W_{\text{LM}} \in \mathbb{R}^{V \times d}$, $b_{\text{LM}} \in \mathbb{R}^V$):
\begin{equation}
z^{(k)} = W_{\text{LM}} h_{idx_k^*} + b_{\text{LM}} .
\label{eq:logits_k_concise}
\end{equation}
From the resulting logits, we isolate those corresponding to the two response tokens (“A” for alert, “B” for no alert) and compute the alert probability for frame $F_k$ as:
\begin{equation}
P(\text{alert} | F_k, S_{\text{prompt}}) = \text{softmax}(z_A^{(k)}, z_B^{(k)})[0] .
\label{eq:prob_k_concise}
\end{equation}
Thus, a single forward pass yields frame-wise alert probabilities across the entire video, enabling efficient evaluation of offline models under streaming conditions.

\vspace{0.5em}
\noindent\textbf{Efficiency and Contextual Awareness.}
Traditional prompting-based methods require one inference per video segment, which scales linearly with sequence length. 
In contrast, TPAD computes token-level activations in one holistic forward pass. 
This design drastically reduces computational cost and, more importantly, allows each frame’s prediction to implicitly condition on all preceding visual context. 
Such contextual awareness proves especially valuable in ambiguous or partially observable scenes, where local frame evidence is insufficient. 
Empirically, we observe that TPAD not only enables offline MLLMs to perform proactive reasoning but also improves stability compared to frame-isolated baselines. 
Detailed comparisons and runtime analyses are presented in the Appendix.

\section{Experiments}
This section presents comprehensive experiments and analyses of VIABench.

\subsection{Evaluation Details}
We evaluate three categories of Multimodal Large Language Models (MLLMs) on VIABench: (1) open-source foundation models, (2) closed-source proprietary models, and (3) online streaming models. 
All three tasks in VIABench---Proactive Reminder, VQA, and Vision-Guided Interaction---are evaluated. Among them, the Proactive Reminder task requires the model to issue outputs proactively in response to the visual stream.

\paragraph{Open-source Foundation MLLMs.}
We evaluate a broad selection of open-source MLLMs, including \textbf{InternVL3.5} \citep{wang2025internvl3},
\textbf{Qwen2.5VL} \citep{bai2025qwen25vltechnicalreport},
\textbf{LLaVA-OneVision-1.5} \citep{an2025llava},
\textbf{LLaVA-Video} \citep{zhang2024video},
\textbf{MiniCPM-V 4.5} \citep{yu2025minicpm},
\textbf{MiniCPM-o 2.6} \citep{yao2024minicpm}, and
\textbf{Kimi-VL} \citep{team2025kimi}.
For the Proactive Reminder task, these models are augmented with TPAD to enable proactive activation. 
To explore settings relevant to low-resource assistive devices, we additionally include smaller variants such as \textbf{InternVL3.5-1B}, \textbf{InternVL3.5-4B}, and \textbf{Qwen2.5VL-3B}.

\paragraph{Proprietary MLLMs.}
We additionally include strong closed-source baselines such as OpenAI’s \textbf{GPT-5} \citep{gpt5}, \textbf{GPT-4o} \citep{gpt4o}, and Google’s \textbf{Gemini-2.5 Pro} \citep{comanici2025gemini}. 
Due to their non-modifiable decoding pipelines, these models are evaluated using a conventional frame-by-frame prompting setup for the Proactive Reminder task, without TPAD.

\paragraph{Online Streaming MLLMs.}
Finally, we evaluate several online multimodal models---\textbf{LiveCC} \citep{chen2025livecc}, 
\textbf{VideoLLM-Online} \citep{chen2024videollm}, and \textbf{StreamingVLM} \citep{xu2025streamingvlm}---which are designed with native streaming and proactive response mechanisms. 
These models inherently align with the requirements of the Proactive Reminder task and are therefore tested directly under their streaming execution modes, while VQA and Vision-Guided Interaction follow the same evaluation protocol as the offline models.

\subsection{Main Results}

\begin{table*}[t]
\centering
\captionsetup{singlelinecheck=false, justification=raggedright}
\resizebox{\textwidth}{!}{%
\begin{tabular}{@{}l|c|cccccccccccccccccccccc|c|c@{}}
\toprule

\multirow{2}{*}{\textbf{Model}} 
& \multirow{2}{*}{\textbf{Overall}}
& \multicolumn{22}{c|}{\textbf{Proactive Reminder}}
& \multirow{2}{*}{\textbf{VQA}}
& \multirow{2}{*}{\textbf{VGI}} \\
 
& & OA & STUD & RSC & CW & DD & COS & CO & SFD & RB & SCUD & BPR & IR & DG & AA & PTL & E & SR & EER & SLUD & CCA & SD & Avg. & \\
\midrule

\multicolumn{24}{c}{\textbf{Open-source Multimodal Models - Offline}} \\
\midrule

InternVL3.5-1B \citep{wang2025internvl3}
& 10.1 & 5.8 & 15.1 & 6.6 & 11.2 & 1.4 & 7.6 & 12.8 & 11.4 & 22.4 & 18.8 & 20.7 & 9.5 & 5.6 & 7.0 & 14.1 & 5.8 & 2.3 & 26.7 & 7.7 & 19.0 & 15.6 &
 9.2 & 25.0 & 16.4 \\

InternVL3.5-4B \citep{wang2025internvl3}
& 18.3 & 13.4 & \textbf{21.5} & 12.3 & 7.3 & \textbf{4.8} & 15.0 & \textbf{28.9} & 22.8 & \textbf{34.3} & 18.3 & \textbf{28.2} & 24.5 & 4.3 & \textbf{17.5} & 9.5 & 13.7 & 12.3 & 31.6 & 16.5 & 46.7 & \textbf{19.2} & 17.7 & 31.1 & 22.6 \\

InternVL3.5-8B \citep{wang2025internvl3}
& \textbf{19.1} & \textbf{13.5} & 20.0 & \textbf{15.0} & 11.7 & 2.4 & 27.5 & 26.4 & \textbf{25.6} & 32.5 & \textbf{32.5} & 19.6 & 27.4 & 5.6 & 17.3 & \textbf{17.0} & 20.6 & \textbf{14.8} & \textbf{37.0} & 14.6 & \textbf{49.7} & 14.0 & \textbf{18.5} & \textbf{39.5} & 20.1 \\

Qwen2.5VL-3B \citep{bai2025qwen25vltechnicalreport}
& 11.4 & 6.1 & 10.0 & 11.6 & 7.3 & 2.0 & 8.0 & 17.5 & 8.6 & 22.1 & 16.5 & 18.8 & 13.7 & 20.7 & 8.9 & 6.3 & 15.6 & 1.1 & 21.5 & 8.7 & 47.0 & 8.4 & 10.1 & 35.2 & 19.2 \\

Qwen2.5VL-7B \citep{bai2025qwen25vltechnicalreport}
& 13.6 & 9.7 & 11.7 & 9.9 & 31.7 & 2.3 & 12.8 & 18.9 & 14.3 & 17.3 & 27.5 & 9.0 & 20.2 & 3.1 & 10.9 & 15.3 & \textbf{27.0} & 5.7 & 22.4 & 7.8 & 12.9 & 17.6 & 11.8 & 38.9 & \textbf{27.1} \\

LLaVA-Video-7B \citep{zhang2024video}
& 12.0 & 7.7 & 15.5 & 6.8 & 25.6 & 2.5 & 16.4 & 14.1 & 9.4 & 23.8 & 12.8 & 12.4 & 11.7 & \textbf{21.4} & 6.9 & 13.4 & 16.0 & 2.3 & 10.8 & 10.9 & 39.0 & 13.2 & 10.6 & 39.0 & 20.1 \\

LLaVA-OV-1.5-8B \citep{an2025llava}
& 12.1 & 7.6 & 18.2 & 8.5 & 9.0 & 3.6 & 15.0 & 11.1 & 7.7 & 26.5 & 21.0 & 9.6 & 18.6 & 0.8 & 11.4 & 14.9 & 20.1 & 8.2 & 17.3 & \textbf{18.1} & 25.2 & 12.4 & 10.8 & 37.3 & 19.3 \\

Kimi-VL-A3B \citep{team2025kimi}
& 15.9 & 10.8 & 19.7 & 7.8 & \textbf{41.7} & 3.0 & \textbf{32.7} & 28.0 & 20.1 & 23.1 & 28.2 & 22.2 & \textbf{32.0} & 8.1 & 12.6 & 10.4 & 23.9 & 3.2 & 33.3 & 16.5 & 32.3 & 9.2 & 15.4 & 33.3 & 16.2 \\

MiniCPM-V 4.5 \citep{yu2025minicpm}
& 15.9 & 11.2 & 13.7 & 11.2 & 19.0 & 3.4 & 21.3 & 24.9 & 17.7 & 27.6 & 23.5 & 18.3 & 22.2 & 12.7 & 14.4 & 9.5 & 13.4 & 7.7 & 25.4 & 16.2 & 47.4 & 8.4 & 15.0 & 29.6 & 22.1 \\

MiniCPM-o 2.6 \citep{yao2024minicpm}
& 8.2 & 4.3 & 6.7 & 4.4 & 12.0 & 0.2 & 8.6 & 13.5 & 6.0 & 13.4 & 9.2 & 7.9 & 10.8 & 0.6 & 6.9 & 6.7 & 8.2 & 5.2 & 11.7 & 5.2 & 24.8 & 8.0 & 6.5 & 28.4 & 23.2 \\

\midrule
\multicolumn{24}{c}{\textbf{Open-source Multimodal Models - Online}} \\
\midrule

VideoLLM-online-8B \citep{chen2024videollm}
& 1.8 & 0.1 & 0.1 & 0.0 & 0.0 & 0.0 & 0.0 & 1.2 & 0.0 & 0.8 & 0.0 & 0.3 & 0.7 & 0.0 & 0.1 & 0.0 & 0.0 & 0.0 & 0.5 & 0.6 & 0.0 & 0.0 & 0.2 & 28.6 & 13.9 \\

LiveCC \citep{chen2025livecc}
& \textbf{6.7} & \textbf{2.9} & \textbf{5.6} & \textbf{2.4} & \textbf{0.3} & \textbf{1.5} & \textbf{3.0} & \textbf{3.5} & \textbf{2.5} & \textbf{2.1} & \textbf{6.5} & \textbf{2.9} & \textbf{6.3} & \textbf{3.7} & \textbf{1.7} & \textbf{4.6} & \textbf{1.6} & 2.1 & \textbf{4.2} & \textbf{2.1} & \textbf{5.1} & \textbf{2.4} & \textbf{3.1} & \textbf{37.1} & \textbf{22.0} \\

StreamingVLM \citep{xu2025streamingvlm}
& - & 0.5 & 0.5 & 0.4 & 0.0 & 0.3 & 1.5 & 0.6 & 0.3 & 1.5 & 1.9 & 1.1 & 0.2 & 1.2 & 0.6 & 2.6 & 0.6 & \textbf{2.7} & 0.8 & 0.5 & 1.0 & 0.8 & 0.6 & - & - \\

\midrule
\multicolumn{24}{c}{\textbf{Proprietary Multimodal Models - Offline}} \\
\midrule

GPT-5-0806 \citep{gpt5}
& \textbf{28.8} & \textbf{18.9} & \textbf{26.2} & 42.5 & 6.0 & 16.0 & 47.5 & 25.7 & \textbf{38.0} & 10.0 & 41.7 & 26.4 & 10.0 & 8.2 & \textbf{19.4} & 8.7 & 24.0 & 20.0 & 28.3 & \textbf{40.0} & 42.9 & 15.0 & \textbf{21.9} & \textbf{62.9} & \textbf{51.7} \\

GPT-4o-0806 \citep{gpt4o}
& 25.2 & 16.6 & 14.5 & 15.0 & \textbf{22.0} & 4.4 & 41.3 & 24.3 & 26.4 & 0.0 & \textbf{55.0} & 26.4 & \textbf{12.5} & 8.2 & 14.7 & 13.3 & 20.0 & \textbf{38.3} & \textbf{31.7} & 20.0 & 52.1 & 18.3 & 19.2 & 56.6 & 43.5 \\

Gemini-2.5 Pro \citep{comanici2025gemini}
& 27.1 & 16.6 & 22.4 & \textbf{52.5} & 4.0 & \textbf{21.0} & \textbf{48.8} & \textbf{27.1} & 18.8 & \textbf{29.0} & 15.0 & \textbf{40.9} & 10.0 & \textbf{9.1} & 17.8 & \textbf{28.7} & \textbf{30.0} & 21.7 & 24.2 & 20.0 & \textbf{63.6} & \textbf{18.3} & 21.3 & 58.8 & 43.5 \\

\end{tabular}}
\caption{
\textbf{Detailed evaluation results on VIABench.}  
We report results for three core tasks: \textbf{Proactive Reminder}, \textbf{Visual Question Answering (VQA)}, and \textbf{Vision-Guided Interaction (VGI)}. For the Proactive Reminder task, results for all sub-tasks are included; detailed definitions of each sub-task can be found in the Appendix.
The abbreviations denote: OA (Obstacle Alert), STUD (Stairs Up/Down), RSC (Road Surface Condition), CW (Crosswalk), DD (Direction Deviation), COS (Crossing to Other Side), CO (Camera Obstruction), SFD (Service Facility Detection), RB (Road Branch), SCUD (Staircase Up/Down), BPR (Blind Path Recognition), IR (Intersection Recognition), DG (Direction Guidance), AA (Active Avoidance), PTL (Pedestrian Traffic Light), E (Escalator), SR (Sign Recognition), EER (Exit/Entry Recognition), SLUD (Slope Up/Down), CCA (Camera Content Anomaly), and SD (Sidewalk Detection).
}

\label{tab:results}
\end{table*}

Table~\ref{tab:results} reports the detailed sub-task scores for the \textit{Proactive Reminder} task, along with the overall results for VQA and Vision-Guided Interaction.
For the VQA and Vision-Guided Interaction tasks, we measure the quality of model responses using a strong external evaluator---GPT\,-5 \citep{gpt5}---to assess the correctness of each response against the ground-truth annotation. 

For the Proactive Reminder task, the score is computed as a weighted combination of recall and response correctness: models must both trigger reminders within the ground-truth temporal intervals and produce accurate, context-aware guidance grounded in the visual content and the provided instruction. 
Full metric definitions and implementation details are provided in the Appendix. Our analysis of these evaluations reveals several key findings:


\textbf{Nascent Capabilities in Proactive Assistance and Interaction:} 
Experimental results show that current MLLMs exhibit only early-stage competence in real-world visual assistance. Even the strongest model, GPT-5 \citep{gpt5}, reaches an average score of just 28.8, revealing substantial performance gaps.  Notably, performance on \textbf{Proactive Reminder} and \textbf{Vision-Guided Interaction} tasks remains notably below that observed in more established benchmarks such as Visual Question Answering (VQA). Specifically, in the \textbf{Proactive Reminder} task, models struggle to meet the dynamic and nuanced requirements of issuing timely alerts. They frequently fail to anticipate navigation-critical events or generate guidance aligned with ground-truth intervals, reflecting difficulties in delivering accurate, context-aware assistance for visually impaired users in real-world navigation scenarios. For the \textbf{Vision-Guided Interaction} task, models often underperform when providing multi-turn, vision-grounded guidance. Two factors contribute to this: first, first-person videos captured by visually impaired users are often low-resolution, blurry, or captured from unusual viewpoints, which complicates accurate scene understanding. Second, even when visual content is correctly processed, models may misinterpret the user’s needs. For example, when asked to help locate an object, a model might assume full vision and provide a generic instruction such as “pick it up” rather than offering spatially specific guidance for a visually impaired user.

\noindent \textbf{Limitations of current proactive-response models:} Our study reveals that existing proactive-response models perform extremely poorly on the Proactive Reminder task. Our analysis suggests two main reasons for this limitation. First, although current streaming designs enable models to generate proactive outputs, their ability to comprehend complex task scenarios remains very limited. Second, the training data for these streaming models is typically narrow in scope, often consisting of dense video captions or automatically generated video narrations, such as Ego4D-Narration \citep{grauman2022ego4d} and SoccerNet-Caption \citep{mkhallati2023soccernet}. While this data allows models to acquire a basic capacity for proactive output, it provides insufficient exposure to the diverse and complex instructions required for real-world guidance, resulting in poor understanding and execution of more sophisticated tasks. We present example proactive reminder responses of these models in the Appendix.

\section{Conclusion and Future Work}

In this work, we have introduced VIABench, a benchmark that pioneers the deep integration of online video understanding for evaluating Multimodal Large Language Models (MLLMs) as assistive tools for visually impaired individuals (VIIs). VIABench features critical tasks such as Proactive Reminder, VQA, and Vision-Guided Interaction. To facilitate robust online evaluation, particularly for proactive alerting, we proposed the efficient Token-Level Prompt Activation Decoding (TPAD). Our experiments underscore the nascent proactive capabilities of current MLLMs, especially when contrasted with their VQA performance, and reveal the limitations of current proactive-response models.

\noindent \textbf{Future Work.}  
Future directions can be organized into three complementary avenues. First, there is a clear need to explore more effective approaches to realize the proactive-response capabilities of MLLMs. Current models often struggle to anticipate navigation-critical events and generate timely guidance, highlighting the necessity for improved model architectures that can reliably deliver context-aware, real-time responses in complex scenarios. Second, existing models are predominantly trained on dense video captions or automatically generated video narrations, which provide only limited exposure to the types of instructions and semantic complexity encountered in real-world blind assistance. Future work should incorporate richer and more diverse video-text datasets, encompassing multiple instruction types, complex interactions, and a wider range of real-world scenarios, to better equip models for proactive reasoning. Third, there is a pressing need to develop models specifically optimized for visually impaired users. Many current MLLMs fail to account for the unique perspective and intentions of blind users, often treating them as sighted individuals, and struggle with first-person videos that are low-quality, blurry, or captured from unconventional viewpoints. Models trained and fine-tuned with blind-centered data could improve alignment with user needs, enhancing the accuracy, safety, and overall utility of assistive guidance.  

Together, these directions—advancing proactive-response mechanisms, expanding and enriching training data, and tailoring models for assistive scenarios—will help drive the development of MLLM-based assistive technologies that are both more capable and dependable, ultimately improving autonomy and safety for visually impaired individuals.

{
    \small
    \bibliographystyle{ieeenat_fullname}
    \bibliography{main}
}

\clearpage
\setcounter{page}{1}
\maketitlesupplementary


\section{More Details of Evaluation}

\subsection{Evaluation Settings}

\noindent\textbf{Proactive Reminder.}
Offline models are evaluated using TPAD, while online models are tested with their default inference interfaces. For closed-source models, we use single-frame inference according to their API constraints.

\noindent\textbf{VQA.}
For each question, we extract the video segment from the beginning of the clip up to the question timestamp and feed it to the model, using the model's default frame sampling and inference configuration.

\noindent\textbf{Vision-Guided Interaction.}
We adopt a multi-turn VQA-style evaluation protocol due to the interactive nature of this task. In each turn, ground-truth responses and timestamps from all previous turns are appended to form the dialogue history, which is supplied to the model along with the current visual input.

\noindent\textbf{Frame Sampling.}
For TPAD and GPT evaluations, video frames are sampled at 1~FPS. All other evaluations use each model’s default input frame rate and inference configuration.

\subsection{Primary Metric for Proactive Reminders}

For each annotated event in the dataset, we define a ground truth interval \([t_s, t_e]\) corresponding to the valid time window during which a model should issue a task-specific prompt. A model is considered to have successfully detected an event if it outputs a matching task prediction and an associated prompt at any time \(t_m\) such that:

\[
t_s \leq t_m \leq t_e
\]

Let \(\mathcal{A}\) denote the set of all annotated events in the dataset, and let \(\mathcal{D}_{\text{model}}\) denote the subset of these events that are correctly detected by the model under this criterion. We define the \textbf{Proactive Detection Rate (PDR)} as:

\begin{equation}
\text{PDR} = \frac{|\mathcal{D}_{\text{model}}|}{|\mathcal{A}|}
\end{equation}

This metric captures the proportion of opportunities where the model successfully performs a task-relevant intervention within the allowed temporal window.

\subsection{Description Accuracy with GPT-5}

While temporal correctness is crucial, a complete evaluation must also assess the semantic quality of the natural language descriptions provided by the model. To this end, we incorporate a secondary scoring mechanism using GPT-5 to compare the generated prompt \(y_{\text{pred}}\) and the ground truth description \(y_{\text{gt}}\) for each matched event.

We prompt GPT-5 to return a similarity score \(s(y_{\text{pred}}, y_{\text{gt}}) \in [0, 1]\), penalizing mismatches in obstacle location, type, or behavioral instruction. The average of these scores over all matched events provides an auxiliary metric termed \textbf{Mean Prompt Similarity (MPS)}:

\begin{equation}
\text{MPS} = \frac{1}{|\mathcal{A}|} \sum_{(y_{\text{pred}}, y_{\text{gt}}) \in \mathcal{D}_{\text{model}}} s(y_{\text{pred}}, y_{\text{gt}})
\end{equation}

\subsection{Metric for Visual Question Answering and Vision-Guided Interaction}

For the VQA and VGI task, we directly apply GPT-5 to assess semantic similarity between the model-generated answer and the ground truth. The model receives both texts and returns a similarity score in the range \([0,1]\). The mean of these scores across the test set constitutes the final \textbf{VQA/VGI Accuracy Score (VAS)}:

\begin{equation}
\text{VAS} = \frac{1}{N} \sum_{i=1}^{N} s\left(y_{\text{pred}}^{(i)}, y_{\text{gt}}^{(i)}\right)
\end{equation}

where \(N\) is the total number of VQA(or VGI) examples and \(s(\cdot, \cdot)\) is the GPT-5 similarity scoring function.

\section{Additional Results}

\subsection{Proactive Reminder}

In the \textbf{Proactive Reminder} task, the problem is inherently a two-stage end-to-end process.
\textbf{Stage~1} is formulated as a frame-level binary classification task: given a sequence of video frames, the model predicts for each frame whether it should trigger a reminder. A prediction is counted as a true positive (TP) when the model marks a frame as a trigger and this frame lies within the ground-truth trigger interval. Stage~1 performance is therefore measured using recall-based metrics. The detailed recall results for all sub-tasks are reported in Table~\ref{tab:pr_stage1_recall}.

\textbf{Stage~2} then takes the true-positive frames retrieved by Stage~1 and generates the final reminder output. However, because each model independently selects trigger frames during Stage~1, the video segments fed into Stage~2 differ across models, making it difficult to isolate the second-stage generation capability.

To more directly compare Stage~2 generation quality, we conduct an ablation study under a unified sliding-window setup. For each task, we sample up to 100 examples and evaluate only the second stage under this controlled condition. Concretely, for every example, we extract a fixed 32-frame window immediately preceding the ground-truth trigger time and feed this identical window to every model's Stage~2 module. This ensures that all models observe exactly the same visual context, enabling a clean and fair comparison of Stage~2 generation quality. The results of this ablation are presented in Table~\ref{tab:pr_stage2_generation}.

\noindent\textbf{Stage-1 Retrieval Performance.} Stage-1 retrieval performance emerges as the key bottleneck in the Proactive Reminder task. Across all offline models, recall remains below 45\%, indicating that current systems fail to detect more than half of the reminder-triggering intervals. Such low retrieval fidelity places a strict upper bound on end-to-end effectiveness, since missing the trigger frame makes successful generation inherently impossible. Although LLaVA-OneVision-1.5 \citep{an2025llava} attains the highest recall among offline models, its limited Stage-2 generation quality prevents it from translating this advantage into strong overall performance, underscoring the need for balanced improvements across both stages. While larger models show a gradual upward trend, their ability to localize anomalous or reminder-relevant events in videos remains fundamentally limited.

In contrast, several online models exhibit seemingly strong retrieval performance, with LiveCC \citep{chen2025livecc} achieving a recall as high as 75.8\%. However, this high score is largely a byproduct of dense, high-frequency outputs that resemble continuous video captioning rather than true task understanding. Instead of identifying the correct reminder-triggering moments, these models tend to narrate whatever appears in the scene, resulting in low-quality and often irrelevant predictions despite their frequent responses. Some detailed example responses are provided in Section~\ref{sec:online_examples}.

\begin{table*}[t]
\centering
\captionsetup{singlelinecheck=false, justification=raggedright}
\resizebox{\textwidth}{!}{%
\begin{tabular}{@{}l|ccccccccccccccccccccc|c@{}}
\toprule

\multirow{2}{*}{\textbf{Model}} 
& \multicolumn{21}{c|}{\textbf{Proactive Reminder}}
& \multirow{2}{*}{\textbf{Avg.}} \\
 
& OA & STUD & RSC & CW & DD & COS & CO & SFD & RB & SCUD & BPR & IR & DG & AA & PTL & E & SR & EER & SLUD & CCA & SD \\
\midrule

\multicolumn{23}{c}{\textbf{Open-source Multimodal Models - Offline}} \\
\midrule

InternVL3.5-1B \citep{wang2025internvl3}
& 26.6 & 36.6 & 31.4 & 51.2 & 46.7 & 36.3 & 53.6 & 31.8 & 37.8 & 42.4 & 29.4 & 43.1 & 47.4 & 38.4 & 41.2 & 45.1 & 20.5 & 42.3 & 29.9 & 41.1 & 52.0 & 32.5 \\

InternVL3.5-4B \citep{wang2025internvl3}
& 35.6 & 44.4 & 40.7 & 46.3 & 43.9 & 44.0 & 67.7 & 41.5 & 43.8 & 45.9 & 44.1 & 47.0 & 49.3 & 47.3 & 44.7 & 36.3 & 31.8 & 46.4 & 43.0 & 74.9 & 60.0 & 41.5 \\

InternVL3.5-8B \citep{wang2025internvl3}
& 37.0 & 43.3 & 45.5 & 48.8 & 44.9 & 44.4 & 68.2 & 47.3 & 42.6 & 52.9 & 40.1 & 54.7 & 46.0 & 48.0 & 44.7 & 45.1 & 45.5 & 52.3 & 49.5 & 72.2 & 40.0 & 43.0 \\

Qwen2.5VL-3B \citep{bai2025qwen25vltechnicalreport}
& 26.4 & 27.6 & 31.1 & 29.3 & 36.4 & 29.0 & 61.0 & 25.6 & 27.8 & 36.5 & 34.5 & 29.3 & 40.0 & 31.9 & 31.5 & 33.0 & 36.4 & 38.3 & 29.0 & 71.3 & 20.0 & 30.6 \\

Qwen2.5VL-7B \citep{bai2025qwen25vltechnicalreport}
& 29.8 & 35.3 & 34.9 & 58.5 & 37.4 & 37.9 & 59.6 & 35.2 & 30.6 & 60.6 & 31.1 & 46.4 & 36.7 & 35.0 & 45.5 & 67.0 & 43.2 & 54.5 & 36.4 & 66.3 & 44.0 & 35.5 \\


LLaVA-OV-1.5-8B \citep{an2025llava}
& 39.6 & 43.8 & 47.8 & 63.4 & 46.7 & 45.2 & 66.0 & 52.0 & 44.4 & 55.3 & 42.4 & 50.3 & 53.0 & 48.6 & 47.1 & 53.8 & 52.3 & 52.1 & 44.9 & 66.6 & 48.0 & 44.9 \\

Kimi-VL-A3B \citep{team2025kimi}
& 33.4 & 36.4 & 37.5 & 73.2 & 36.4 & 53.6 & 63.3 & 48.4 & 33.6 & 61.2 & 33.9 & 56.9 & 36.7 & 40.1 & 52.5 & 64.8 & 38.6 & 61.0 & 44.9 & 48.2 & 36.0 & 39.4 \\

MiniCPM-V 4.5 \citep{yu2025minicpm}
& 35.2 & 42.2 & 39.7 & 56.1 & 44.9 & 36.3 & 65.0 & 40.1 & 35.6 & 37.6 & 46.9 & 43.6 & 46.0 & 43.6 & 36.2 & 34.1 & 34.1 & 53.2 & 45.8 & 74.3 & 40.0 & 40.1 \\

MiniCPM-o 2.6 \citep{yao2024minicpm}
& 16.6 & 17.4 & 24.0 & 26.8 & 10.3 & 19.0 & 38.7 & 14.8 & 17.0 & 22.9 & 15.8 & 21.0 & 20.5 & 20.8 & 19.5 & 23.1 & 20.5 & 23.3 & 19.6 & 56.8 & 16.0 & 19.5 \\

\midrule
\multicolumn{23}{c}{\textbf{Open-source Multimodal Models - Online}} \\
\midrule

VideoLLM-online-8B \citep{chen2024videollm}
& 3.4 & 0.9 & 5.5 & 0.0 & 0.0 & 0.0 & 16.9 & 2.0 & 1.2 & 0.0 & 5.1 & 2.2 & 7.4 & 5.4 & 4.7 & 0.0 & 2.3 & 6.8 & 1.9 & 32.7 & 0.0 & 4.6\\

LiveCC \citep{chen2025livecc}
& 71.7 & 80.9 & 82.7 & 81.8 & 85.1 & 86.9 & 90.7 & 74.8 & 77.0 & 84.0 & 79.5 & 80.5 & 79.8 & 80.0 & 65.1 & 77.9 & 69.7 & 84.8 & 81.6 & 75.5 & 68.0 & 75.8\\

StreamingVLM \citep{xu2025streamingvlm}
& 65.9 & 76.7 & 68.8 & 75.8 & 80.6 & 81.8 & 78.0 & 65.9 & 75.9 & 68.9 & 75.8 & 73.2 & 75.0 & 71.0 & 74.3 & 57.4 & 60.6 & 77.5 & 71.1 & 70.6 & 72.0 & 69.2 \\





\end{tabular}}
\caption{
\textbf{Detailed recall performance on VIABench (Proactive Reminder — Stage 1).} 
Results are reported for all sub-tasks.
}
\label{tab:pr_stage1_recall}
\end{table*}

\noindent\textbf{Stage-2 Generation Quality}. Under the unified sliding-window setting, all models observe the same 32-frame visual context, enabling a clean evaluation of their generation capabilities independent of retrieval variance. Among the offline open-source models, InternVL3.5-8B ~\citep{wang2025internvl3} achieves the strongest overall performance, followed by InternVL3.5-4B ~\citep{wang2025internvl3} and Kimi-VL-A3B ~\citep{team2025kimi}. This ordering reflects the benefits of increased model capacity as well as the advantages conferred by stronger instruction-tuned language generation.

Across sub-tasks, the weakest performance is observed on \textbf{Direction Deviation (DD)}. This task requires continuously tracking the user’s walking direction and issuing timely alerts when the trajectory shifts toward a potentially dangerous orientation. The consistently low scores indicate that current models struggle with direction-sensitive tasks that demand fine-grained trajectory monitoring and short-term motion reasoning. In contrast, models perform noticeably better on tasks involving larger and more easily distinguishable visual targets, such as \textbf{Exit/Entry Recognition (EER)}, where the system only needs to identify the presence and state of nearby doors. The disparity between DD and EER highlights a clear limitation: precise directional reasoning and dynamic path-following remain difficult for existing multimodal systems.

Taken together, these observations show that, even after eliminating Stage~1 variability, high-quality proactive reminder generation remains challenging across all models. Both within and across architecture families, language modeling capacity continues to play a decisive role in determining Stage~2 performance.

\begin{table*}[t]
\centering
\captionsetup{singlelinecheck=false, justification=raggedright}
\resizebox{\textwidth}{!}{%
\begin{tabular}{@{}l|ccccccccccccccccccccc|c@{}}
\toprule

\multirow{2}{*}{\textbf{Model}} 
& \multicolumn{21}{c|}{\textbf{Proactive Reminder}}
& \multirow{2}{*}{\textbf{Avg.}} \\
 
& OA & STUD & RSC & CW & DD & COS & CO & SFD & RB & SCUD & BPR & IR & DG & AA & PTL & E & SR & EER & SLUD & CCA & SD \\
\midrule

\multicolumn{23}{c}{\textbf{Open-source Multimodal Models - Offline}} \\
\midrule

InternVL3.5-1B \citep{wang2025internvl3}
& 25.2 & 38.3 & 19.6 & 20.2 & 4.3 & 16.6 & 20.1 & 37.6 & 58.3 & 39.3 & 63.1 & 16.5 & 27.6 & 10.2 & 47.8 & 10.2 & 0.9 & 61.2 & 24.6 & 42.5 & 30.8 & 30.4 \\

InternVL3.5-4B \citep{wang2025internvl3}
& 28.5 & 54.0 & 29.8 & 16.1 & 9.5 & 31.7 & 49.1 & 51.8 & 77.6 & 50.6 & 59.2 & 51.8 & 11.8 & 31.8 & 23.6 & 33.6 & 34.3 & 68.2 & 29.4 & 69.5 & 27.2 & 41.4 \\

InternVL3.5-8B \citep{wang2025internvl3}
& 40.8 & 46.6 & 34.2 & 26.1 & 5.2 & 56.0 & 52.9 & 53.3 & 74.5 & 65.9 & 50.9 & 48.6 & 12.3 & 26.1 & 41.0 & 34.0 & 16.4 & 67.0 & 29.6 & 71.0 & 39.2 & 43.9 \\

Qwen2.5VL-3B \citep{bai2025qwen25vltechnicalreport}
& 23.0 & 38.2 & 30.4 & 39.0 & 6.1 & 30.3 & 32.4 & 33.7 & 78.4 & 40.1 & 53.2 & 50.9 & 42.7 & 25.4 & 13.8 & 41.0 & 4.3 & 51.0 & 30.8 & 52.5 & 47.2 & 36.8 \\

Qwen2.5VL-7B \citep{bai2025qwen25vltechnicalreport}
& 37.8 & 34.1 & 16.0 & 44.9 & 6.4 & 35.6 & 32.9 & 50.1 & 72.4 & 47.6 & 24.6 & 49.0 & 13.5 & 32.8 & 39.1 & 38.1 & 24.6 & 39.0 & 22.2 & 34.3 & 46.4 & 34.9 \\


LLaVA-OV-1.5-8B \citep{an2025llava}
& 13.7 & 40.7 & 19.8 & 13.2 & 6.5 & 32.1 & 19.4 & 12.1 & 77.1 & 36.2 & 21.2 & 33.1 & 0.9 & 17.5 & 42.0 & 33.0 & 5.2 & 37.0 & 31.4 & 38.8 & 21.6 & 27.5 \\

Kimi-VL-A3B \citep{team2025kimi}
& 29.4 & 56.7 & 20.5 & 54.1 & 5.4 & 63.0 & 48.4 & 38.1 & 68.1 & 48.5 & 53.6 & 51.5 & 26.3 & 26.2 & 28.4 & 39.7 & 9.8 & 47.5 & 32.5 & 72.7 & 20.0 & 41.3 \\

MiniCPM-V 4.5 \citep{yu2025minicpm}
& 29.5 & 16.0 & 23.7 & 40.0 & 7.0 & 47.5 & 42.9 & 41.4 & 77.4 & 56.0 & 37.2 & 49.4 & 26.9 & 32.9 & 25.1 & 33.2 & 28.2 & 54.5 & 39.2 & 57.3 & 24.0 & 38.3 \\

MiniCPM-o 2.6 \citep{yao2024minicpm}
& 24.4 & 45.1 & 16.0 & 39.3 & 5.3 & 63.5 & 35.8 & 35.8 & 67.3 & 42.6 & 58.3 & 52.4 & 19.5 & 23.9 & 43.0 & 37.9 & 16.6 & 48.8 & 34.8 & 47.3 & 21.2 & 38.2 \\









\end{tabular}}
\caption{
\textbf{Detailed generation performance on VIABench (Proactive Reminder - Stage 2).}
Results are reported for all sub-tasks. Since proactive-output models cannot control the timing of their outputs and the evaluation cost of closed-source models is prohibitively high, we report results only for offline models.
}
\label{tab:pr_stage2_generation}
\end{table*}

\subsection{The impact of input frames}

As shown in Table~\ref{tab:viabench_frame_ablation}, increasing the number of input frames leads to only marginal improvements on VIABench. This stands in contrast to many video understanding benchmarks where larger temporal context typically brings substantial gains. The key reason is that visual assistance is inherently real-time and situational: in navigation scenarios, for instance, the environment changes rapidly, and visual observations from even a few minutes earlier (e.g., obstacles on the sidewalk) quickly become outdated and offer limited value for the current query.

Instead of relying on long-term temporal dependencies, VIABench is dominated by challenges specific to interactive visual assistance. These challenges include:
\begin{itemize}
    \item \textbf{Proactive visual monitoring}: continuously tracking the visual stream and responding only when necessary, while remaining silent otherwise.
    \item \textbf{Intention-aware question understanding}: accurately interpreting the blind user’s verbal questions and providing genuinely meaningful answers. For example, when asked “Is there a tactile paving nearby?”, the system should not simply reply “Yes”; instead, it must infer the user’s intent—i.e., “If it exists, tell me where; if not, say no.”
    \item \textbf{User-state tracking and adaptive guidance}: monitoring the user’s actions and intentions and offering iterative, context-aware guidance that adapts to the user’s evolving state and environment.
\end{itemize}

These characteristics highlight why visual assistance differs fundamentally from conventional video understanding tasks, and why increasing frame count alone offers limited benefit.

\begin{table*}[t]
\centering
\captionsetup{singlelinecheck=false, justification=raggedright}
\resizebox{\textwidth}{!}{%
\begin{tabular}{@{}l|ccc|ccc|ccc|ccc|ccc@{}}
\toprule

\multirow{2}{*}{\textbf{Model}} 
& \multicolumn{3}{c|}{\textbf{8 frames}} 
& \multicolumn{3}{c|}{\textbf{16 frames}} 
& \multicolumn{3}{c|}{\textbf{32 frames}} 
& \multicolumn{3}{c|}{\textbf{64 frames}} 
& \multicolumn{3}{c}{\textbf{128 frames}} \\
\cmidrule(lr){2-16}
& PR & VQA & VGI 
& PR & VQA & VGI 
& PR & VQA & VGI 
& PR & VQA & VGI 
& PR & VQA & VGI \\

InternVL3.5-8B \citep{wang2025internvl3}
& 44.8 & 34.3 & 21.0 & 44.7 & 36.5 & 21.7 & 43.9 & 35.9 & 20.0 & 43.9 & 35.3 & 21.8 & 42.1 & 38.4 & 22.5 \\

Qwen2.5VL-7B \citep{bai2025qwen25vltechnicalreport}
& 34.8 & 37.8 & 26.6 & 34.6 & 36.9 & 26.8 & 34.9 & 35.6 & 25.1 & 35.6 & 36.5 & 26.7 & 37.5 & 37.2 & 26.3 \\

Kimi-VL-A3B \citep{team2025kimi}
& 43.0 & 33.2 & 17.1 & 40.4 & 32.2 & 17.4 & 41.3 & 33.4 & 16.1 & 41.2 & 32.4 & 16.4 & 41.5 & 30.1 & 16.2 \\

MiniCPM-V 4.5 \citep{yu2025minicpm}
& 42.7 & 32.1 & 21.9 & 41.0 & 29.8 & 21.5 & 38.3 & 28.5 & 21.0 & 37.4 & 30.5 & 20.0 & 40.4 & 30.5 & 21.6 \\

\end{tabular}}
\caption{
\textbf{Effect of input frame count on task performance.}
Performance across \textbf{PR}, \textbf{VQA}, and \textbf{VGI} tasks is reported for different input frame settings (8–128) on VIABench.
}
\label{tab:viabench_frame_ablation}
\end{table*}

\subsection{Inference Latency}

\begin{figure}[H]
  \centering
  \includegraphics[width=0.95\linewidth]{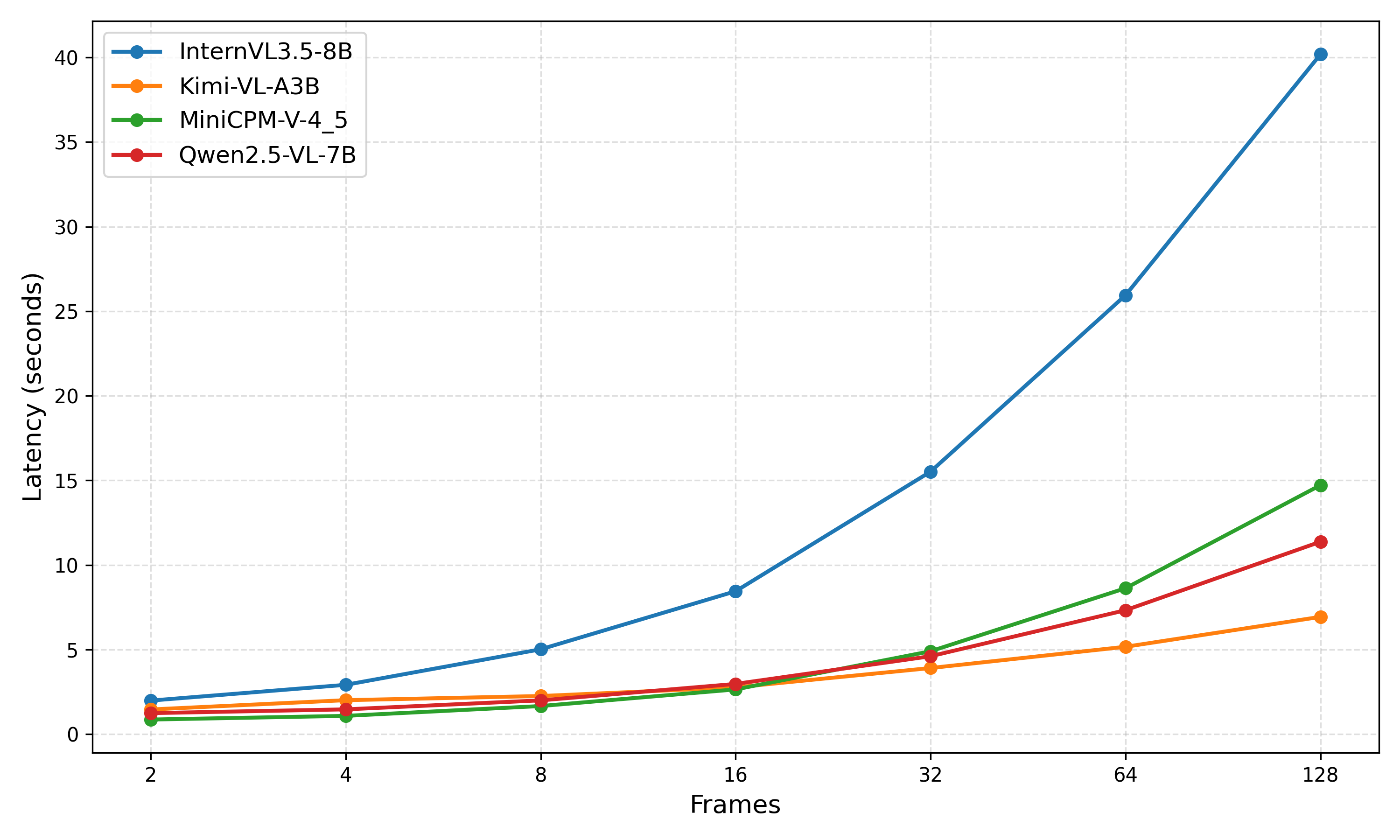}
  \caption{Inference Latency (y-axis) v.s. Frames Number (x-axis). All models are evaluated on a single NVIDIA L20 GPU.}
  \label{fig:latency}
\end{figure}

Current Video-LLMs lack an efficient inference framework to support real-time visual assistance. As shown in Figure ~\ref{fig:latency}, the inference latency of existing models increases exponentially with the number of input frames. When using 32 visual frames, Qwen2.5-VL-7B requires around 4.6 seconds per response, while InternVL-3.5 takes notably longer—around 15.5 seconds. Such latency makes the vision of an always-on, real-time assistance system for blind users still far from practical.

Although online models like VideoLLM-Online can reach 10 FPS, their overall performance remains significantly lower, making them unsuitable for reliable assistance applications.

\subsection{Qualitative Analysis}

\textbf{Hallucination remains one of the major obstacles for Visual Impairment Assistance (VIA) systems.}
Since blind users are unable to perceive visual content, their queries often do not align with the actual visual scene. VIA models are expected to ground user questions in the real environment and provide safe, context-aware guidance. However, when the model is misled by an irrelevant or mismatched query, it may generate hallucinated descriptions that contradict the visual evidence. Such hallucinations can be particularly dangerous in VIA applications, as incorrect scene interpretations may lead to unsafe instructions and potentially severe real-world consequences.

In terms of robustness against hallucination, \textbf{closed-source models generally perform substantially better than open-source alternatives.} Even the most advanced open-source vision-language models—such as InternVL-3.5~\citep{wang2025internvl3} and Qwen2.5-VL~\citep{bai2025qwen25vltechnicalreport}—still exhibit severe hallucinations when user queries conflict with the visual context. In contrast, closed-source models (e.g., GPT-5 \citep{gpt5} and Gemini-2.5 Pro \citep{comanici2025gemini}) are more capable of rejecting or correcting mismatched queries by grounding their responses in the actual video frames.

\begin{figure*}[t]
    \centering
    \includegraphics[width=\linewidth]{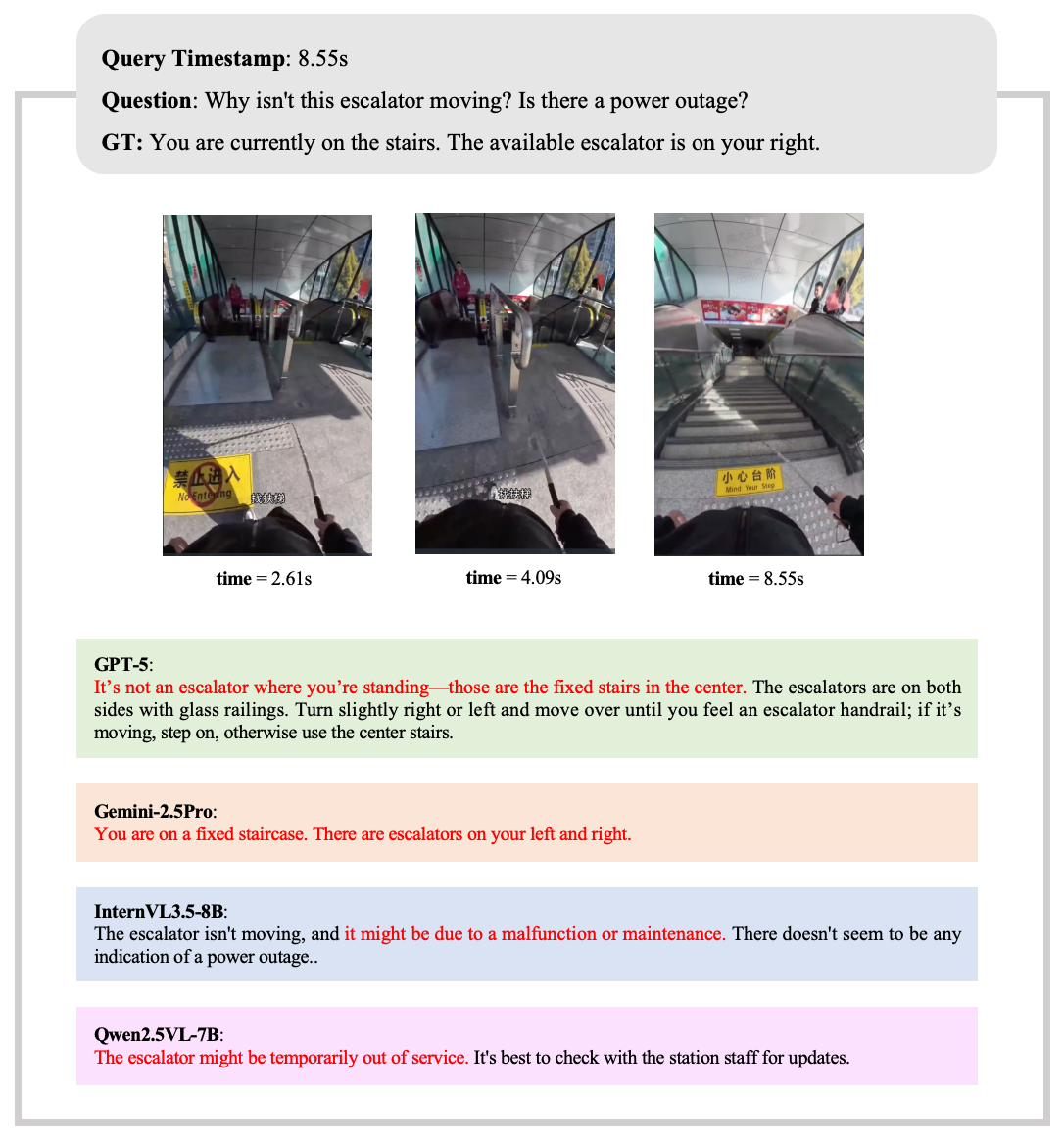}
    \caption{Example of hallucination in open-source multimodal models.}
    \label{fig:halucination}
\end{figure*}

Qualitative examples are shown in Fig.~\ref{fig:halucination}, where open-source models hallucinate nonexistent escalator malfunctions despite the user being on a fixed staircase, whereas closed-source models correctly identify the scene.

\begin{figure*}[t]
    \centering
    \includegraphics[width=\linewidth]{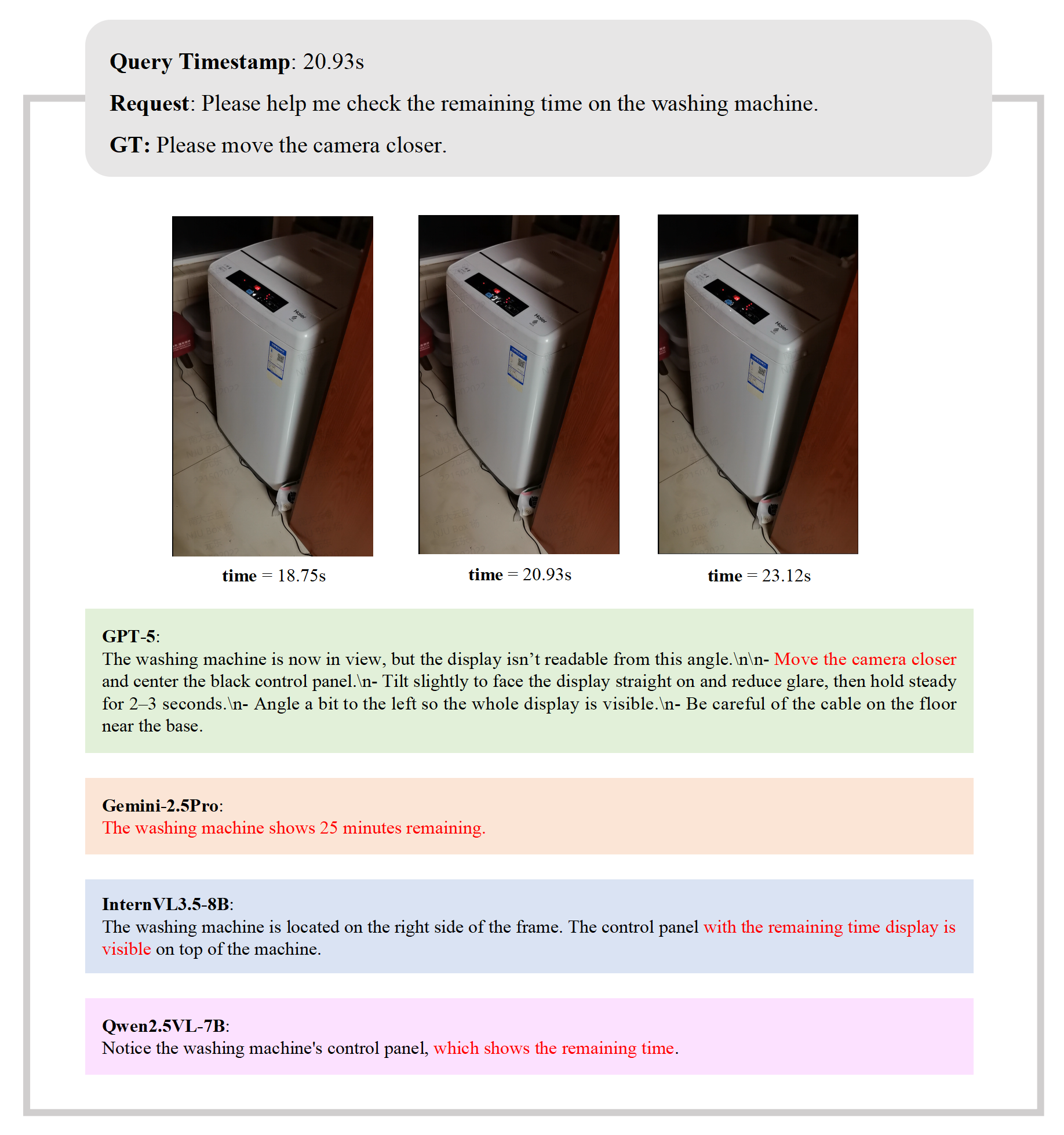}
    \caption{Model responses to the request: “Please help me check the remaining time on the washing machine.” 
While the ground truth, GPT-5 \citep{gpt5}, and Gemini 2.5 Pro \citep{comanici2025gemini} correctly interpret the user as blind—either reading out the remaining time (although Gemini 2.5 Pro \citep{comanici2025gemini} read the wrong number) or providing actionable camera-adjustment guidance—open-source models such as InternVL3.5-8B \citep{wang2025internvl3} and Qwen2.5-VL-7B \citep{bai2025qwen25vltechnicalreport} assume a sighted user. Their responses simply state that the remaining-time display is “visible,” revealing a failure to adapt instructions to non-visual interaction.}

    \label{fig:identity}
\end{figure*}

\noindent\textbf{Understanding the identity of a blind user remains a fundamental challenge for current open-source Video-LLMs.}
In many Vision-Guided Interaction scenarios, the model must provide assistance that explicitly accounts for the fact that the user cannot rely on visual information. As shown in Fig.~\ref{fig:identity}, this requirement is frequently overlooked by open-source models.

When a blind user requests help checking the remaining time on a washing machine, the expected behavior is to acknowledge the user's non-visual needs. Both advanced proprietary models GPT-5 \citep{gpt5} and Gemini 2.5 Pro \citep{comanici2025gemini} interpret the request correctly: they either read out the remaining time or guide the user to adjust the camera to make the display legible.

However, open-source models such as InternVL3.5-8B \citep{wang2025internvl3} and Qwen2.5-VL-7B \citep{bai2025qwen25vltechnicalreport} tend to respond as if assisting a sighted user. Their answers simply point out that the control panel “shows the remaining time” or that it is “visible on top of the machine,” implicitly assuming that the user can see the screen. This mismatch reveals a deeper limitation: the models do not truly internalize the user’s identity as a blind person, leading to instructions that are unusable in non-visual interaction.

\subsection{Efficiency Comparison on a Proactive Reminder Video Segment}

To further substantiate the runtime advantage of TPAD discussed in the Experiment section, we conduct a concrete comparative experiment on a 53-second video clip containing six proactive reminder events. This clip was selected as a representative example where multiple warnings are needed across time, and both TPAD and the conventional MLLM prompting method achieve the same detection accuracy.

\begin{figure}[H]
  \centering
  \includegraphics[width=0.95\linewidth]{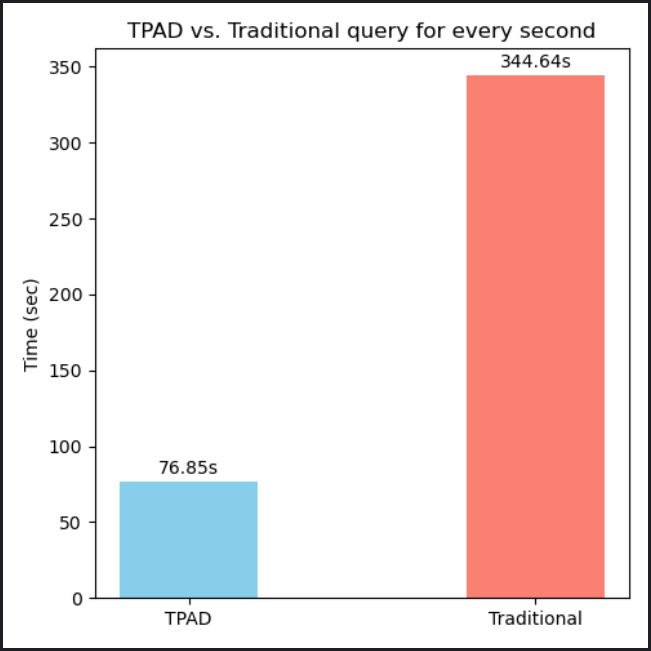}
  \caption{Runtime comparison on a 53s video segment with 6 proactive reminder tasks (same accuracy). TPAD processes the video in 76.85s, while conventional prompting requires 344.64s.}
  \label{fig:C_1}
\end{figure}

\textbf{Baseline Method.} The baseline method follows a frame-wise querying strategy: for each second of the video, a single frame is sampled (i.e., 1 FPS) and passed to the MLLM along with a fixed prompt. This results in 53 separate forward passes. Due to runtime constraints, the query at each timestamp can only include the current frame as input. While in theory it is possible to include the entire 0-to-$t$ video segment as input for better temporal reasoning, doing so for every second would make the memory and computation cost explode. This limitation greatly reduces the model's ability to utilize contextual information over time.

\textbf{TPAD Method.} In contrast, TPAD performs a single forward pass for the entire 53-second video. Instead of processing every second independently, TPAD first performs a lightweight frame selection step, then processes all selected frames \textit{together}, encoding each one in the context of its preceding frames. Specifically, for each selected frame, TPAD constructs an input sequence consisting of the prompt followed by the tokenized representations of frames from the beginning of the video up to that frame. This design not only reduces the number of forward passes to one, but also provides stronger temporal context when computing per-frame alert probabilities.

This comparison reveals that TPAD achieves over a \textbf{4.5$\times$ speedup} compared to traditional MLLM prompting while maintaining identical detection accuracy. The efficiency gain becomes even more significant as the video length and alert density increase, making TPAD particularly suitable for real-time or low-resource deployment scenarios.

\subsection{Example Responses of Online Models}
\label{sec:online_examples}

\begin{figure}[H]
  \centering
  \includegraphics[width=0.95\linewidth]{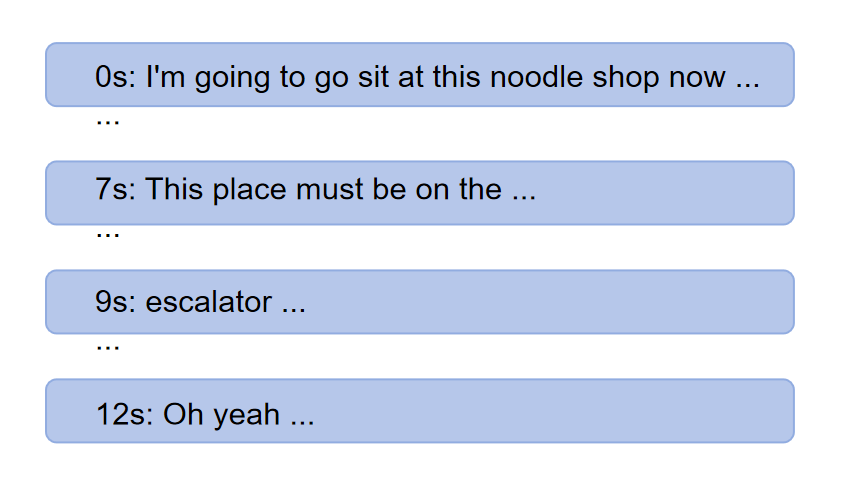}
  \caption{Responses of the LiveCC model on the ``Obstacle Alert'' task. Instead of interpreting the instructed goal, online models tend to default to narrating visible scene content, resulting in reminders that are unrelated to the actual task requirements.}
  \label{fig:online_examples}
\end{figure}

As discussed in the main \textit{Experiments} section, existing online models that are designed to trigger responses autonomously still struggle when deployed in realistic proactive–reminder scenarios. Although these models can in principle decide \emph{when} to speak, they frequently fail to interpret complex, open-ended instructions or sustain coherent guidance over long, first-person video streams. In practice, they either respond too early, miss critical events, or produce captions unrelated to the user's intent. Figure~\ref{fig:online_examples} illustrates several representative outputs: even with clear visual cues, the models offer semantically incorrect reminders, revealing a substantial gap between their nominal proactive capabilities and the demands of real-world assistive navigation.

From these examples, it becomes clear that the high recall of online models mainly reflects their tendency to produce continuous, unsolicited outputs rather than an ability to issue the correct reminders. Although they speak frequently, the content of their responses rarely aligns with the instructed task: most predictions ignore the user's actual intent, misinterpret the scene, or drift into generic descriptions unrelated to navigation. This mismatch highlights a fundamental limitation of current online models—their proactive behavior is driven by output frequency rather than genuine task understanding.

\section{VIABench}

\subsection{Overview}

Our goal is to leverage multimodal large language models (MLLMs) as assistive agents for blind individuals, enabling more independent and safe navigation in real-world environments. We envision deploying these models in smart glasses or smartphones to realize a Jarvis-like real-time visual assistant. Beyond proactively issuing alerts and providing navigation guidance when a user encounters hazardous obstacles, the assistant can also answer situational visual questions and engage in multi-step, interactive guidance to help the user complete tasks in dynamic environments.

\subsection{Timing of Model Alerts in the Proactive Reminder Task}

Model-generated alerts are designed to mimic human attentional timing—they are neither triggered too early (which may be irrelevant or confusing), nor too late (which may be unsafe). Instead, alerts should be issued within a contextually appropriate temporal window that reflects the blind user's spatial relation to the target. Two key paradigms of alert timing are distinguished:

\textbf{(1) Immediate alerts upon event onset:} For tasks marked with \textcolor{red}{\textcircled{a}} in our task list \ref{tab:task-list}, alerts must be issued as soon as the event begins. This includes scenarios such as camera occlusion, unintended walking-direction drift, or unintentionally stepping outside the crosswalk while crossing.

\begin{figure*}[htbp]
    \centering
    \begin{subfigure}{0.49\textwidth}
        \centering
        \includegraphics[width=\linewidth]{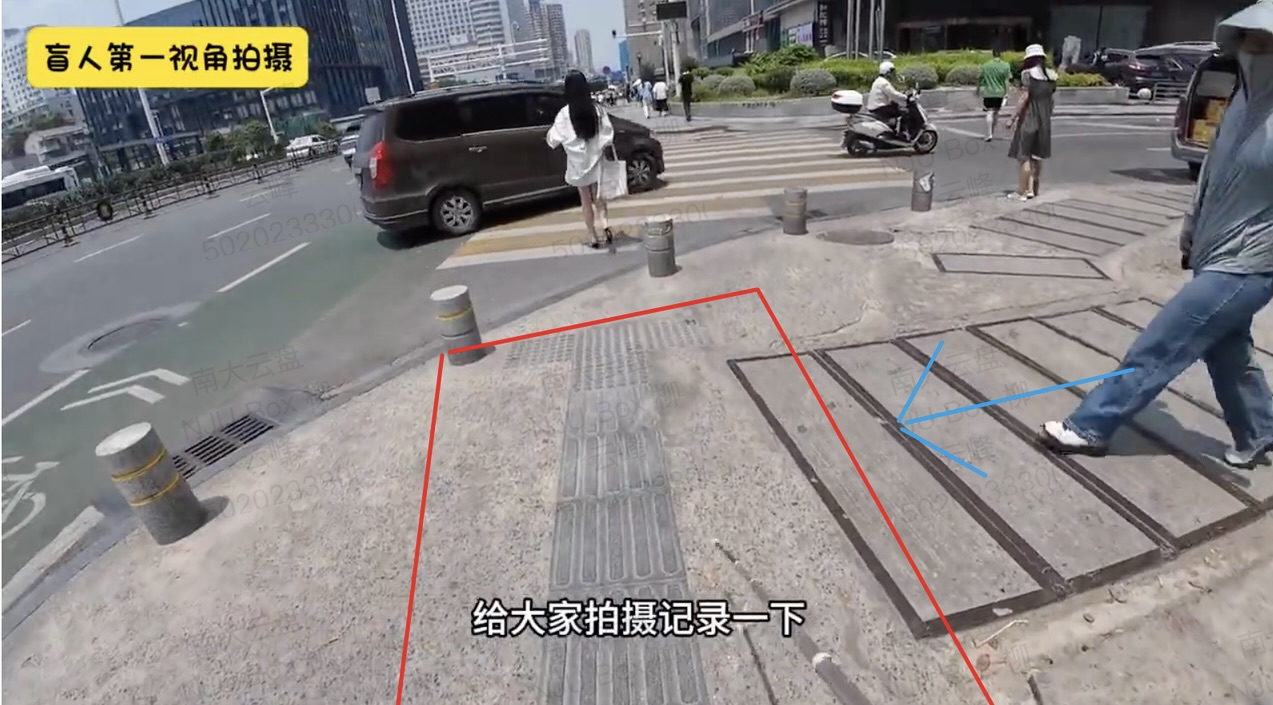}
    \end{subfigure}
    \hfill
    \begin{subfigure}{0.49\textwidth}
        \centering
        \includegraphics[width=\linewidth]{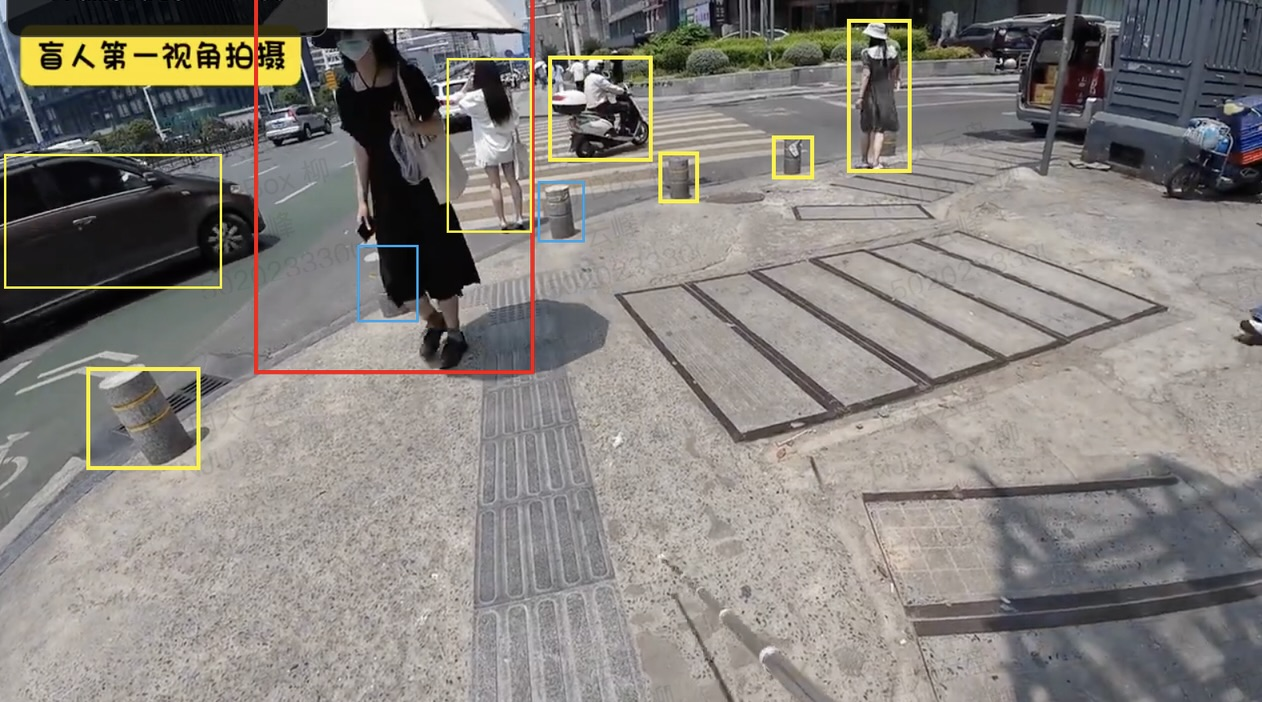}
    \end{subfigure}
    \caption{(\textbf{Left}) Visualization of the defined safe alert range (red bounding box). (\textbf{Right}) Examples of alert-timing categories: targets requiring immediate alerts (red), targets that should be alerted later when the user approaches them (blue), and targets that do not require alerts (yellow).}
    \label{fig:A3_combined}
\end{figure*}

\textbf{(2) Distance-aware alerts:} For other tasks, alerts are only issued when the blind user is within a safe but actionable range from the target object or hazard. Several examples illustrate this principle:

\begin{itemize}
    \item \textbf{Traffic light alerts:} A green light seen from a long distance may become red by the time the user reaches the intersection. Alerts should only be triggered when the user is sufficiently close to make an informed crossing decision.
    \item \textbf{Obstacle avoidance:} As shown on the right side of Figure~\ref{fig:A3_combined}, when walking on the street, immediate alerts are necessary for nearby hazards such as the woman in the black dress, whereas the distant woman in the white shirt should not trigger an alert until it becomes clear that she will affect the user’s path.
\end{itemize}

To formalize this, we define a \textbf{safe alert range}, as illustrated by the red bounding box in left side of Figure~\ref{fig:A3_combined}.

\begin{itemize}
    \item \textbf{Front range:} Approximately 1.5 to 2 meters ahead of the user, which corresponds to 4–6 average walking steps depending on the user’s height.
    \item \textbf{Lateral range:} Around 50 cm on each side of the user, roughly corresponding to one person-width per side. This lateral space is typically probed by the white cane, although probing is asymmetric due to single-hand usage.
\end{itemize}

\textbf{Illustrative categorization example:} In a sample video (right side of Figure~\ref{fig:A3_combined}), we categorize potential alert targets into three groups:

\begin{itemize}
    \item \textbf{Red (alert required):} A woman with an umbrella enters the safe zone at 0.89 seconds and may collide with the user or be hit by the cane—an alert should be triggered.
    \item \textbf{Yellow (no alert):} Pedestrians, cyclists, or vehicles at greater distances or on non-intersecting paths, even if visible.
    \item \textbf{Blue (deferred alert):} Objects close in lateral direction but distant longitudinally (e.g., bollards at the end of a crosswalk) should trigger alerts only when the user approaches.
\end{itemize}

\textbf{Special cases:} In certain edge cases, alerts are still required even when targets fall outside the predefined safe zone, provided their trajectory intersects with the user's future path and may pose a risk (left side of Figure~\ref{fig:A3_combined}).

\subsection{Task List}
As the central component of VIABench, \textbf{Proactive Reminder} encompasses a comprehensive suite of fine-grained tasks designed to evaluate an AI assistant’s capacity for online video understanding and timely, anticipatory feedback. Each sub-task targets a unique navigational scenario commonly encountered by visually impaired individuals in real-world environments. Below, we list all sub-tasks with their abbreviations, full names, and concise descriptions to clarify their function within the benchmark in Table \ref{tab:task-list}. More visual examples can be found in Section \ref{Task Examples}.

\begin{table*}[htbp]
\centering
\begin{tabular}{|c|l|p{8.5cm}|}
\hline
\textbf{Abbr.} & \textbf{Task Name} & \textbf{Description} \\
\hline
OA & Obstacle Alert & Detect and describe nearby obstacles that are close to the user, serving as a reminder rather than prompting active avoidance. \\

\hline
STUD  & Stairs Up/Down & Identify small steps or short stair segments that lead slightly up or down—usually 1 to 3 steps for caution. \\
\hline
RSC   & Road Surface Condition & Describe the surface ahead, including uneven, slippery, or obstructed conditions. \\
\hline
CW    & Crosswalk \textcolor{red}{\textcircled{a}} & Detect crosswalks and help the user stay aligned within them. \\
\hline
DD    & Direction Deviation \textcolor{red}{\textcircled{a}} & Detect user veering off the intended path and guide them back. \\
\hline
COS   & Crossing to Other Side & Assist the user in completing a road crossing to the other side. \\
\hline
CO    & Camera Obstruction \textcolor{red}{\textcircled{a}} & Detect partial camera blockage and alert for adjustment. \\
\hline
SFD   & Service Facility Detection & Identify nearby public service facilities such as benches or trash bins. \\
\hline
RB    & Road Branch & Detect forks or splits in the road and notify the user to prepare. \\
\hline
SCUD  & Staircase Up/Down & Recognize full staircases and indicate their direction (up/down). \\
\hline
BPR   & Blind Path Recognition & Detect and align the user with designated blind paths. \\
\hline
IR    & Intersection Recognition & Identify intersections or junctions and prepare the user for navigation. \\
\hline
DG & Direction Guidance & Notify the user to turn or change direction when the current path ends or is obstructed. \\
\hline
AA    & Active Avoidance & Proactively alert users to hazards requiring path changes. \\
\hline
PTL   & Pedestrian Traffic Light & Recognize pedestrian signal status and advise crossing decisions. \\
\hline
E     & Escalator & Detect escalators and guide safe boarding by indicating their direction. \\
\hline
SR    & Sign Recognition & Detect and read relevant environmental signs for navigation. \\
\hline
EER   & Exit/Entry Recognition & Detect entrances or exits and inform about their accessibility. \\
\hline
SLUD  & Slope Up/Down & Identify slopes and describe their direction to ensure safe traversal. \\
\hline
CCA   & Camera Content Anomaly \textcolor{red}{\textcircled{a}} & Detect anomalies in camera view (e.g., facing sky, overexposure) and prompt correction. \\
\hline
SD   & Sidewalk Detection & Detect the sidewalk in the scene and help the user align themselves with it. \\
\hline
\end{tabular}
\caption{List of Proactive Reminder sub-tasks in VIABench.}
\label{tab:task-list}
\end{table*}

\section{Data Collection Details}

To rigorously and fairly evaluate MLLMs in real-world blind assistance, the quality of data collection and annotation is paramount. VIABench is built upon a carefully designed pipeline that prioritizes the authenticity, richness, and precision of first-person video annotations. This section outlines our data sourcing, annotation methodology, and quality control strategies, with particular focus on principled guidelines, task-specific labeling protocols, and strict quality assurance—all tailored to the unique challenges of real-time, vision-based blind navigation.

\subsection{General Guidelines}

To maintain consistency and high annotation quality across the dataset, we developed a comprehensive set of general guidelines. Prior to large-scale annotation, all annotators received a detailed instruction manual covering the full range of VIABench task types, including definitions, boundary conditions, and illustrative examples of both correct and incorrect annotations. This manual was supplemented by hands-on training videos based on real data, providing step-by-step demonstrations for each task type.

In addition, we organized formal training sessions during the vendor onboarding stage. These sessions played a key role in standardizing task understanding and clarifying common ambiguities identified in early pilot annotations.

Given VIABench’s focus on real-time visual assistance, we enforced two fundamental principles throughout the annotation process:

\begin{itemize}
    \item \textbf{Visual-Only Basis:} All annotations were grounded strictly in visual input. Annotators were instructed to ignore audio cues, ensuring that every label reflected only what a vision-based model could perceive and infer.
    
    \item \textbf{Online Temporal Constraint:} Annotation decisions had to be made without knowledge of future frames. Annotators were not allowed to scrub forward, preserving the real-time nature of assistive inference.
\end{itemize}

By emphasizing these core principles and investing in comprehensive annotator education, we produced annotations that are not only fine-grained and temporally precise but also aligned with the real-world constraints faced by assistive AI systems operating in online settings.

\subsection{Dataset Annotation Submission Guidelines}

To standardize the annotation process, we established a clear submission protocol that all annotators were required to follow. These guidelines were carefully tailored to the structure of each task—especially for those involving temporally localized proactive prompts—and were strictly enforced during both pilot and full-scale annotation phases.

Each annotation represents a single instance of a predefined task type (see Section~\ref{tab:task-list}) and includes the following components:

\begin{itemize}
    \item \textbf{Start and End Timestamps:} Annotators were required to mark the precise temporal interval during which a model should issue a reminder. This interval consists of a \textit{start time} and an \textit{end time}. For example, in the \textit{Obstacle Alert} (OA) task, the \textit{start time} denotes the earliest moment at which a blind user should be warned to avoid a potential collision, while the \textit{end time} represents the last frame where evasive action is still feasible or the obstacle is no longer present.

    \item \textbf{Task Type Selection:} After determining the start–end interval, annotators assigned the corresponding task label according to the benchmark’s task definitions. 

    \item \textbf{Instructional Description:} For each annotated interval, annotators wrote a natural-language reminder that specifies what the assistant should say to the user. For example, in the \textit{Active Avoidance} (AA) task, this reminder typically consists of two parts: (i) an obstacle description that states the obstacle type and its coarse direction or position relative to the user (e.g., "A wall blocks the end of the tactile path ahead"), and (ii) a walking instruction that provides concrete guidance on how to avoid it (e.g., "Please move slightly to the left").
\end{itemize}

Annotators were encouraged to provide reminders that are accurate, informative, and concise, ensuring they remain practical and easy to convey in real-time assistive scenarios.

Crucially, annotations were expected for \textit{every qualifying event} in the video. Annotators were explicitly instructed not to skip any occurrences, ensuring full coverage of assistive opportunities. This standardized schema underpins VIABench’s role as a dependable benchmark for evaluating and training vision-based, real-time assistive models.





\subsection{Common Annotation Issues for Annotators}

Despite extensive training and well-defined guidelines, our audits revealed several recurring annotation errors. These primarily stemmed from misunderstandings about the model’s capabilities or incorrect temporal reasoning. We summarize below two particularly common pitfalls, along with clarifications provided during QA rounds and annotator feedback:

\paragraph{Issue 1: Relying on Audio, Subtitles, or Prior Knowledge.}
Some annotators mistakenly based their decisions on auditory cues, on-screen subtitles, or prior contextual knowledge rather than visual input.

\paragraph{Issue 2: Using Future Events to Justify Present Labels.}
Another common mistake was backward reasoning: annotating a moment based on what the user does afterwards. For instance, if a user turns right a few seconds later, annotators would prematurely mark a “turn right” prompt before the visual scene indicates such a decision.

This misrepresents the model’s role. Unlike watching offline videos—where the viewer can rewind or skip ahead—real-world navigation is continuous and forward-only. Our model simulates this real-time constraint: decisions must be based on the current frame, not on future knowledge.

Moreover, the model does not perform destination-aware navigation like Baidu Maps. It doesn’t know where the user is going; instead, it reacts to immediate visual context to help avoid hazards and navigate safely. Hence, guidance like “turn right” should only be annotated when the current visual scene demands it. For example, if both the front and left are blocked by walls, and only the right is open, then a prompt is appropriate. In contrast, if all paths are open and the user later chooses to turn right, no guidance should be given at that earlier moment.

\section{Limitations}


Despite the strengths of VIABench in promoting real-time, vision-based assistive intelligence, several limitations remain.

First, although we reviewed extensive real-world blind-user videos to identify recurring assistive needs and distilled them into three major task families, the space of real-world visual assistance is fundamentally open-ended. Similar to autonomous driving, assistive navigation must operate in an unconstrained, dynamic environment, where rare events, long-tail edge cases, and highly situational challenges cannot be fully enumerated. Thus, strong performance on VIABench does not guarantee real-world readiness, as even a minor misjudgment or hallucination may pose serious safety risks to blind users.

Additionally, the Vision-Guided Interaction data were collected via role-played simulation by sighted individuals(unlike videos in Proactive Reminder and VQA tasks). While great care was taken to emulate realistic behavior patterns, these simulated recordings inevitably introduce a domain gap relative to actual blind user behaviors. This limits the Interaction part’s current capacity to fully evaluate model generalization across real-world interactive contexts.

In future iterations, we plan to broaden the task coverage and collaborate with blind communities to ethically collect real-world interactive footage. This would help ensure VIABench remains a comprehensive and representative resource for advancing assistive multimodal intelligence.

\section{Prompt Templates}

\subsection{Inference Prompts}

\begin{figure*}[htbp]
  \centering
  \caption{Detailed prompt templates for all Proactive Reminder sub-tasks.}
  \includegraphics[width=\linewidth]{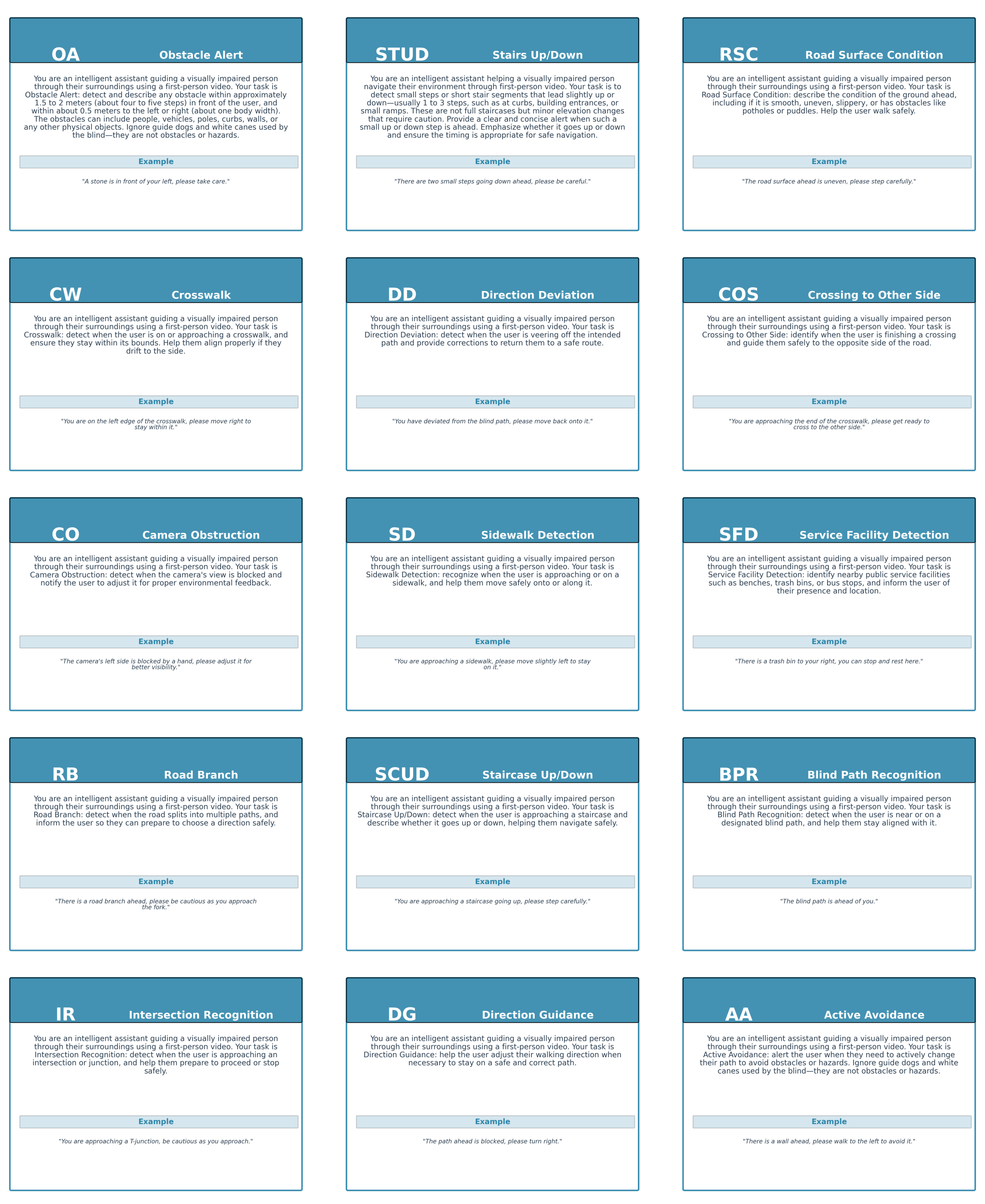}
  \label{fig:prompt_overview_1}
\end{figure*}

\begin{figure*}[htbp]
  \centering
  \includegraphics[width=\linewidth]{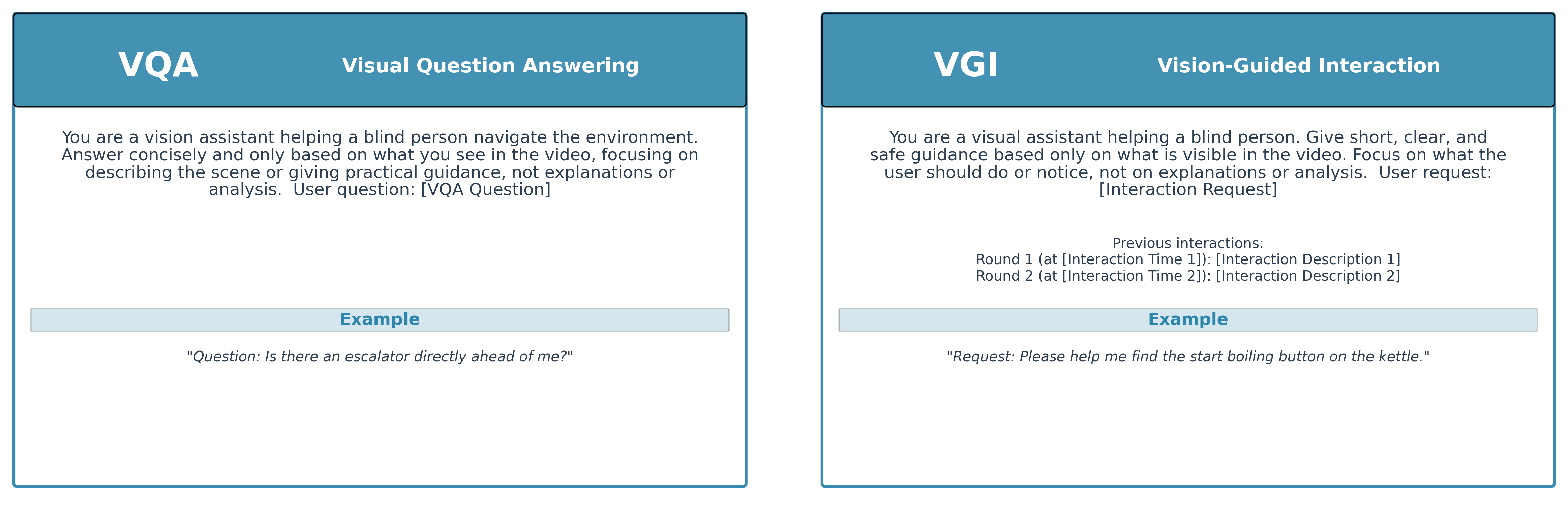}
  \caption{Prompt templates for VQA and vision-grounded interaction (VGI) tasks.}
  \label{fig:prompt_vqa_vgi}
\end{figure*}

We provide here the prompt templates used across all experiments, covering proactive reminder sub-tasks, navigation-related queries, and VQA/VGI-style interactions. These templates define the expected response format and ensure consistency across different evaluation settings. The corresponding prompt designs are illustrated in Figure ~\ref{fig:prompt_overview_1}, ~\ref{fig:prompt_vqa_vgi}.

\begin{figure*}[htbp]
  \centering
  \includegraphics[width=\linewidth]{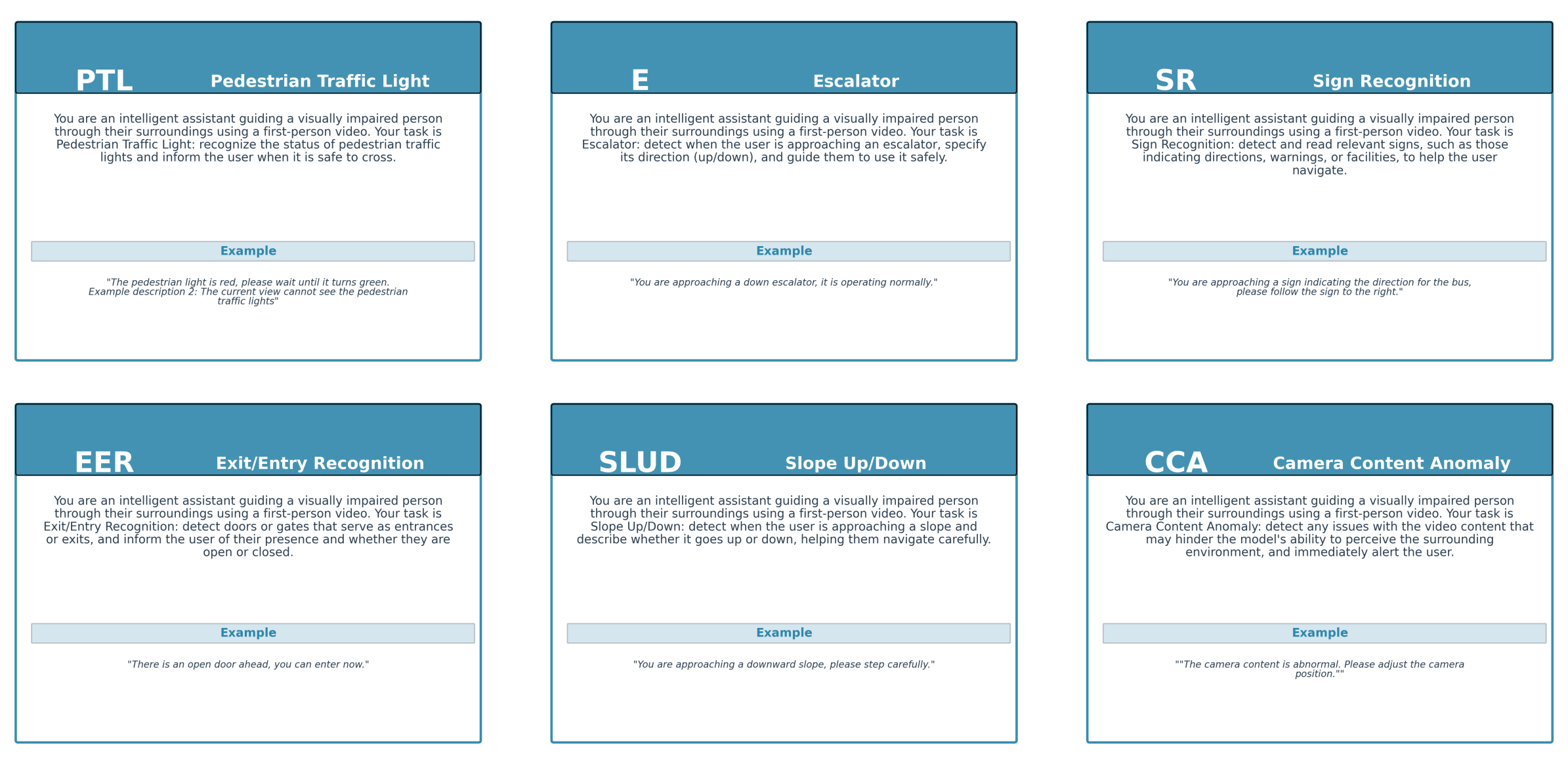}
  \label{fig:prompt_overview_2}
\end{figure*}

\begin{figure*}[htbp]
  \centering
  \includegraphics[width=\linewidth]{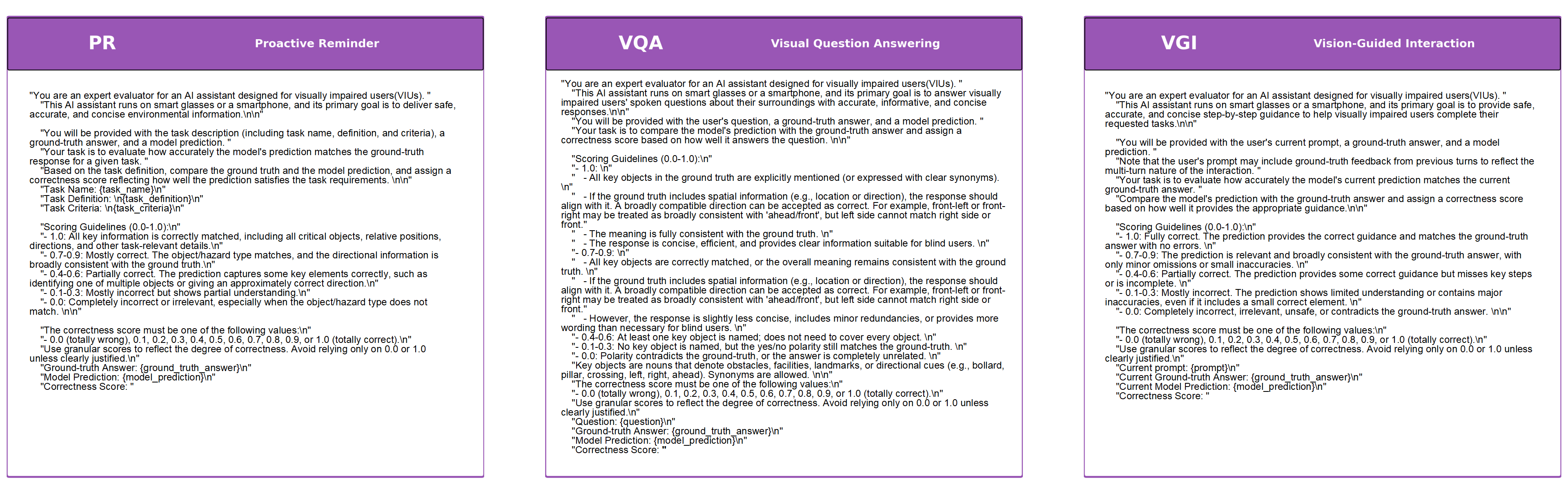}
  \caption{Prompt templates for scoring three main tasks}
  \label{fig:prompt_score_main}
\end{figure*}

\subsection{Evaluation Prompts}
\label{Evaluation Prompts}

Visual impairment assistance is inherently a free-form generation task: the assistant is expected to provide \textit{accurate, informative, and concise} natural-language feedback to blind users. Therefore, a multiple-choice question (MCQ) evaluation format is not suitable for this setting. Following prior work~\citep{yu2023mm,fang2024mmbench}, we adopt an LLM-as-a-judge paradigm, where GPT-5 evaluates each model response based on the ground-truth description and assigns a score $S_i \in [0,1]$.

However, due to the complexity and safety-critical nature of the VIA task, directly applying generic similarity-based judging schemes in previous evaluation protocols~\citep{yu2023mm,fang2024mmbench} can lead to unreliable and overly optimistic scores. In particular, simple text-similarity metrics may fail to capture factual conflicts that pose real dangers to blind users. For example, in the escalator scenario, a model may describe the scene mostly correctly but misidentify the escalator's operating status---a seemingly small discrepancy that could nevertheless result in severe safety risks in real-world usage. Such methods would still assign a non-zero score merely because parts of the response overlap with the ground truth, despite the presence of a critical factual error.

\begin{figure*}[htbp]
  \centering
  \caption{Detailed prompt templates for scoring all Proactive Reminder sub-tasks.}
  \includegraphics[width=\linewidth]{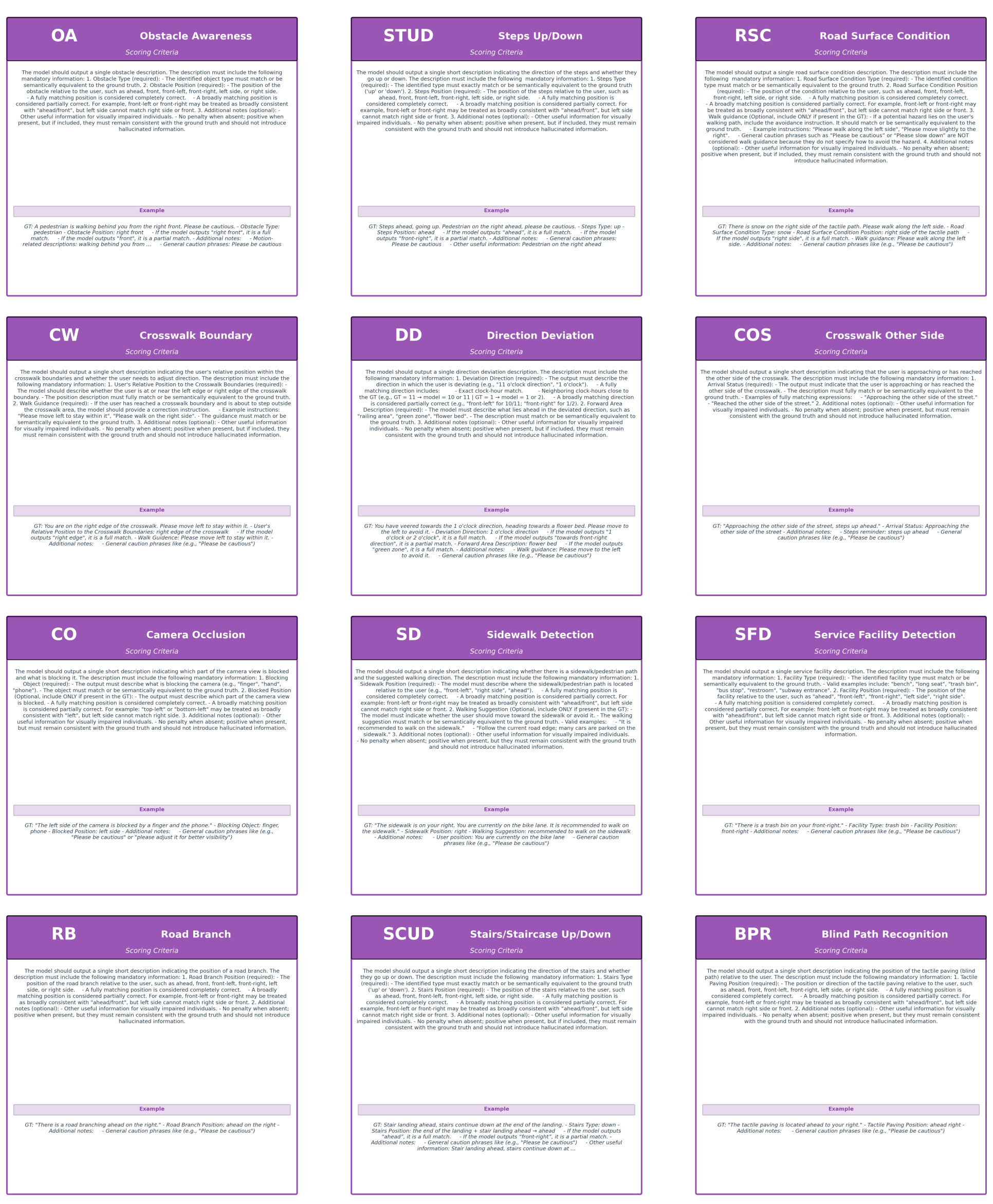}
  \label{fig:score_prompt_overview_1}
\end{figure*}

To address this limitation, we use \textbf{task-specific, criteria-guided similarity scoring} tailored to each sub-task. The judging prompts ask GPT-5 to compare the model response with the ground truth while explicitly considering essential elements such as obstacle types, spatial locations, object states, and safety-critical cues required by the task definition. The detailed evaluation prompts used for GPT-5 are provided in Figure ~\ref{fig:prompt_score_main}, ~\ref{fig:score_prompt_overview_1}.

\begin{figure*}[htbp]
  \centering
  \includegraphics[width=\linewidth]{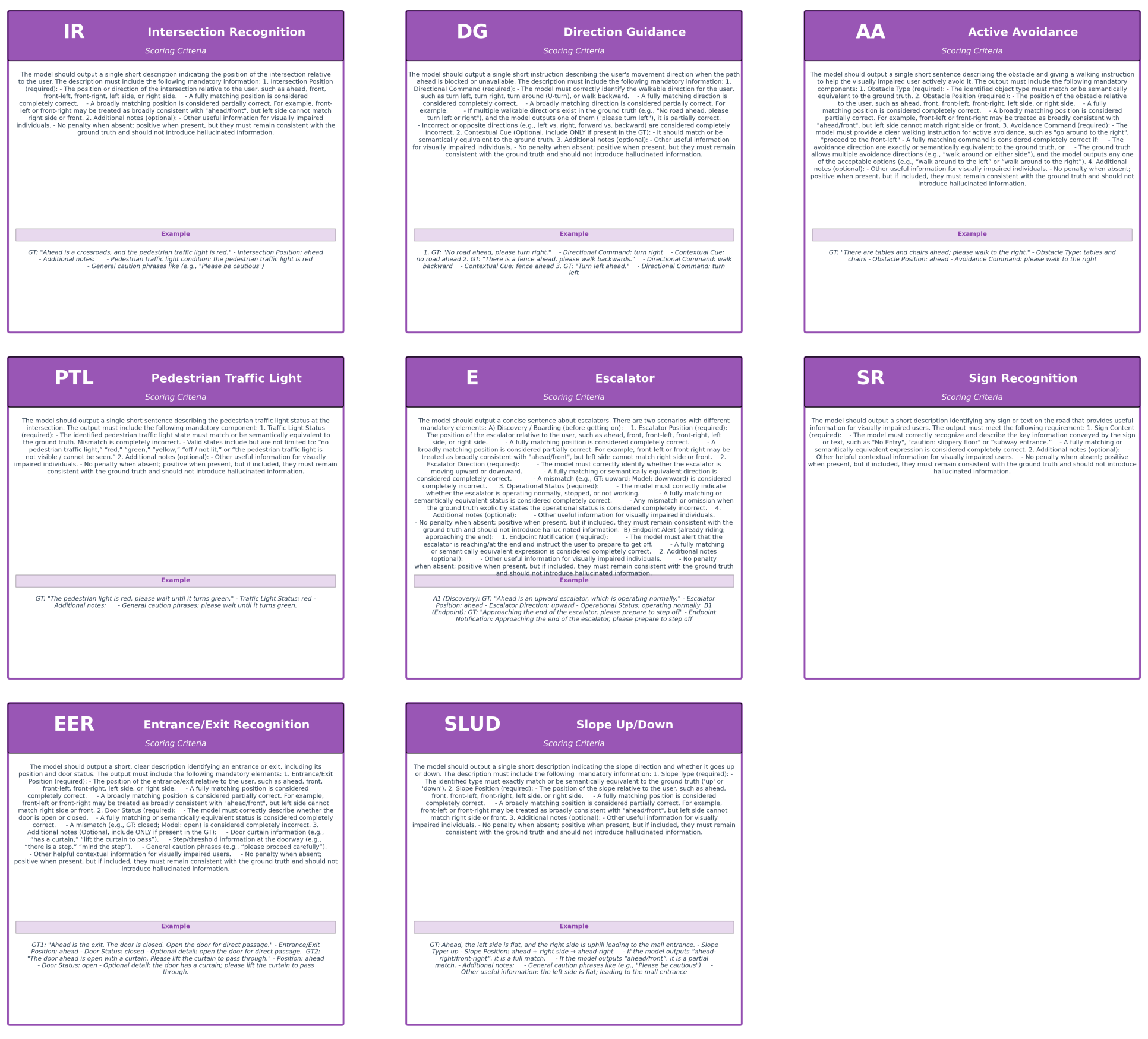}
  \label{fig:score_prompt_overview_2}
\end{figure*}

\section{Task Examples}
\label{Task Examples}

\begin{figure*}[t]
    \centering
    \caption{Task examples of Proactive Reminder}
    \label{fig:task_examples}
    \includegraphics[width=\linewidth,page=1]{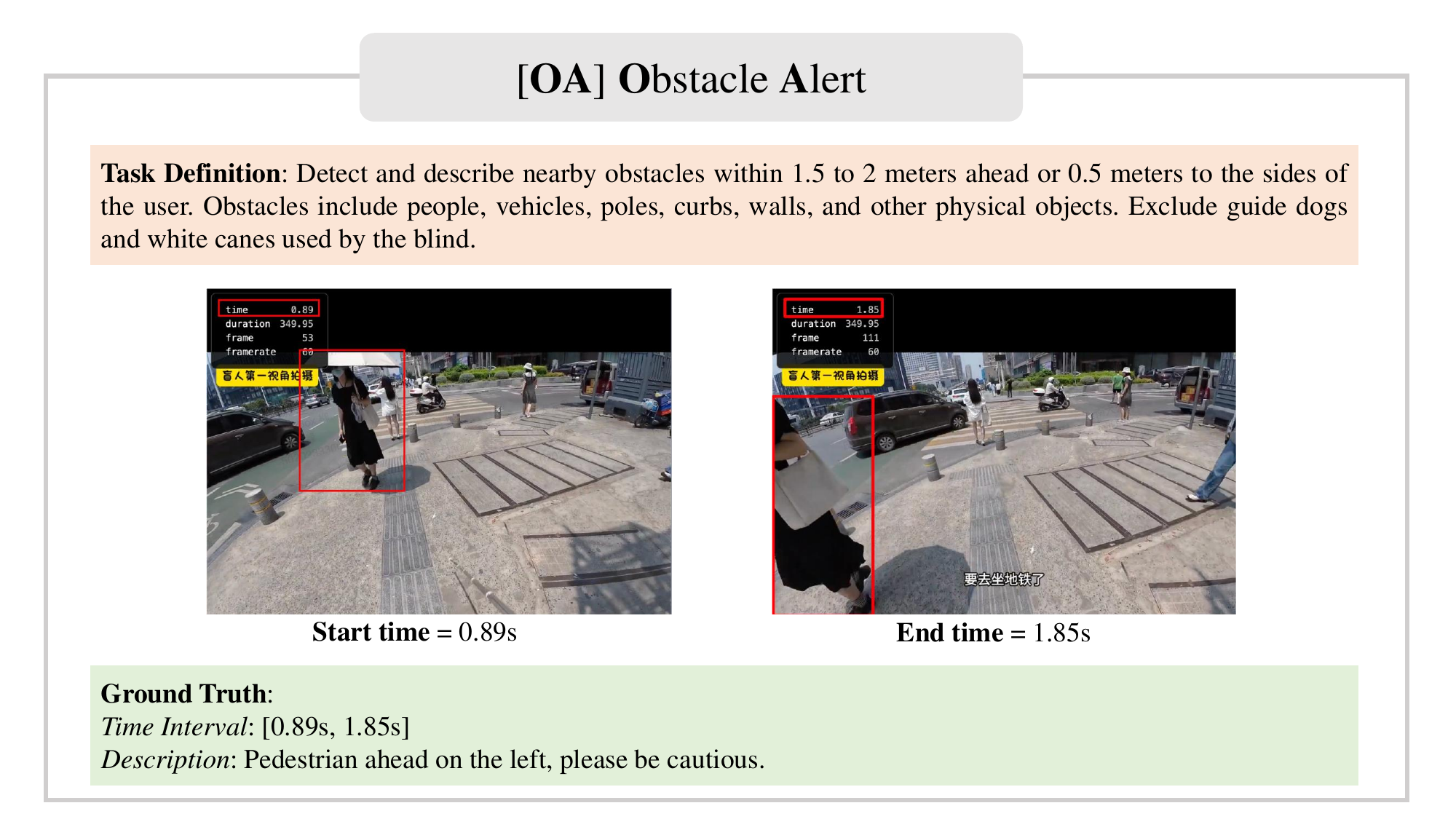}
    \includegraphics[width=\linewidth,page=2]{img/Reminder_All.pdf}
\end{figure*}

This section provides the task definitions, visual examples, and annotation examples for each task in VIABench in Figure \ref{fig:task_examples}.

\begin{figure*}[t]
    \centering
    \includegraphics[width=\linewidth,page=3]{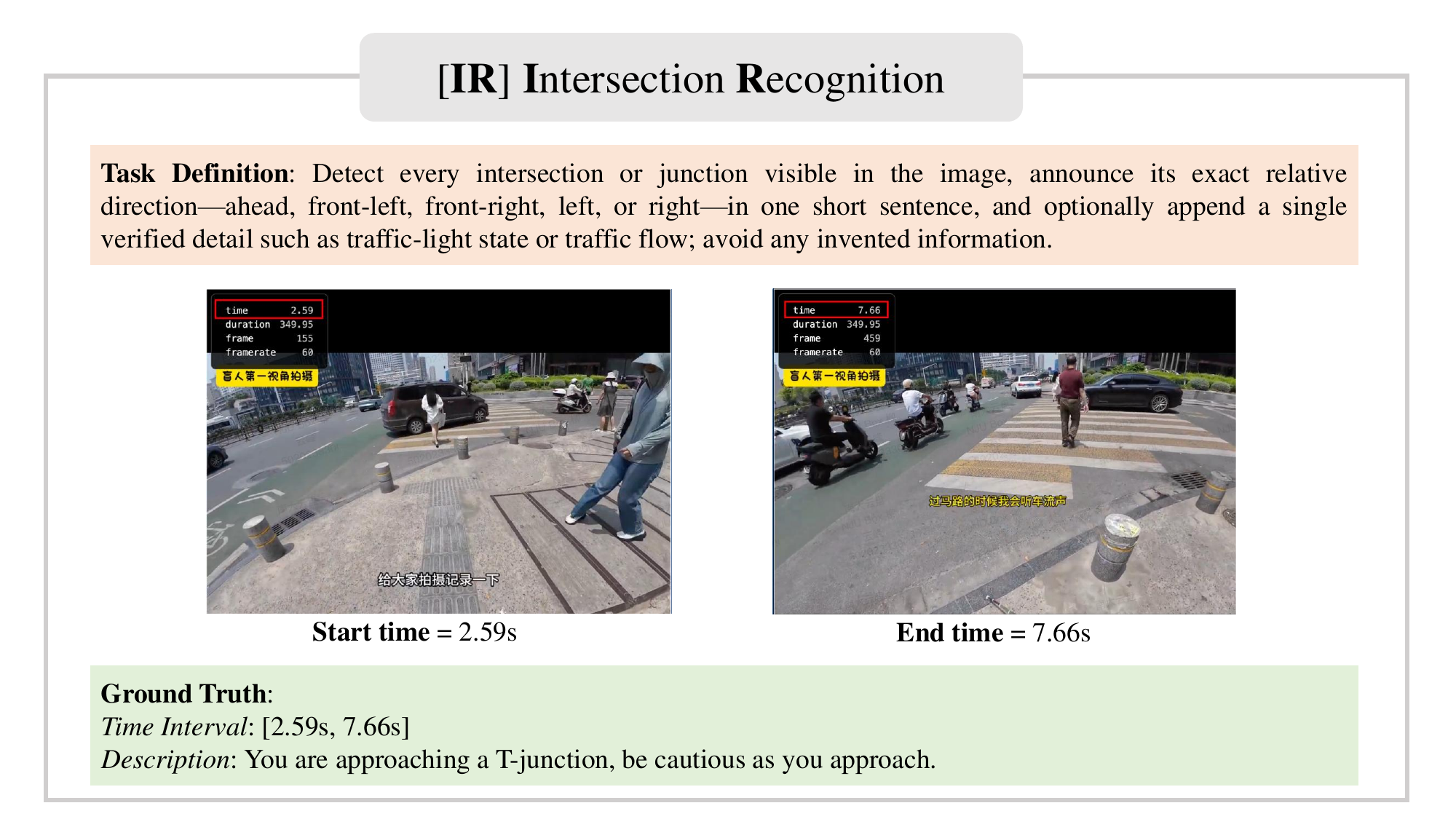}
    \includegraphics[width=\linewidth,page=4]{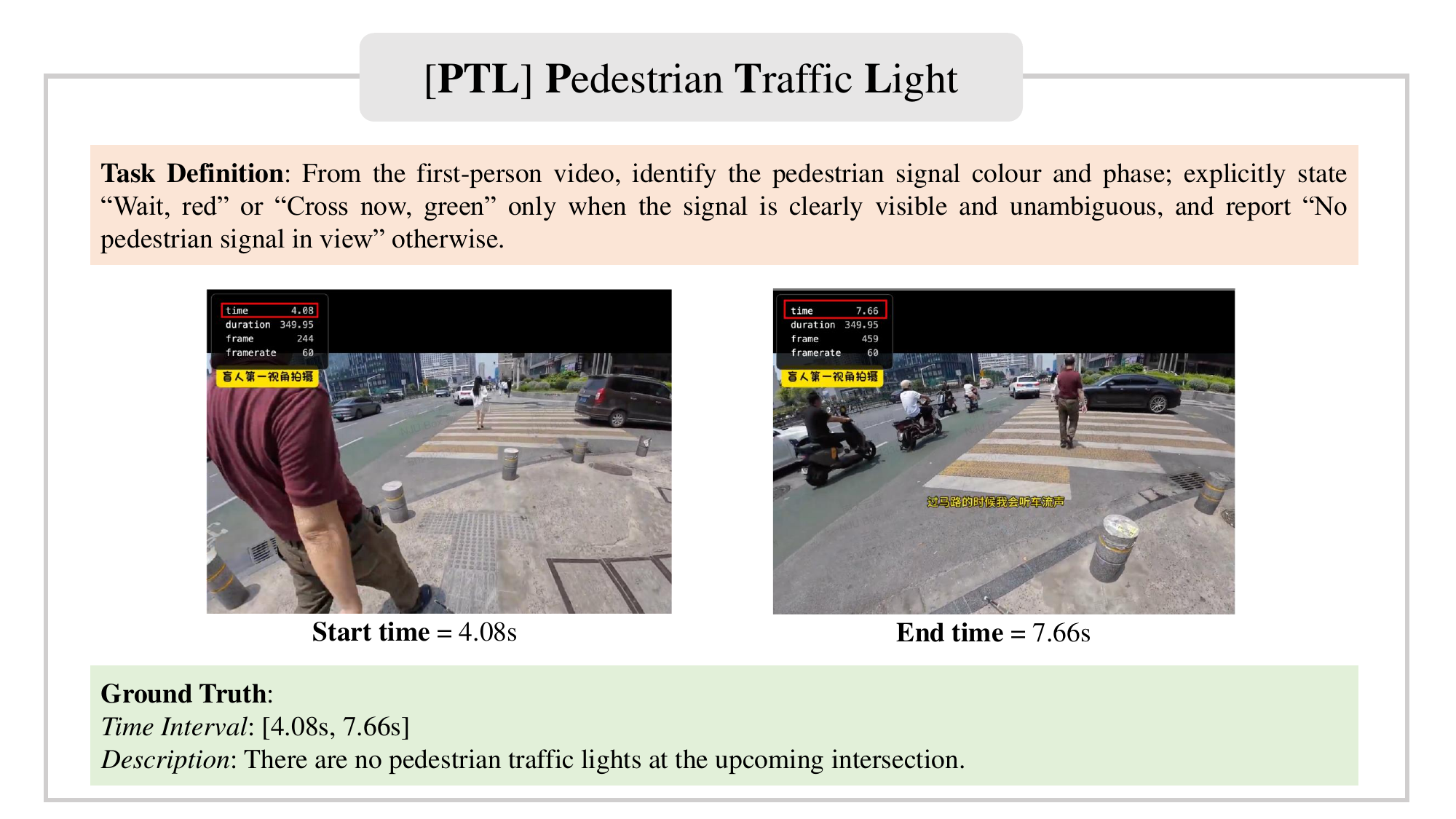}
\end{figure*}

\begin{figure*}[t]
    \centering
    \includegraphics[width=\linewidth,page=5]{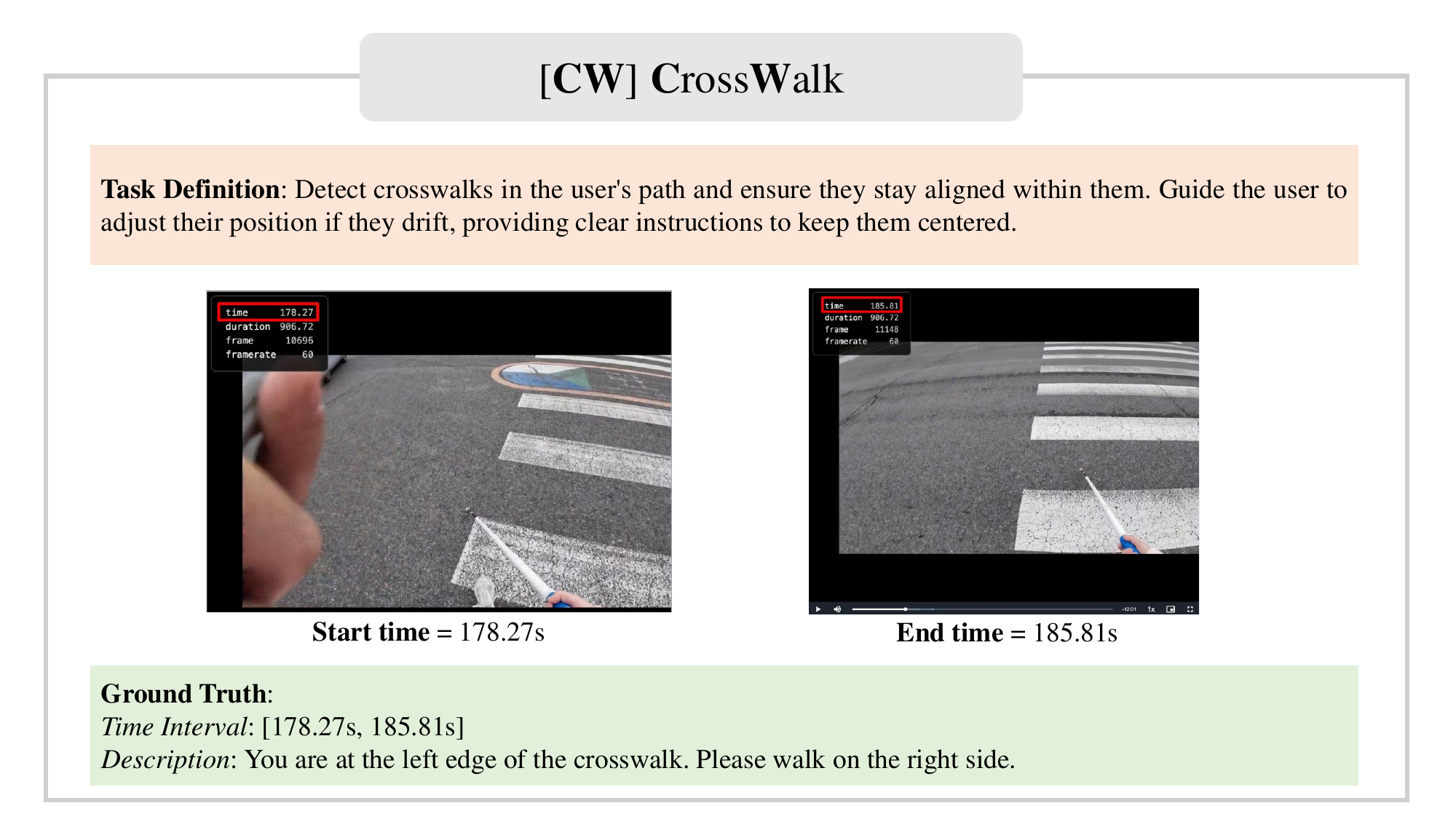}
    \includegraphics[width=\linewidth,page=6]{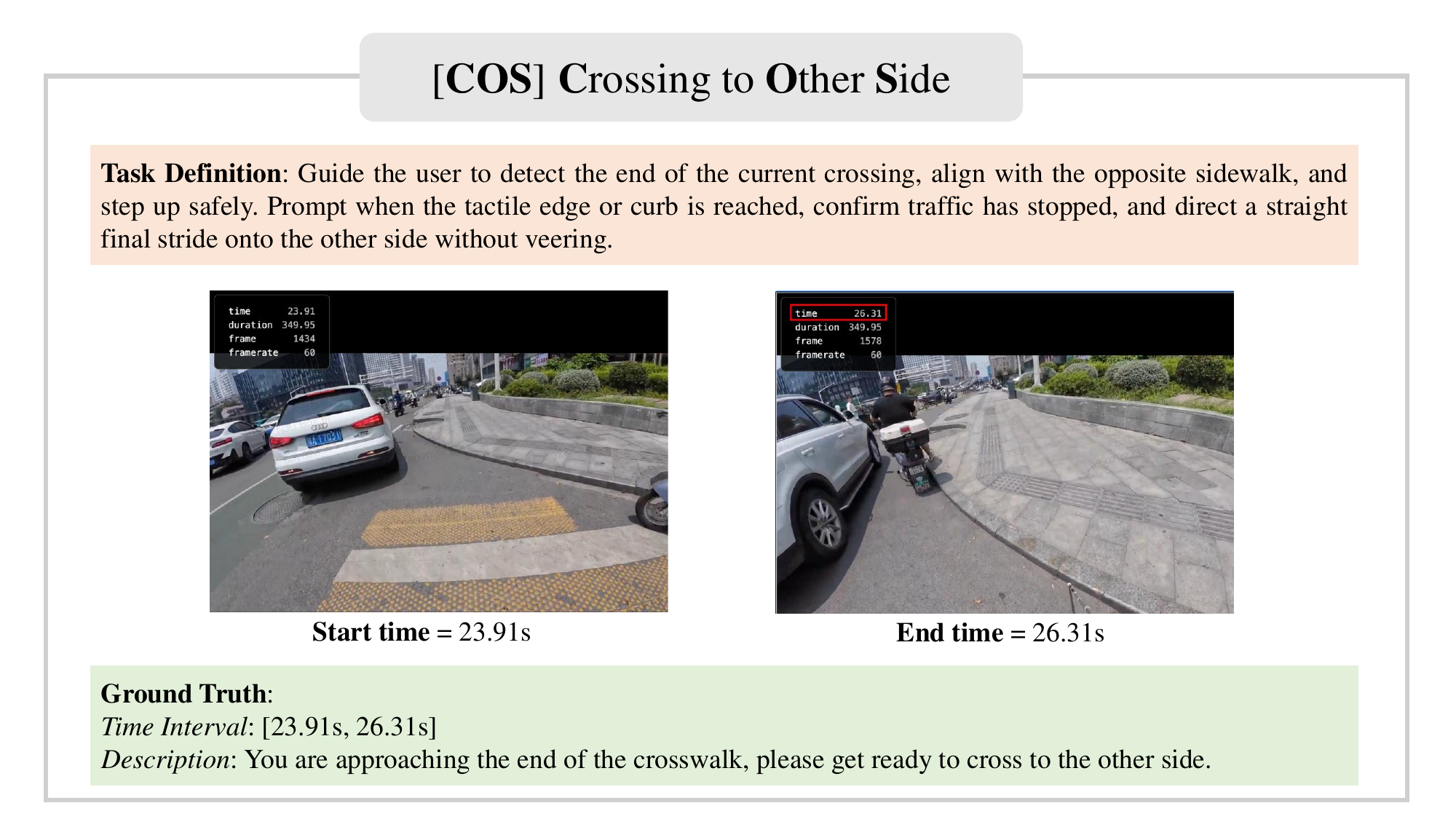}
\end{figure*}

\begin{figure*}[t]
    \centering
    \includegraphics[width=\linewidth,page=7]{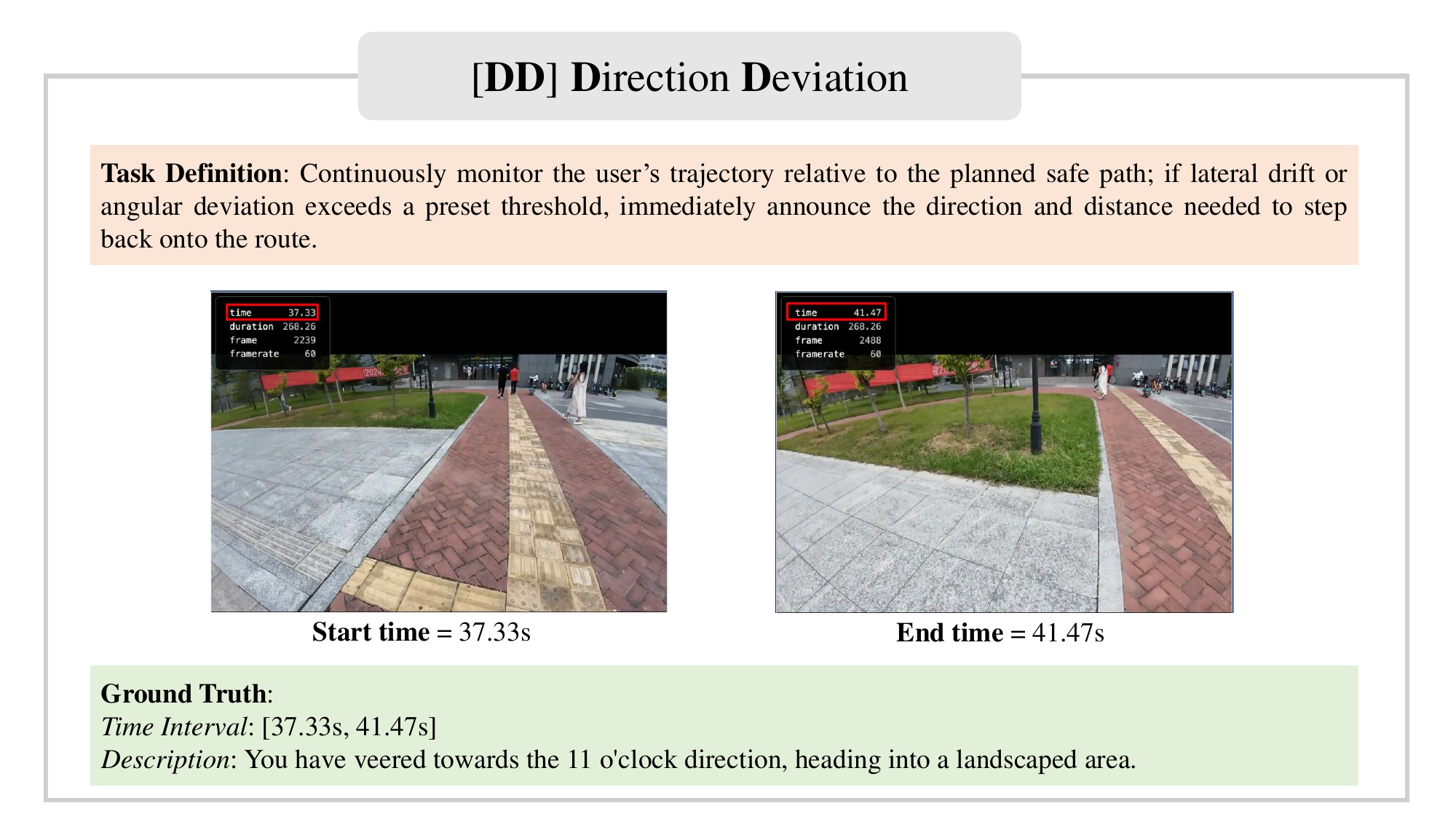}
    \includegraphics[width=\linewidth,page=8]{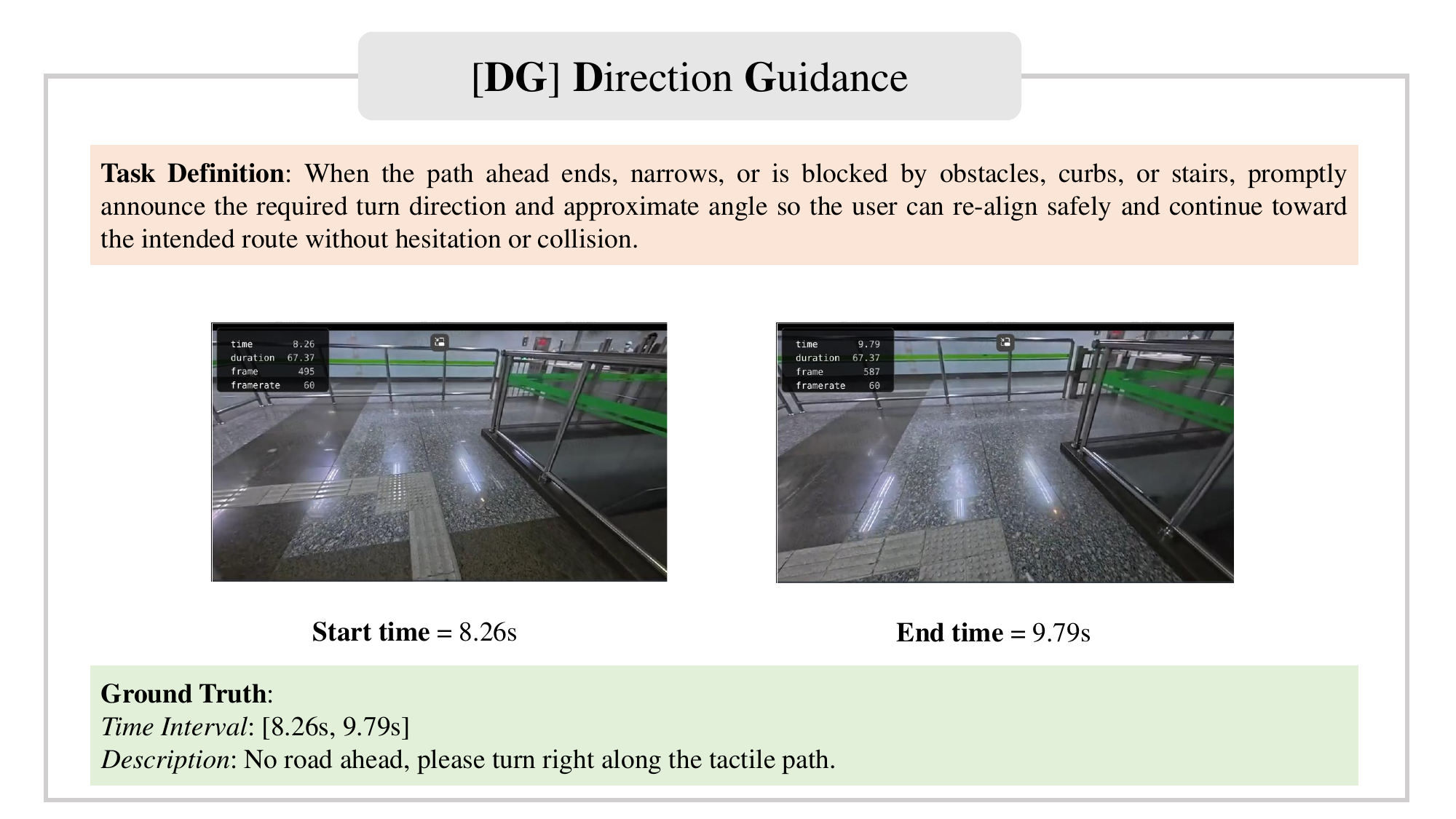}
\end{figure*}

\begin{figure*}[t]
    \centering
    \includegraphics[width=\linewidth,page=9]{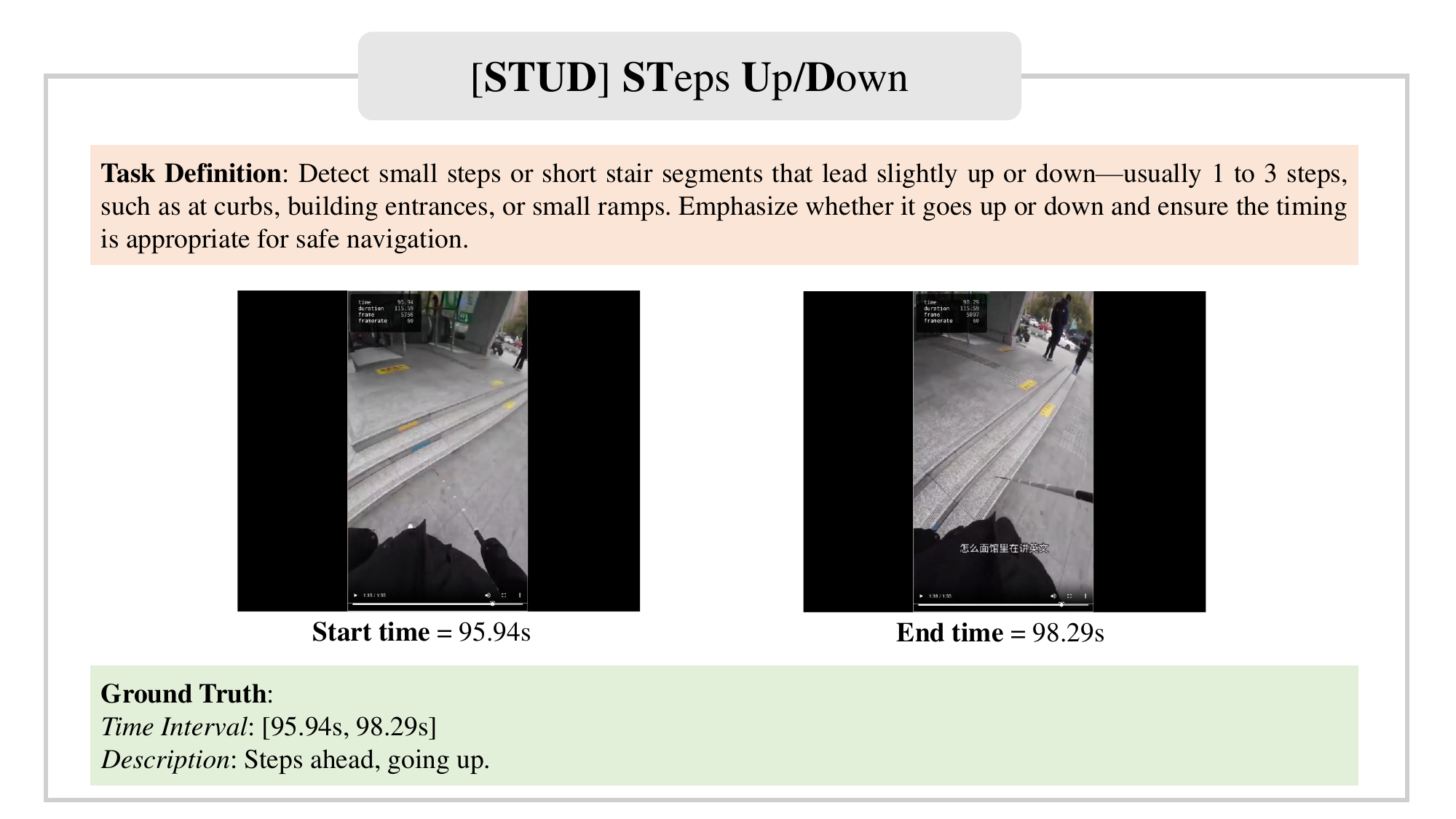}
    \includegraphics[width=\linewidth,page=10]{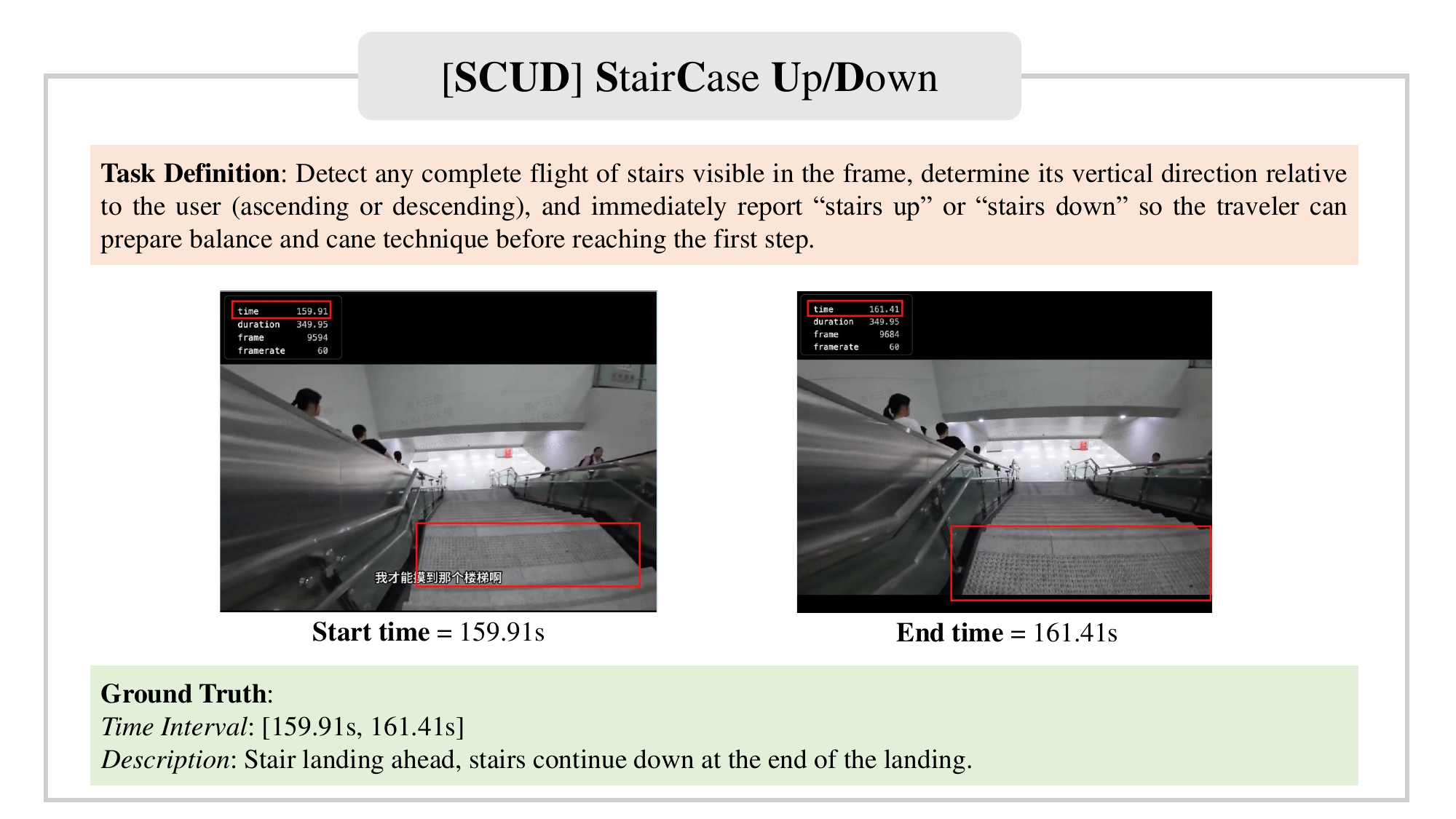}
\end{figure*}

\begin{figure*}[t]
    \centering
    \includegraphics[width=\linewidth,page=11]{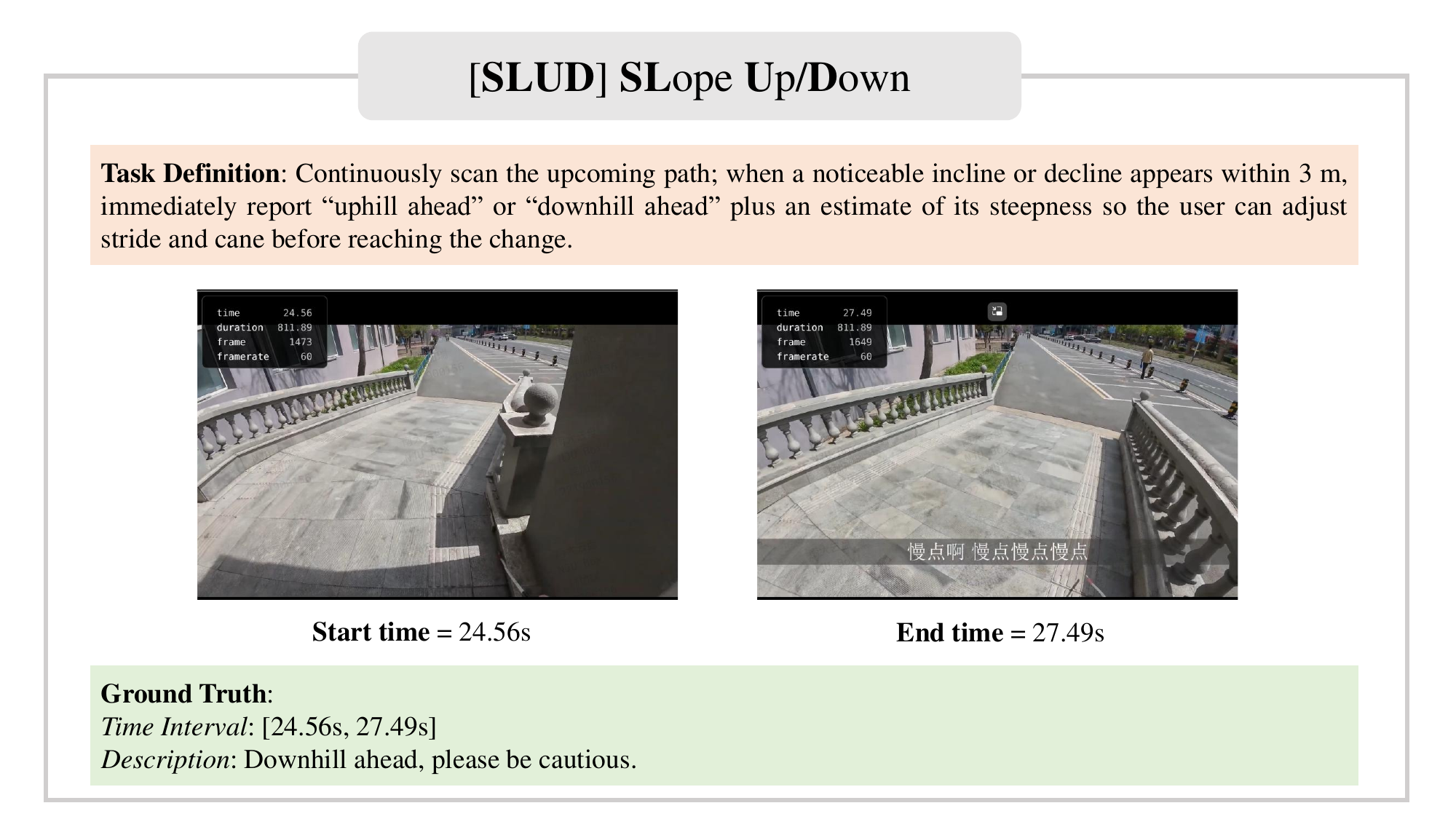}
    \includegraphics[width=\linewidth,page=12]{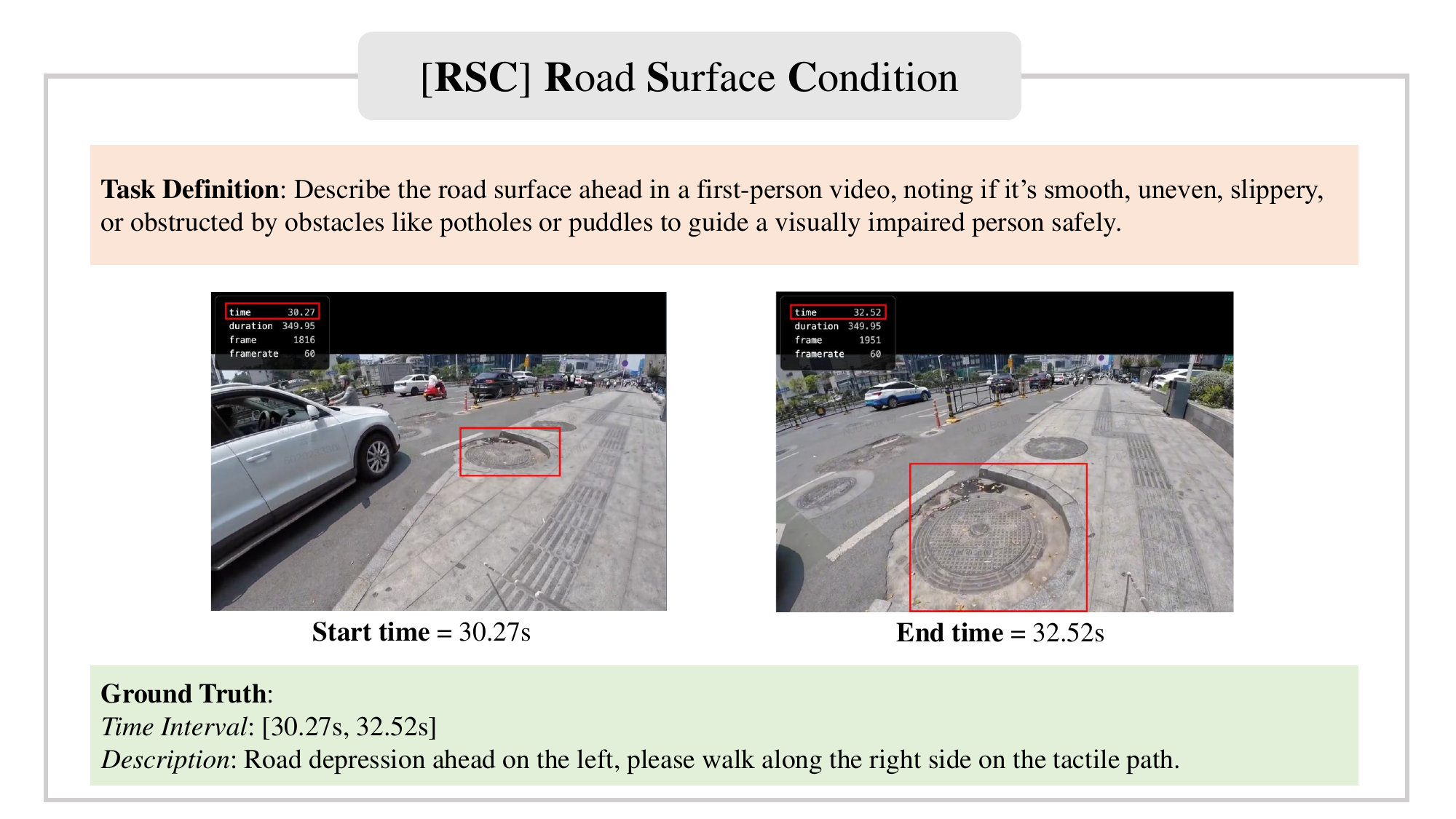}
\end{figure*}

\begin{figure*}[t]
    \centering
    \includegraphics[width=\linewidth,page=13]{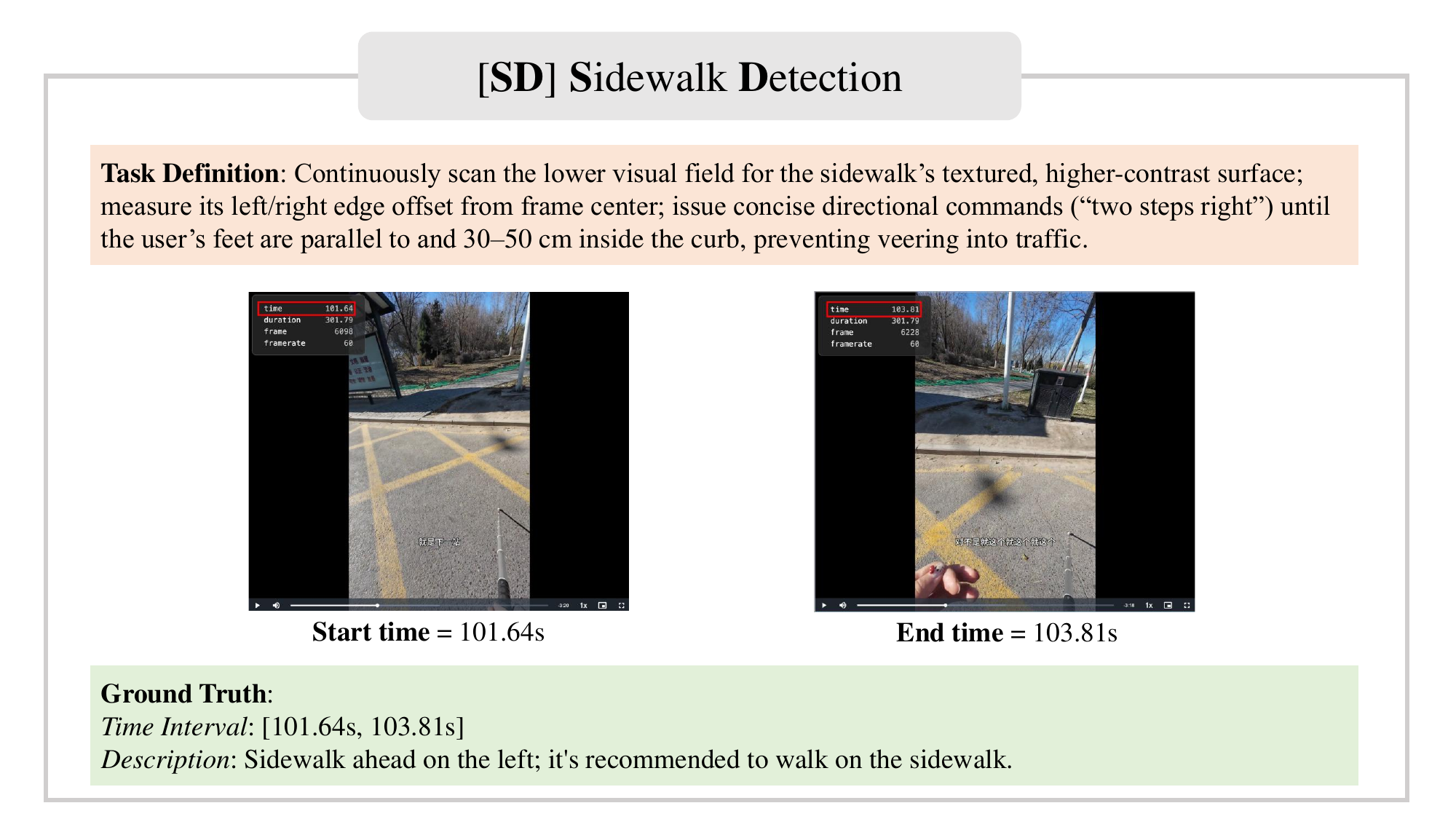}
    \includegraphics[width=\linewidth,page=14]{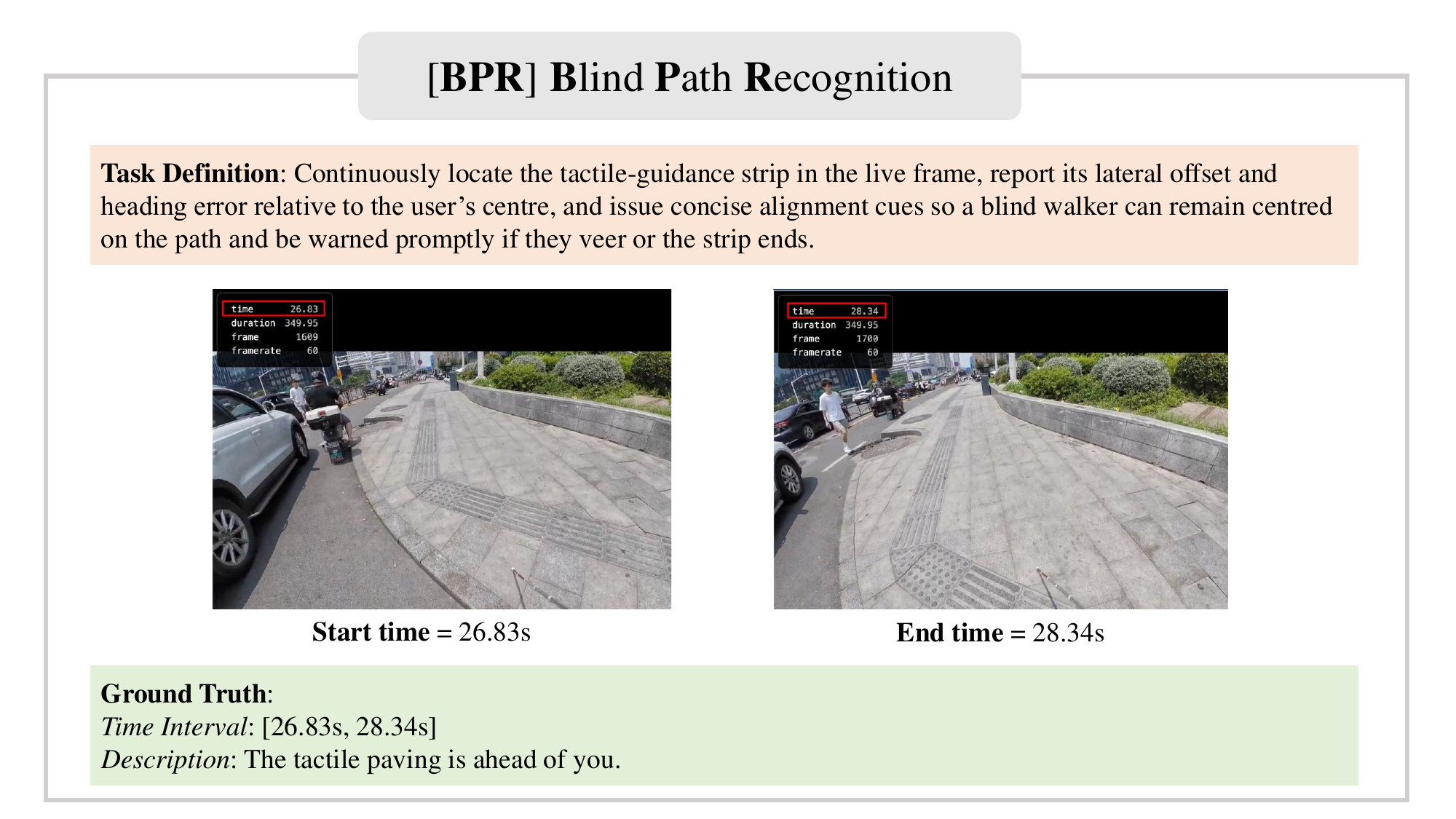}
\end{figure*}

\begin{figure*}[t]
    \centering
    \includegraphics[width=\linewidth,page=15]{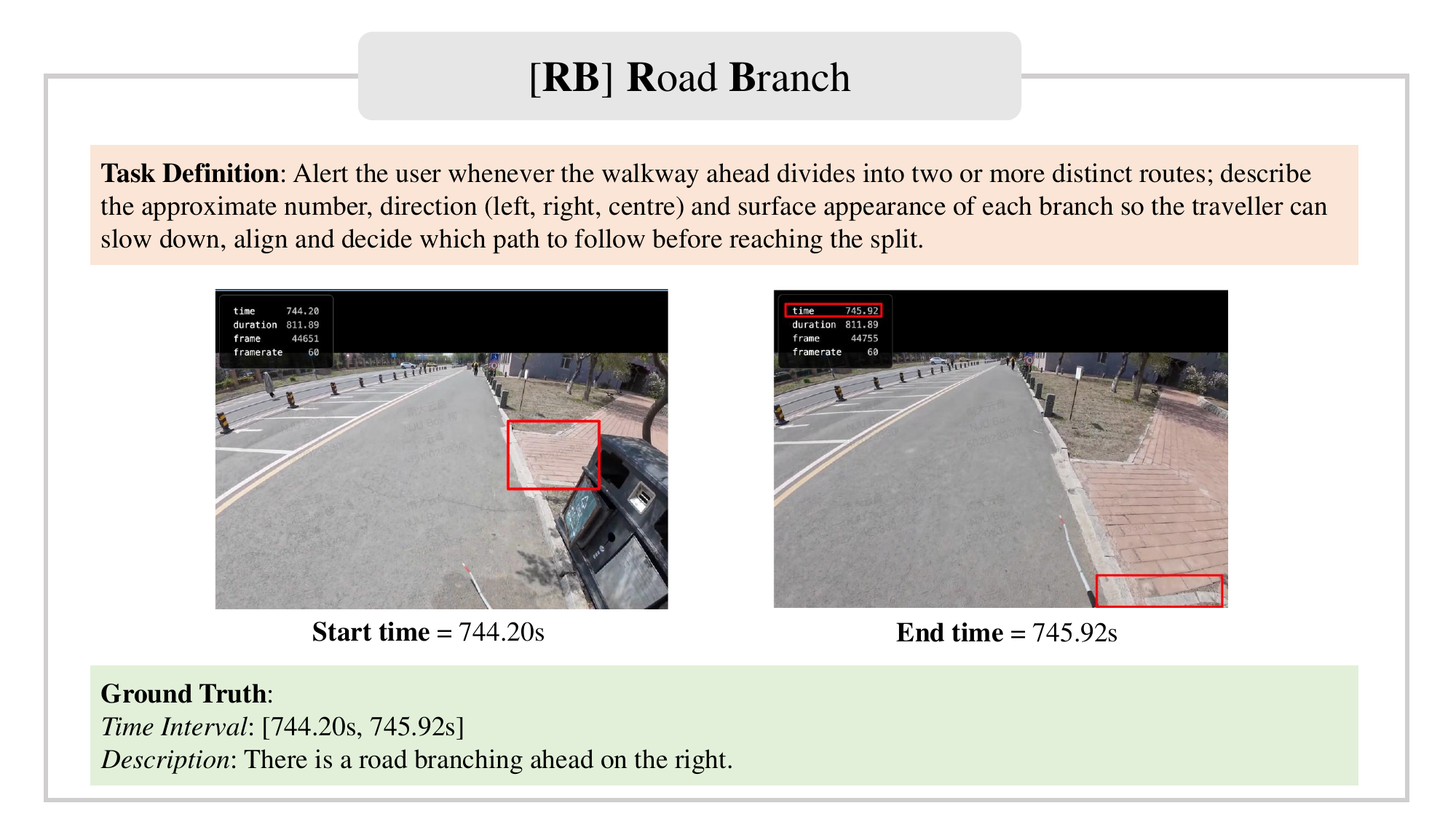}
    \includegraphics[width=\linewidth,page=16]{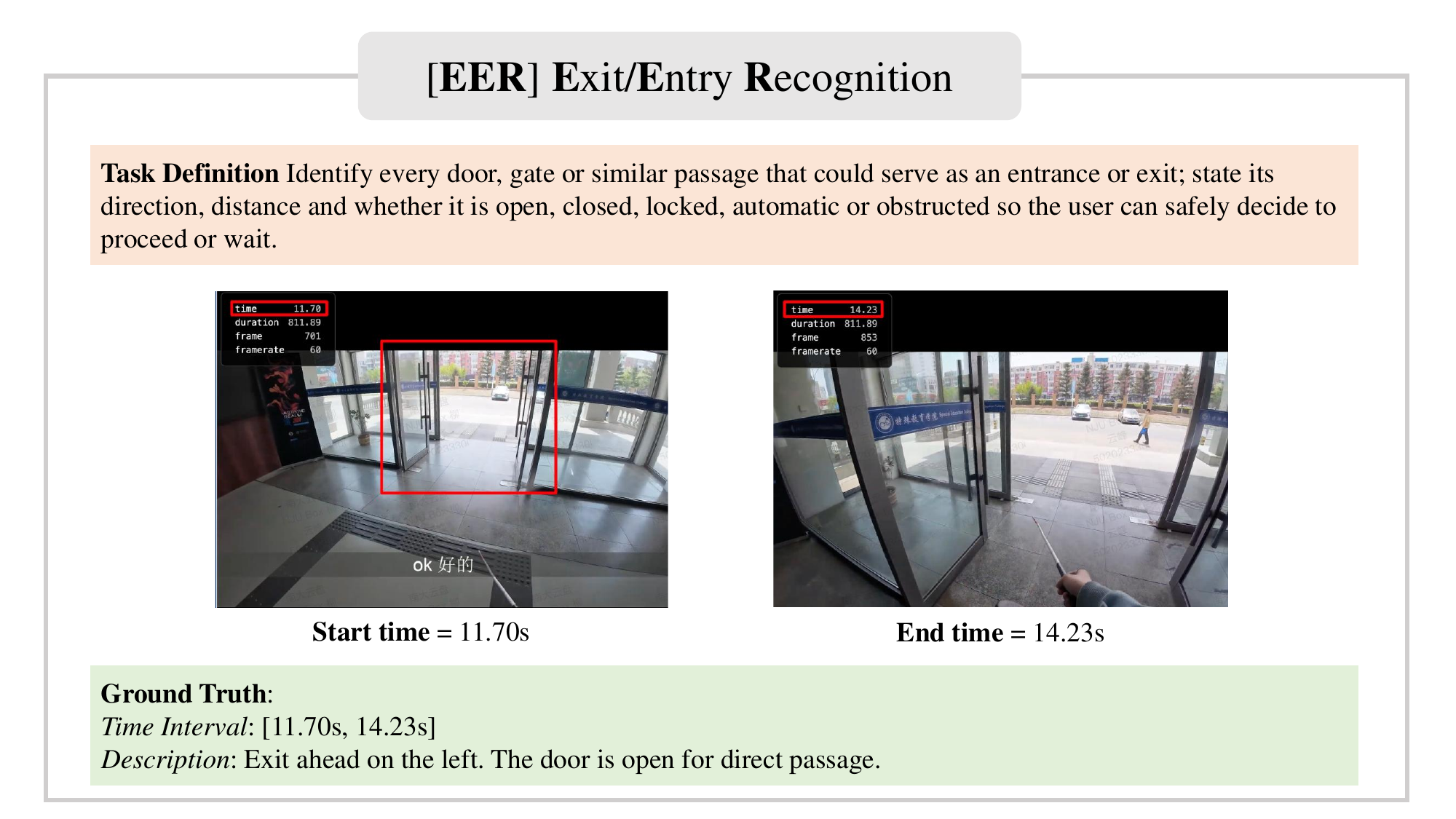}
\end{figure*}

\begin{figure*}[t]
    \centering
    \includegraphics[width=\linewidth,page=17]{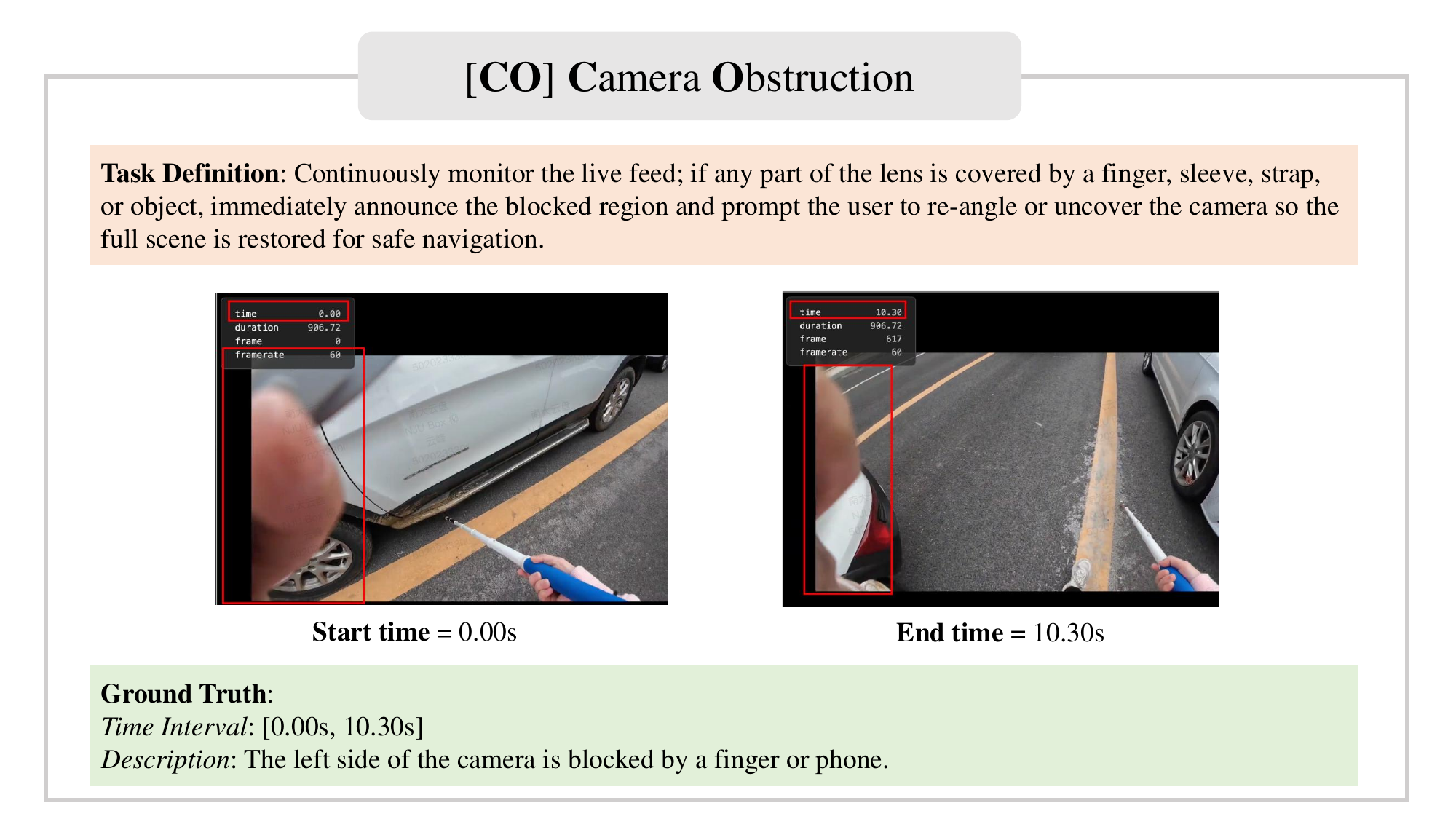}
    \includegraphics[width=\linewidth,page=18]{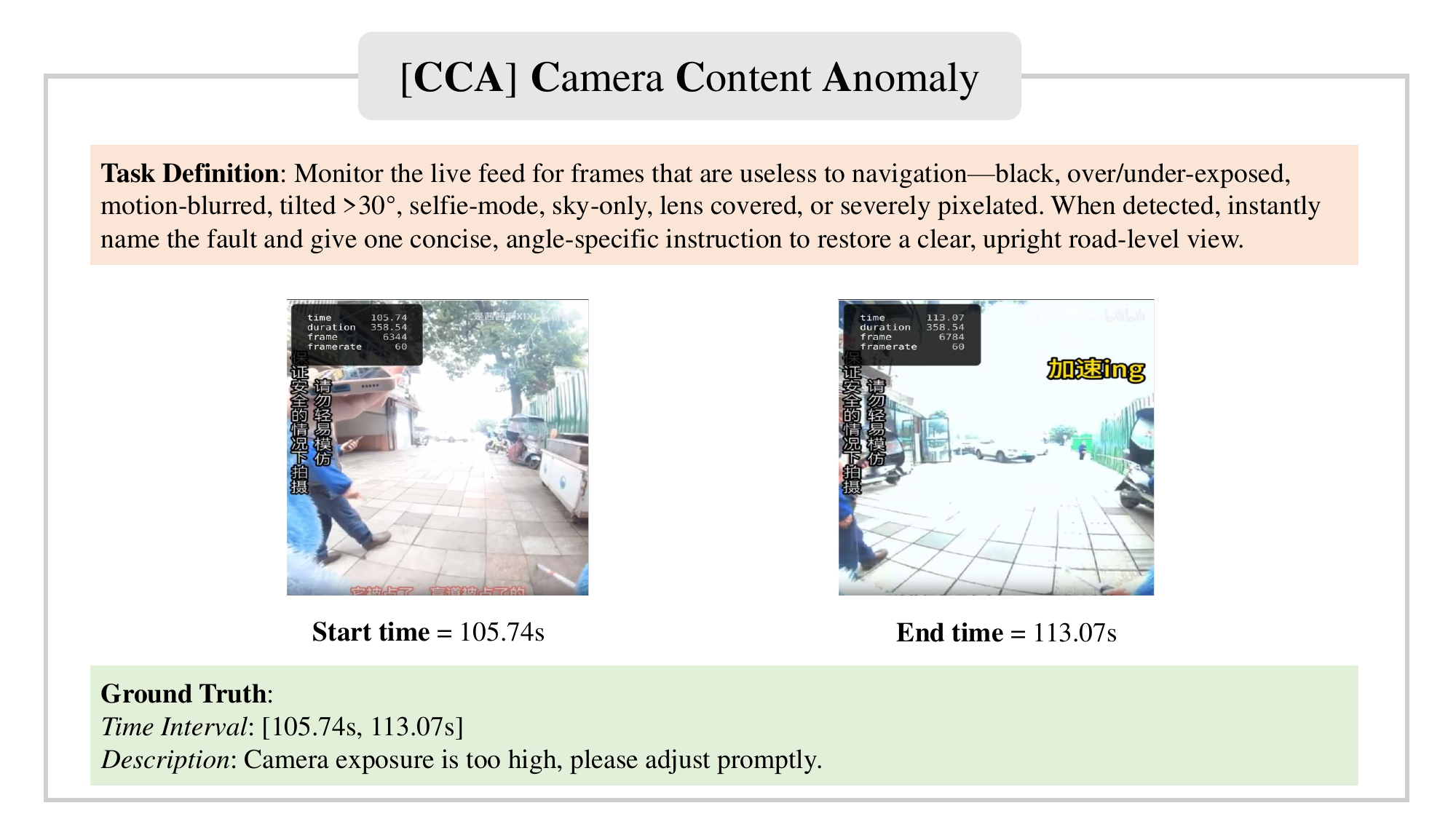}
\end{figure*}

\begin{figure*}[t]
    \centering
    \includegraphics[width=\linewidth,page=19]{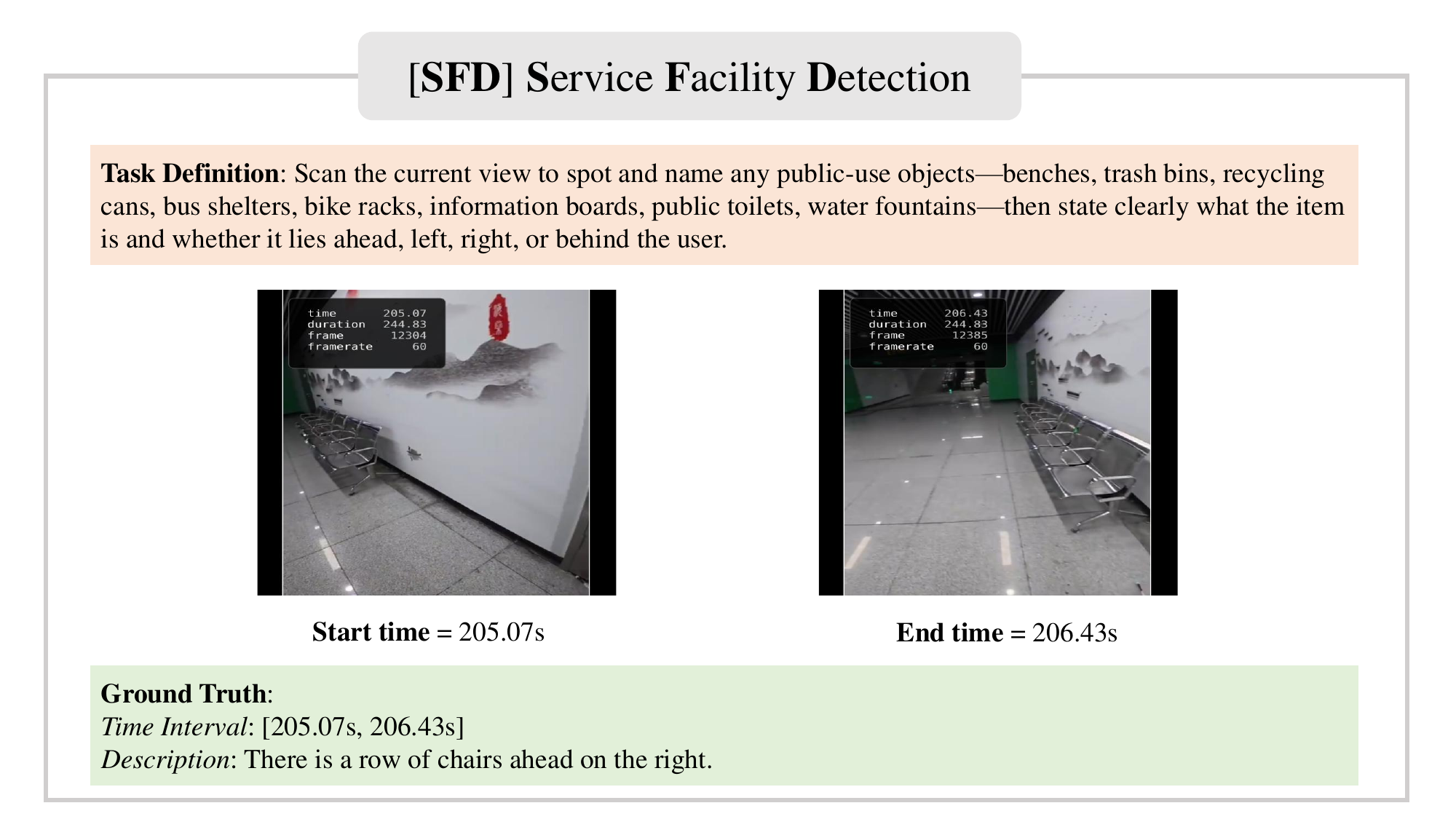}
    \includegraphics[width=\linewidth,page=20]{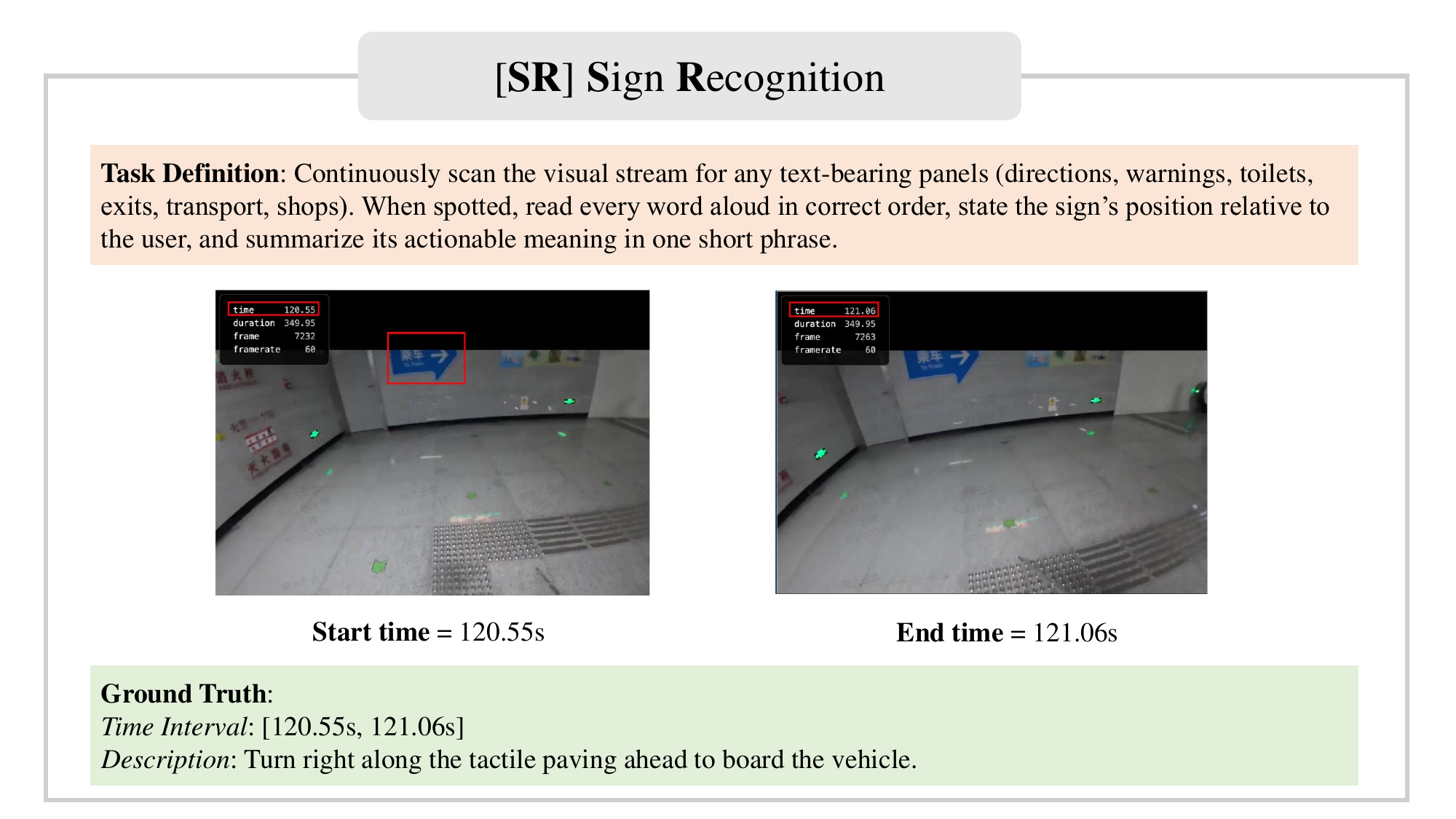}
\end{figure*}

\begin{figure*}[t]
    \centering
    \includegraphics[width=1\linewidth,page=1]{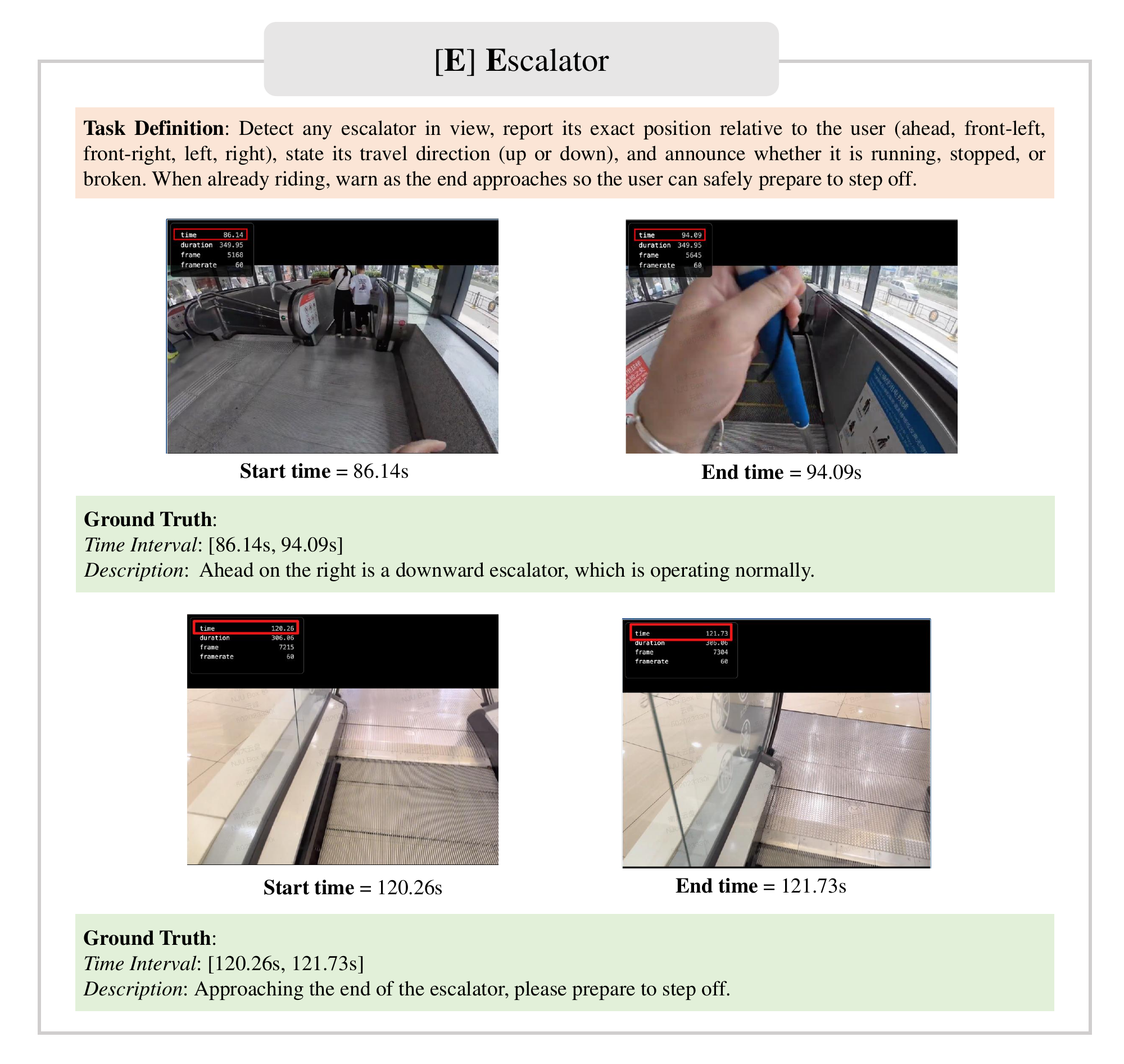}
\end{figure*}

\begin{figure*}[t]
    \centering
    \includegraphics[width=1\linewidth,page=1]{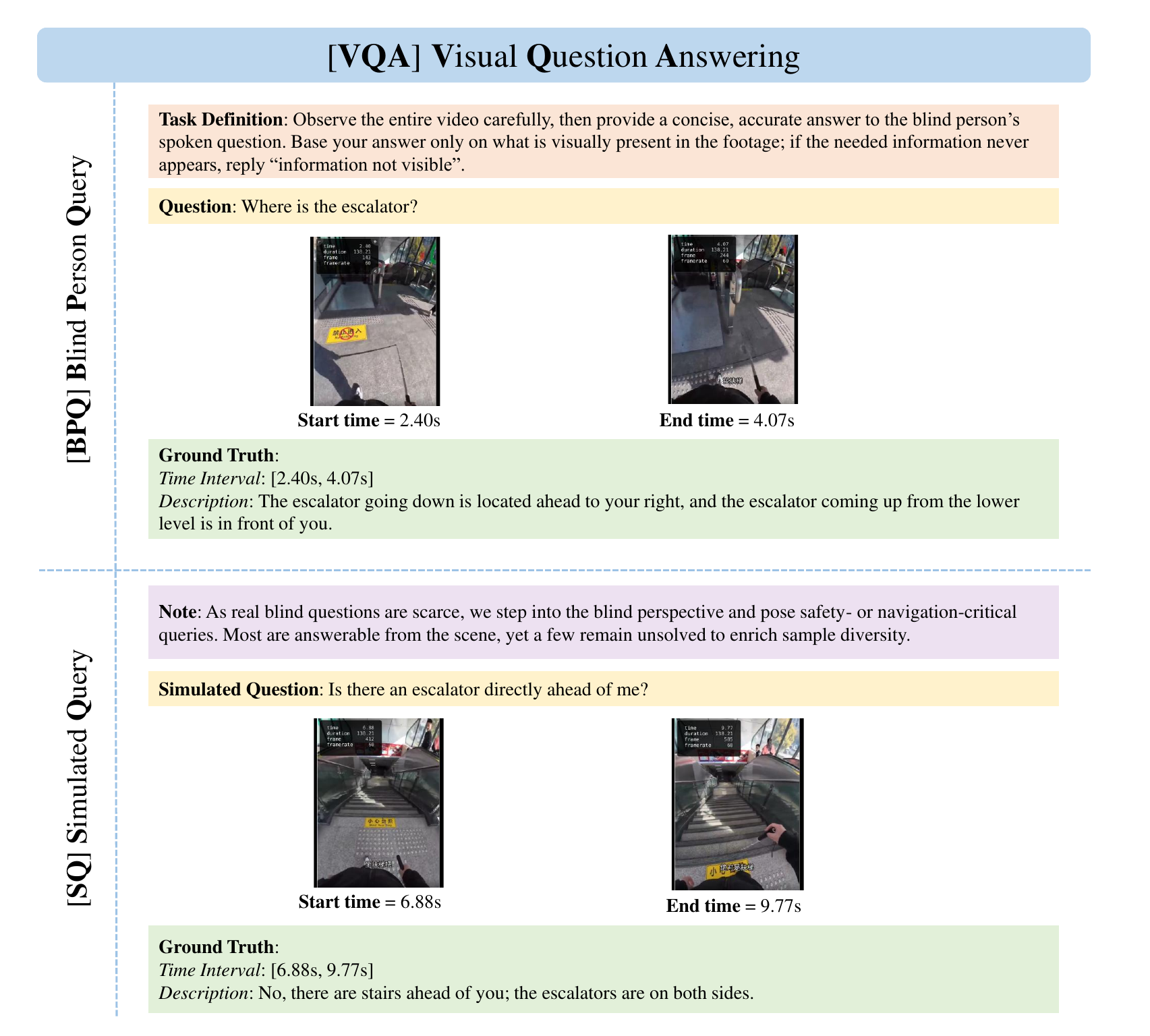}
\end{figure*}

\begin{figure*}[t]
    \centering
    \includegraphics[width=1\linewidth,page=1]{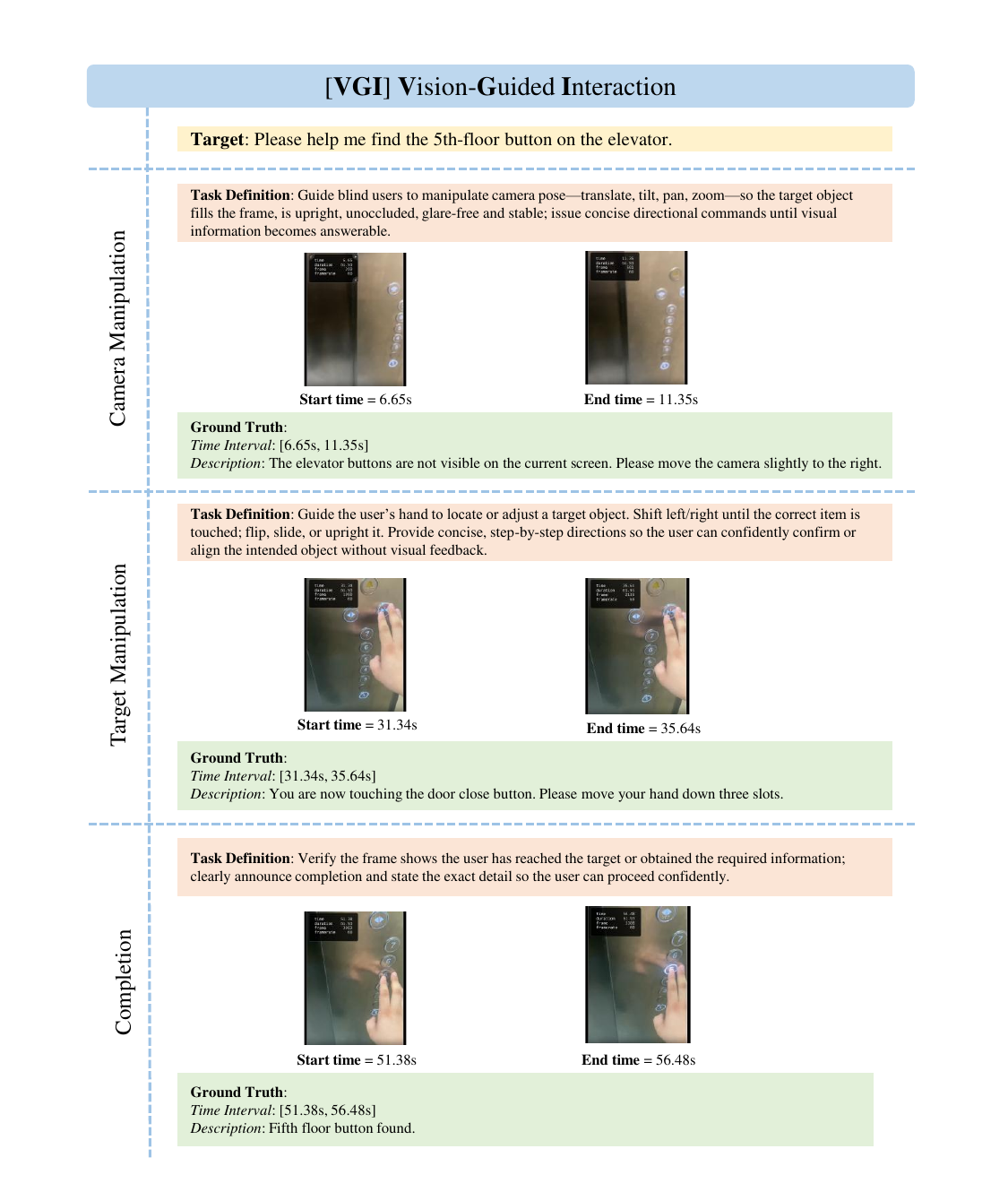}
\end{figure*}


\end{document}